\documentclass{article}

\usepackage[preprint, nonatbib]{neurips_2023}

\usepackage{times}
\usepackage{epsfig}
\usepackage{graphicx}
\usepackage{amsmath}
\usepackage{amssymb}
\usepackage{graphicx}
\usepackage{float}
\usepackage{adjustbox}
\usepackage{multirow}
\usepackage{caption}
\usepackage{subcaption}

\usepackage[utf8]{inputenc} 
\usepackage[T1]{fontenc}    
\usepackage{hyperref}       
\usepackage{url}            
\usepackage{booktabs}       
\usepackage{amsfonts}       
\usepackage{nicefrac}       
\usepackage{microtype}      
\usepackage{xcolor}         

\newcommand{\ours}[0]{CLIP-Layout}
\newcommand{\adj}[1]{\adjustbox{trim={0.1\width} {0.1\height} {0.1\depth} {0\totalheight},clip}{#1}}

\title{\ours{}: Style-Consistent Indoor Scene Synthesis with Semantic Furniture Embedding}

\author{%
Jingyu Liu \quad
Wenhan Xiong \quad
Ian Jones \quad
Yixin Nie \quad
Anchit Gupta \quad
Barlas O\u{g}uz \\
Meta AI \\
{\texttt \{jingyuliu, xwhan, ijones, ynie, anchit, barlaso\}@meta.com}
}

\begin{document}

\makeatletter
\let\@oldmaketitle\@maketitle%
\renewcommand{\@maketitle}{\@oldmaketitle%
    \vspace{-10pt}
    \rotatebox[origin=c]{90}{Automatic}
    \begin{minipage}{0.195\textwidth}
        \centering
        \adj{\includegraphics[width=\textwidth]{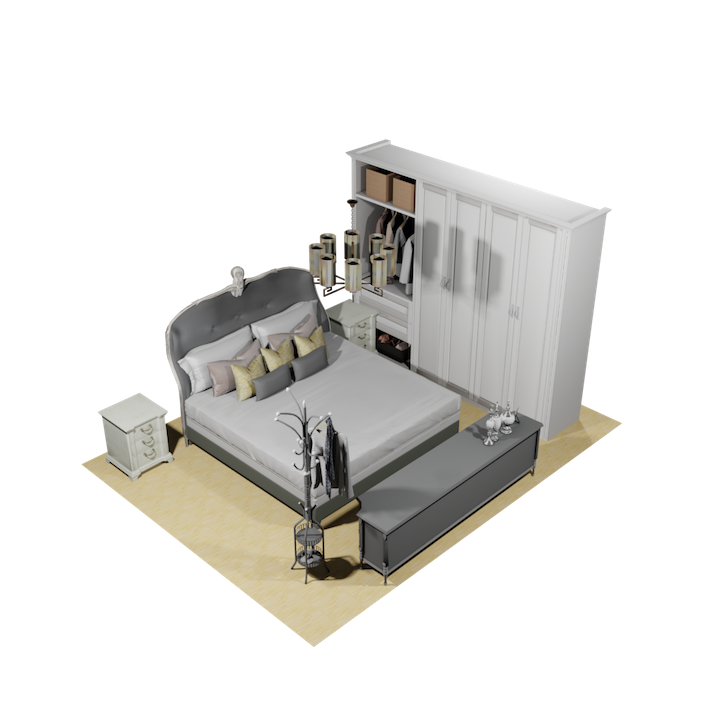}}
    \end{minipage}\hfill
    \begin{minipage}{0.195\textwidth}
        \centering
        \adj{\includegraphics[width=\textwidth]{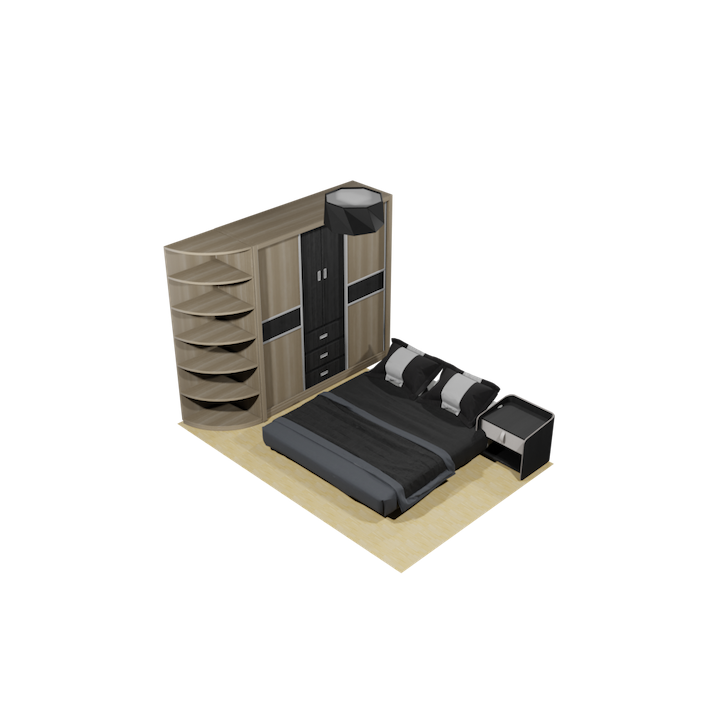}}
    \end{minipage}\hfill
    \begin{minipage}{0.195\textwidth}
        \centering
        \adj{\includegraphics[width=\textwidth]{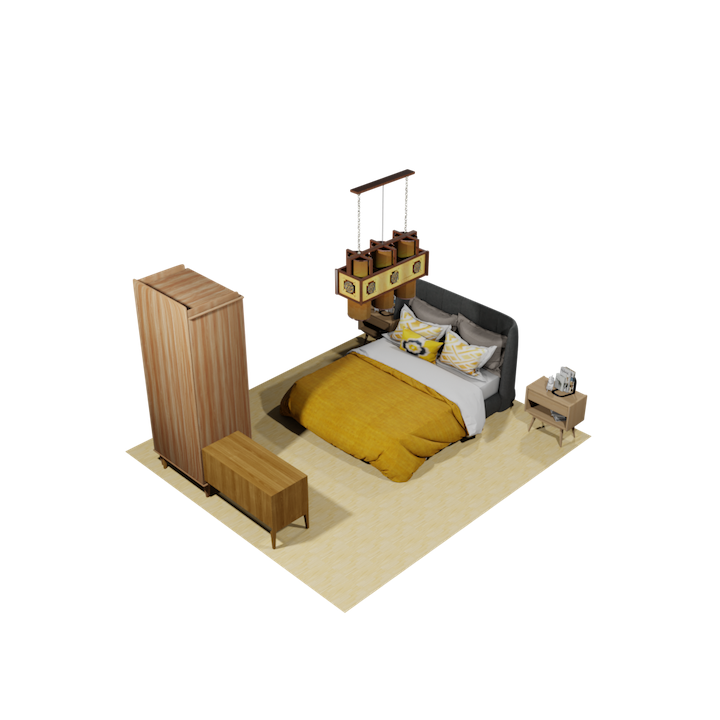}}
    \end{minipage}\hfill
    \begin{minipage}{0.195\textwidth}
        \centering
        \adj{\includegraphics[width=\textwidth]{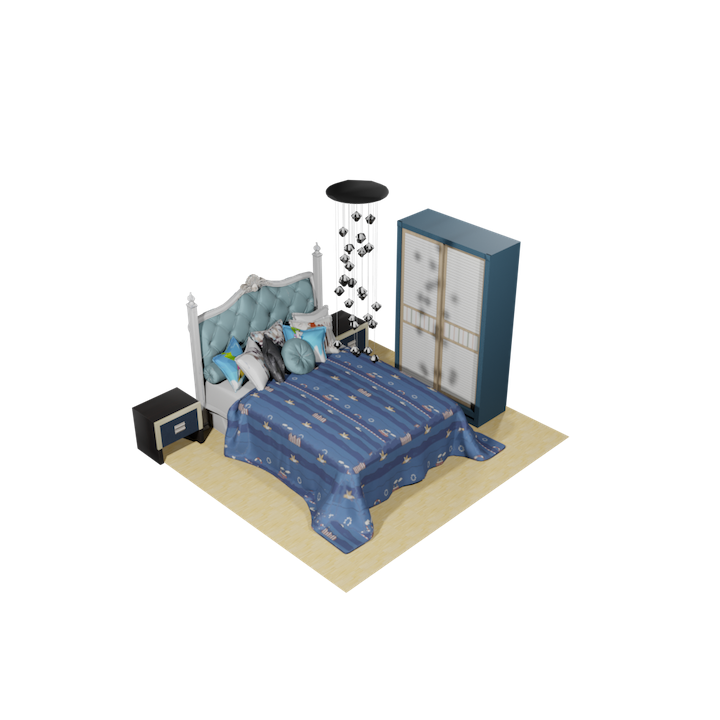}}
    \end{minipage}\hfill
    \begin{minipage}{0.195\textwidth}
        \centering
        \adj{\includegraphics[width=\textwidth]{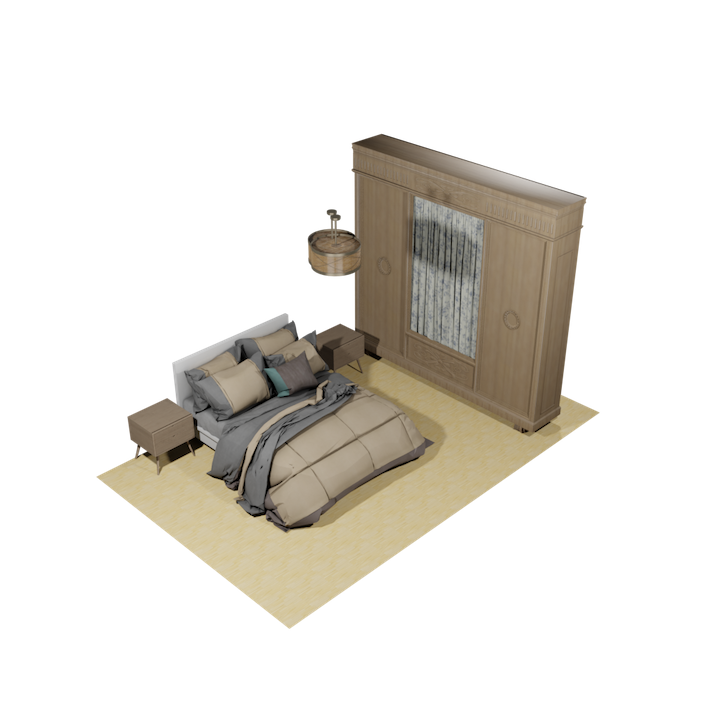}}
    \end{minipage}
    \centering{\raisebox{-15pt}{\rule{0.9\linewidth}{0.4pt}}}

    \rotatebox[origin=c]{90}{Text-guided by Users}
    \begin{minipage}{0.19\textwidth}
        \centering
        \adj{\includegraphics[width=\textwidth]{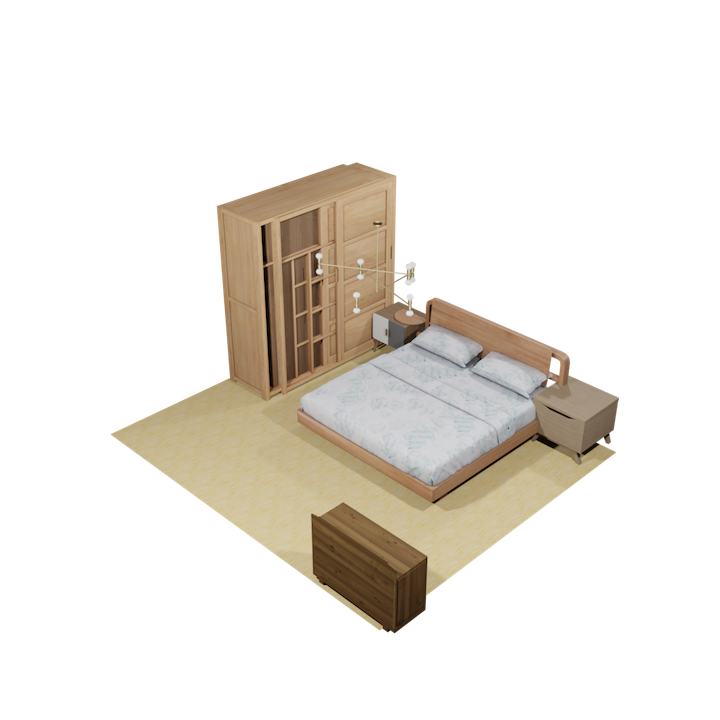}}
        Light Wooden
    \end{minipage}\hfill
    \begin{minipage}{0.19\textwidth}
        \centering
        \adj{\includegraphics[width=\textwidth]{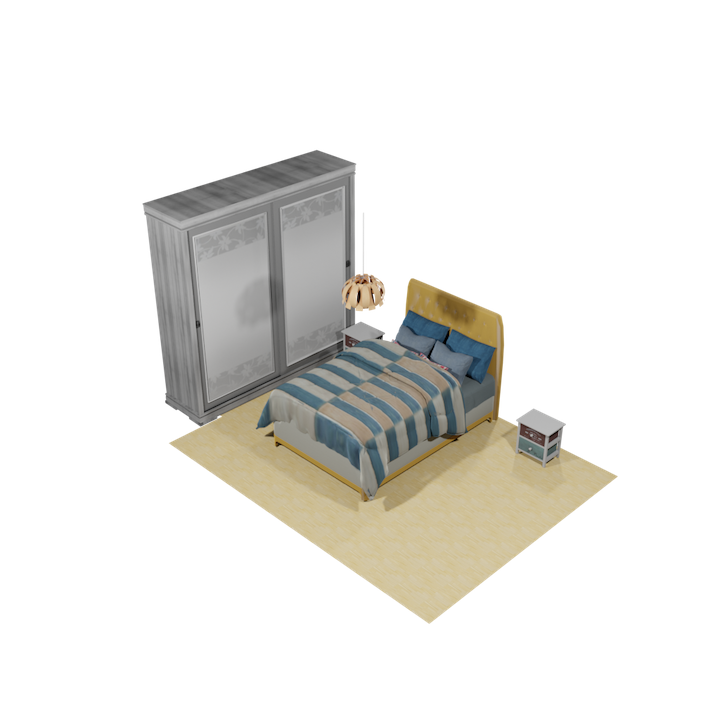}}
        Blue Striped
    \end{minipage}\hfill
    \begin{minipage}{0.19\textwidth}
        \centering
        \adj{\includegraphics[width=\textwidth]{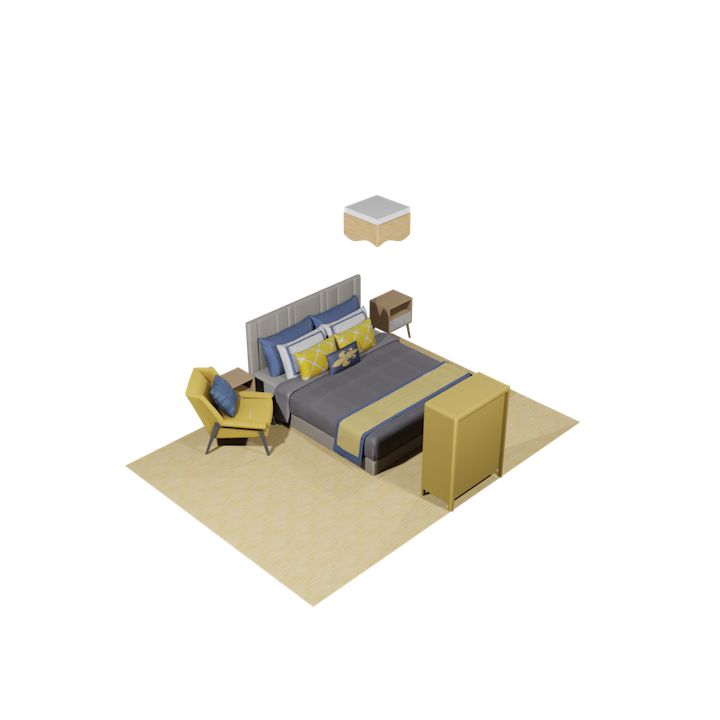}}
        Dark Yellow
    \end{minipage}\hfill
    \begin{minipage}{0.19\textwidth}
        \centering
        \adj{\includegraphics[width=\textwidth]{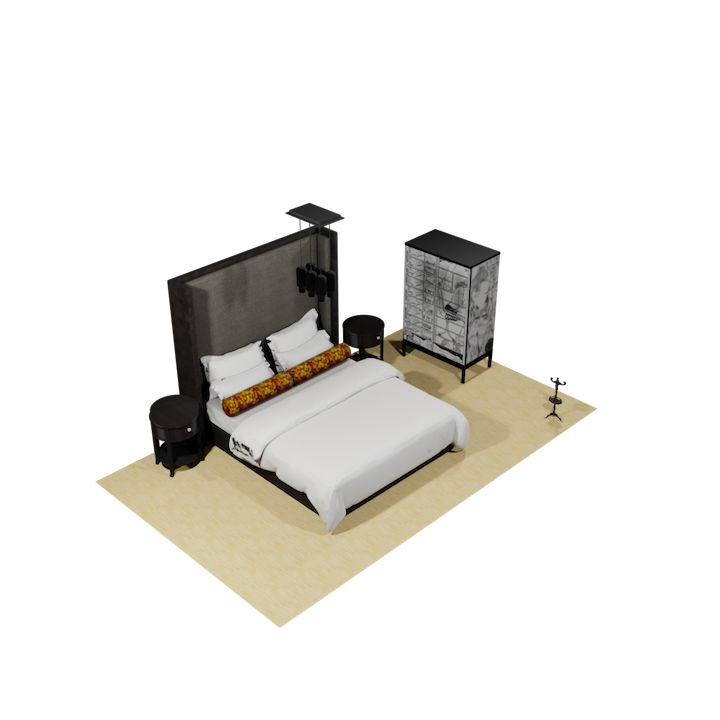}}
        European
    \end{minipage}\hfill
    \begin{minipage}{0.19\textwidth}
        \centering
        \adj{\includegraphics[width=\textwidth]{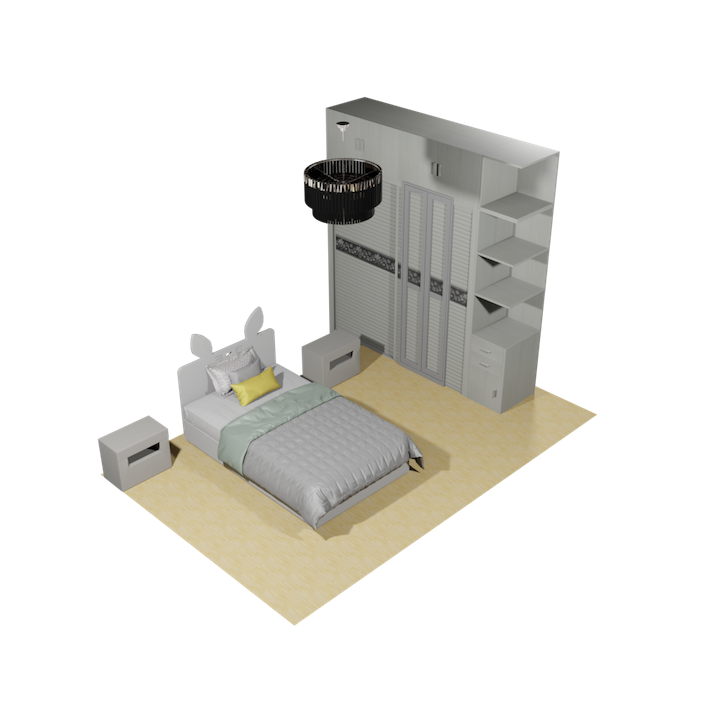}}
        Space Silver
    \end{minipage}

    \captionof{figure}{\label{fig:title} \textbf{Automatic scene synthesis:} With a floor plan and an optional style description prompt as conditions, our model can synthesize rooms with style-consistency, diversity, and visual realism.} 
    \vspace{-5pt}
\bigskip}%
\makeatother

\maketitle

\begin{abstract}
Indoor scene synthesis involves automatically picking and placing furniture appropriately on a floor plan, so that the scene looks realistic and is functionally plausible. Such scenes can serve as homes for immersive 3D experiences, or be used to train embodied agents. Existing methods for this task rely on labeled categories of furniture, e.g. bed, chair or table, to generate contextually relevant combinations of furniture. Whether heuristic or learned, these methods ignore instance-level visual attributes of objects, and as a result may produce visually less coherent scenes.

In this paper, we introduce an auto-regressive scene model\footnote{Code will be released upon acceptance.} which can output instance-level predictions, using general purpose image embedding based on CLIP. This allows us to learn visual correspondences such as matching color and style, and produce more functionally plausible and aesthetically pleasing scenes. Evaluated on the 3D-FRONT dataset, our model achieves SOTA results in scene synthesis and improves auto-completion metrics by over 50\%. Moreover, our embedding-based approach enables zero-shot text-guided scene synthesis and editing, which easily generalizes to furniture not seen during training.
\end{abstract}

\section{Introduction}
Automatically generating 3D models of indoor spaces such as rooms, houses and buildings has many important applications \cite{survey}. These include creating virtual environments to enable immersive experiences in the metaverse, generating virtual test beds to train and evaluate embodied agents, and aiding the design of physical spaces in the real world. A well-designed indoor scene synthesis model needs to produce scenes which are:
\begin{enumerate}
\item \emph{Physically and functionally plausible}. Furniture should be placed appropriately, such as beds on the floor rather than floating in the air, fridges against the wall instead of in the middle of a room etc. The composition of furniture should make sense given the type of room, e.g. not placing a bed in a bathroom nor having multiple floor lamps side by side.
\item \emph{Visually consistent and aesthetically pleasing}. Humans pay attention to more than just functional properties when composing scenes, such as whether two pieces of furniture match in style, the contrast between the colors etc. The model should be able to capture such properties.  
\item \emph{Diverse and generalizable}. Real world interiors can vary greatly in type and complexity, from a simple bedroom in a typical house to an open office space. The number of unique objects which may appear in such spaces is nearly infinite. A good model needs to be able to capture this diversity, and generalize well to new furniture libraries not seen at training time.
\end{enumerate}

Previous works either use procedural modeling techniques \cite{obj_associations, constraint_based, proc_arrange, interactive_layout, meta_sim, meta_sim_1, sdr, dr, procthor} for synthesis, or treat the problem as scene graph generation \cite{metroplis_0, metroplis_1, grains, planit, scene_graph_net, e2e, sg_vae, hybrid, fast_synth}. Both depend on manually encoded rules and heuristics to ensure plausibility. While these methods are efficient, the diversity of generated scenes are ultimately limited by the finite set of hand crafted rules. These methods could capture some aesthetic constraints if these were explicitly encoded, however only if the needed attributes are in the object metadata, and to a limited degree.

Another recent bucket of works (which ours falls into) propose learning furniture selection and placement from data \cite{deep_synth_1, deep_synth_2, sceneformer, atiss}. This approach has the advantage of being more generalizable as we increase the training data. However, previous works model the scene distribution until the level of object categories, limiting their capacity to learn about unique furniture attributes such as color and style. Such models can learn e.g. to place a chair next to a table, however they can not recommend a particular chair. This is in contrast to how humans design interiors, where lots of efforts are spent deciding between variations of similar furniture, making sure that styles match and the composition is appealing to the eye.

To capture such subtle visual correspondences, we need to go beyond simple category labels to richer visual representations of objects. To this end, we use a pooled CLIP \cite{clip} image embedding from several renders of the object under multiple view directions. Using an image encoder which was pre-trained on large scale image data from the web allows us to represent the full diversity of real world objects and styles. We pick the CLIP model in particular, because it also opens up text-guided editing capabilities. We train an auto-regressive model, \ours{}, which at each step outputs a predicted furniture embedding and its placement. During inference, we use the predicted embedding to retrieve the next furniture conditioned on what already exists in the scene, with optional style description text guidance from the user. 

\ours{} can learn to generate natural looking, diverse and style-consistent scenes (Figure \ref{fig:title}) from data. It can easily generalize to new furniture without re-training, and opens up text-guided scene synthesis applications (see Section \ref{sec:text_guided}). Modeling objects at the instance level also gives \ours{} more representation capacity, resulting in more accurate scene synthesis. When evaluated on the 3D-FRONT dataset \cite {3dfront}, we improve auto-completion precision and recall substantially over previous state-of-the-art, with improvements ranging from 22\% - 129\% relatively.

\section{Related Work}
\subsection{Scene Synthesis}
The goal of indoor scene synthesis is to generate a furniture layout that satisfies both functional and aesthetic criteria \cite{survey}. Early attempts \cite{obj_associations, constraint_based, proc_arrange, interactive_layout} primarily rely on hand-crafted rules and constraints for object placement. Following this, many techniques such as procedural modeling \cite{dr, sdr, procthor} and probabilistic grammars \cite{meta_sim, meta_sim_1} are explored, which can generate very impressive samples but might ignore visual coherence during composition. Rather than placing explicit objects in a space, \cite{scenedreamer, gaudi} attempt to represent scenes using implicit functions to model occupancy and attributes of 3D points in the space. \cite{text2room} uses text-to-image models and continuously aligning and fusing the images into rooms. Despite being realistic, implicit methods might be less ideal for downstream tasks which require reasoning of individual objects or involve physical interactions with the scenes. Another branch of work \cite{metroplis_0, metroplis_1, grains, planit, scene_graph_net, e2e, fast_synth, sg_vae, scene_gen} focus on designing specialized architectures and using graphs to capture the underlying structure of scenes. Recently, many works start to consider aspects of scene synthesis beyond predicting furniture categories and poses. \cite{human_0, human_1} aim to incorporate annotated human motions, \cite{mutual} tackles the problem of merging several input rooms into a new one with maximum mutual functional spaces, \cite{gsacnet} proposes a model that can perform well with limited scene data. Inspired by prior data-driven efforts which use CNN and transformer-based architectures for auto-regressive generation \cite{deep_synth_1, deep_synth_2, sceneformer, atiss}, we also want to additionally make our model take into account visual cues of furniture and produce style-consistent scenes. Similar to our direction, \cite{hybrid} uses a pre-trained volumetric module to encode only un-textured 3D point clouds. But \ours{} works with any renderable 3D shapes, and requires only furniture models and their transformations as labels. Built upon \cite{sceneformer, atiss}, our model uses permutation-invariant transformers \cite{atiss, attention} to generate a set of furniture predictions consisting of semantic embeddings and poses.

\subsection{Multi-modal Embedding}
Being able to encode 3D shapes with semantics and visual information is key to constructing aesthetics-aware scenes. There are many efforts towards 3D shape representations such as using irregular latent grids \cite{3dilg}, triplane features \cite{get3d}, and neural fields \cite{neural_fields}. However, they focus more on topological than semantic information and require a large amount of 3D or multi-view data for training. In the 2D domain, CLIP \cite{clip} learns a joint embedding space of texts and images from hundreds of millions of image-text pairs. CLIP consists of a text and an image encoder, which are optimized with a constrastive loss to enforce that images and texts with similar underlying semantics will be encoded to feature vectors closer in the embedding space. With this large-scale pretrained and multi-modal CLIP encoder, we can delegate the task of textured shape embedding from 3D to 2D, generalize to unseen furniture instances, and leverage the learned prior knowledge of various visual patterns to accomplish text-driven applications. 

\section{Method}
\subsection{Problem Setup}
Let $X = {X_1, ..., X_2}$ denote a collection of scenes where each scene $X_i = (O_i, F_i)$ comprises the unordered set of objects $O_i = \{o^i_j\}^M_{j=1}$ and its floor layout $F_i$. Our goal is to model the distribution over scenes $p_\theta(X)$, which can be written as:
\begin{align}
    p_\theta(X_i) = p(F_i)p_\theta(O_i|F_i) = p(F_i)\sum_{O\in\pi(O_i)}\prod_{j\in O}p_\theta(o^i_j|o^i_{<j}, F_i)
\end{align}
where $p(F_i)$ is the given prior over scene floor plans and $p_\theta(o^i_j|o^i_{<j}, F_i)$ is the probability of the $j$-th object, conditioned on the floor plan and all the objects in the $i$-th partial scene, and $\pi(\cdot)$ is the permutation operator. The objective is to then maximize the log-likelihood of this function. Following \cite{atiss}, we switch the summation to a product, which gives the objective function:
\begin{align}\label{eq:obj}
    \log p_\theta(X_i) = \log (\prod_{O\in\pi(O_i)}\prod_{j\in O}p_\theta(o^i_j|o^i_{<j}, F_i)) = \sum_{O\in\pi(O_i)}\sum_{j\in O}\log p_\theta(o^i_j|o^i_{<j}, F_i)
\end{align}
Changing from summation to product makes it more computationally feasible and could encourage all permutations to have higher density\cite{atiss}. In our implementation, each furniture instance has a semantic embedding and a 3D transformation consisting of translations, sizes, and an angle around the vertical axis. The floor plan takes the form of binary masks, indicating the boundary of furniture placement. 

\subsection{Architecture}


Our model primarily consists of three parts: 1) encoders for the room floor plan, the transformation, and the embedding in Section \ref{sec:encoder} 2) a transformer decoder for the set of all encoded instances in the scene in Section \ref{sec:transformer} and 3) decoders for the next instance at each timestamp in Section \ref{sec:decoder}. 


\subsubsection{Floor Plan, Transformation and Embedding Encoder} \label{sec:encoder}
Our synthesis conditions on a given floor plan, which is encoded as a binary mask on a square grid. This mask image is then fed into a pretrained ResNet-18 feature extractor \cite{resnet}, the final layer of which is used as the feature vector $F$ for the floor plans. 

Each furniture instance has a 7D tensor representing its transformation in the 3D space: 3D transformation, 3D size, and 1D rotation along the vertical axis. We choose to model only the rotation along the vertical axis because all furniture in our dataset are only rotated like this, and it is straightforward to generalize to 3D rotation if needed. All transformations are normalized linearly into $[0, 1]$ based on training dataset statistics with the consideration of possible scene augmentations discussed in Section \ref{sec:training}. We follow the fixed sine-cosine positional encoding scheme $\gamma(\cdot)$ for each dimension of the attributes as in \cite{attention}. A furniture instance also has a semantic embedding vector $h$ (Section \ref{sec:emb}), which we project to a smaller space with a MLP. The encoded attributes are concatenated into a single feature vector, projected to 512 dimension linearly as $C_j$, which represents the feature of the $j$-th transformed object instance, and will then be fed into the transformer decoder.

\subsubsection{Transformer Decoder} \label{sec:transformer}
Following \cite{atiss}, we use a decoder-only transformer \cite{attention, fast_trans_0, fast_trans_1} without any positional encoding because we want the model to ignore the order of the given set of furniture, hence producing permutation-invariant features. The input set of the transformer is $I = \{F\} \cup \{C_j\}_{j<i} \cup q$ for predicting the $i$-th instance where $q \in\mathbf{R}^{64}$ here is a learned query vector\cite{atiss}.

\subsubsection{Instance Decoder} \label{sec:decoder}
At each step, we first predict a stop token to decide whether the model should stop adding more furniture or not. This is implemented as a simple MLP head with two logits as the output. The output is then appended to $\hat{q}$ to predict the next object instance by feeding into the embedding decoder, which outputs a vector in the same dimension as the raw object embedding. From this, we project into the output space of all possible furniture in the dataset. We output a distribution over this vocabulary by taking the softmax over the dot products of the predicted raw embedding $\hat{h}$ with all furniture embeddings. During instance retrieval, we apply temperature scaling and top-K sampling.

Once decided on the stop condition and instance embedding, the model will output translation, rotation angle, and sizes accordingly, in a cascading fashion. We choose to model all of the transformation attributes as mixture of logistics distributions as in \cite{pixelcnn_pp, atiss} since we experimentally found them to out-perform other design choices such as mixtures of Gaussians. 

\subsubsection{Object Embedding} \label{sec:emb}
Learning representations for 3D objects has been studied in many recent work \cite{clip_forge, dreamfusion, get3d, lion}. Despite being very effective for their aimed application, these approaches require large amounts of 3D data to train, which might not be available at the scale we need. Also these representations are geared towards mesh surface reconstruction, which is not necessarily a good fit for our application. In order to capture the wide variety of objects and styles which may appear in real indoor scenes, we propose to utilize an existing pre-trained image representation, which has been trained on large scale image data from the web. In particular, we pick the multi-modal CLIP model \cite{clip}, for the additional advantage that it comes with aligned text embeddings, opening up multi-modal applications such as text guided synthesis and editing.
\begin{figure}
    \begin{center}
    \includegraphics[width=0.35\textwidth]{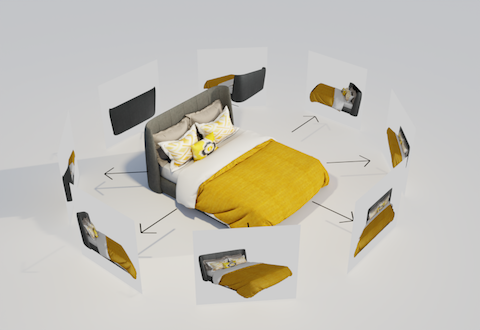}
    \end{center}
\caption{\label{fig:eight_view} \textbf{Object embedding:} \ours{} calculates the semantic embedding of a 3D mesh by feeding images rendered from eight canonical directions to the CLIP image encoder.}
\vspace{-5pt}
\end{figure}

In order to use an off-the-shelf 2D image representation to encode a 3D object, we rely on renderings of the object from multiple view points, and then pooling the embeddings from this set of renders. For each object instance, we first normalize it to make sure it lies within the centered unit cube with correct forward and upward directions. We then manually set the diffuse, ambient, and specular properties to a fixed value for consistency. After that, we render images of this normalized shape from eight canonical camera directions (azimuth) on the upper hemisphere with a fixed camera elevation and distance, the same lighting, and shading setting (details in Appendix \ref{app:rendering}) as illustrated in Figure \ref{fig:eight_view}. The resulting set of eight images $\{I_i\}_8$ will be fed into a pretrained CLIP image encoder\cite{clip}, which takes in a RGB image and outputs a tensor in $\mathbf{R}^{512}$. The final embedding will be the normalized average of the eight individual ones, $h = \frac{1}{8}\sum^8_{i=1} normalized(\text{CLIP}(I_i))$. We find this encoding method to work well for image or text-based object retrieval, which serves as a strong building block for our style-aware scene generator.

Using a pre-trained image encoder allows us to represent fine visual details of each furniture instance while remain agnostic to object format. As we do not rely on category labels any more, we also relax the data requirement of category annotation for each furniture.


\subsection{Training and Inference} \label{sec:training}
To optimize the objective in \eqref{eq:obj} during training, for each scene in the dataset, we first take a subset of a randomly chosen size $T$ of objects from the scene and randomly permute them. Then the subset of objects along with the floor plan will be encoded using their respective encoders. Conditioned on the encoded features, our network first predicts whether to add any new objects, and then decodes the embedding and transformation. We use standard binary cross entropy loss with label smoothing for the stop token. For the embedding, we first take the dot product of the predicted embedding with all instance embedding in the furniture space and then use cross entropy loss. Finally, we use the analytical negative log-likelihood for the transformations as they're modeled as mixtures of logistics.

To enhance the quality and diversity of our training data, we attempted scene rotations with random angles, mirroring along the $x$ or $z$ axis, and instance transformation jittering, where we find the last one to be less effective and therefore include the combination of the first two in our experiments.

During inference, we provide the network with a floor plan sampled from a given prior distribution, with an optional additional set of initial furniture as in the scene completion task. When a new object is chosen from the candidate pool, the corresponding ground truth embedding will be fed back to the network as the input, together with the previous step outputs for predicting other attributes. And finally, we transform the furniture with predicted pose and add it to the scene.

\section{Experiments}
In this section, we evaluate \ours{} in partial scene completion, unconditional synthesis, text-guided synthesis and editing. We compare to ATISS \cite{atiss} as our main baseline and use their released codebase to retrain their model on our version of the dataset for fair comparison. 

\subsection{Dataset}
We use 3D-FRONT\cite{3dfront} as our dataset for all experimental settings, which consists of human-designed indoor scenes with annotations for each furniture object from the complementary 3D-FUTURE dataset \cite{3dfuture}. For each scene, we require only annotations for the transformation and the raw mesh model of each furniture. To facilitate learning and fair comparison, we followed the same dataset filtering procedures used in \cite{atiss} which, for example, removes rooms with unnatural dimensions, crowded furniture, and ill-posed shapes. Unfortunately, after filtering, the new version of this dataset contains less data than the original which is no longer available\footnote{https://github.com/nv-tlabs/ATISS/issues/18} (dataset statistics summarized in Table \ref{tab:dataset}).   

\begin{table}
  \caption{Dataset statistics after filtering}
  \label{tab:dataset}
  \centering
  \begin{tabular}{llllll}
    \toprule
    \multicolumn{1}{c}{} & \multicolumn{1}{c}{Old dataset} & \multicolumn{4}{c}{New dataset} \\
    \midrule
    Room type & Total & Train & Test & Total & Furniture \\
    \midrule
    Bedroom & 5996 & 3879 & 162 & 4041 & 2398 \\
    Dining Room & 2625 & 723 & 177 & 900 & 2247 \\
    Library & 622 & 230 & 56 & 286 & 464 \\
    Living Room & 2962 & 621 & 192 & 813 & 2380 \\
    \bottomrule
  \end{tabular}
\end{table}

\subsection{Scene Synthesis from Floor plans}
We first evaluate the performance of our scene model for its ability to generate plausible and diverse scenes conditioned on the floor plan, with optional existing furniture in a partial scene. 

\subsubsection{Quantitative Evaluations} \label{sec:scene_comp}

\begin{figure}
    \captionsetup[subfigure]{labelformat=empty}
    \centering
    \begin{subfigure}[b]{0.25\textwidth}
        \centering
        \adj{\includegraphics[width=\textwidth]{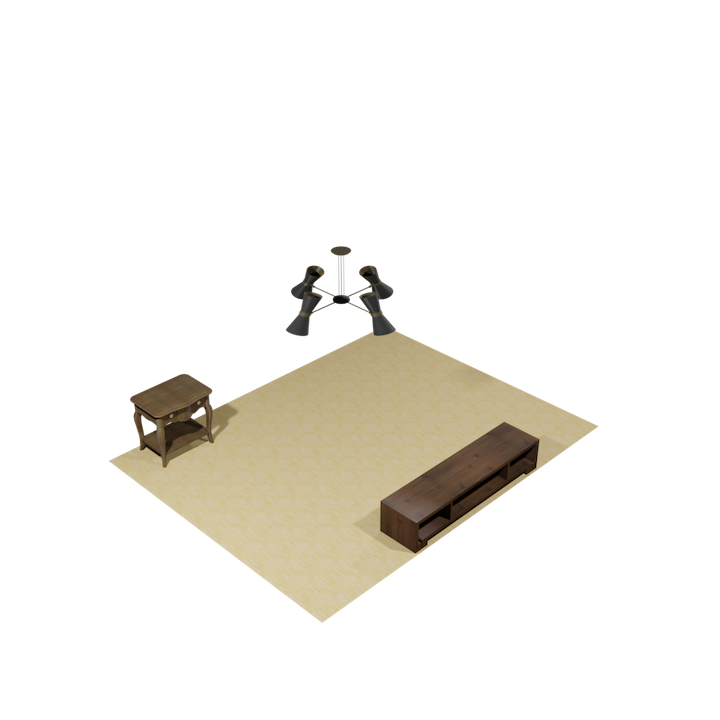}}
    \end{subfigure}
    \begin{subfigure}[b]{0.25\textwidth}
        \centering
        \adj{\includegraphics[width=\textwidth]{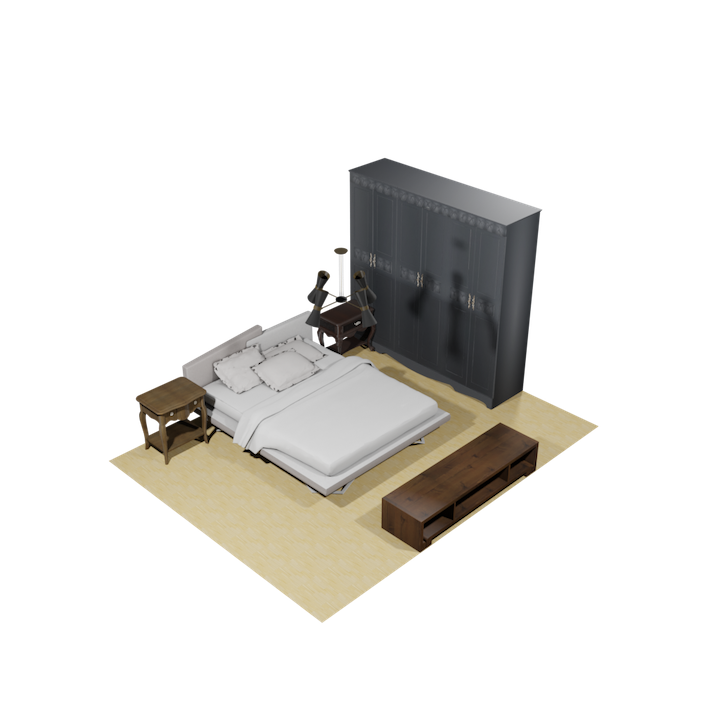}}
    \end{subfigure}
    \begin{subfigure}[b]{0.25\textwidth}
        \centering
        \adj{\includegraphics[width=\textwidth]{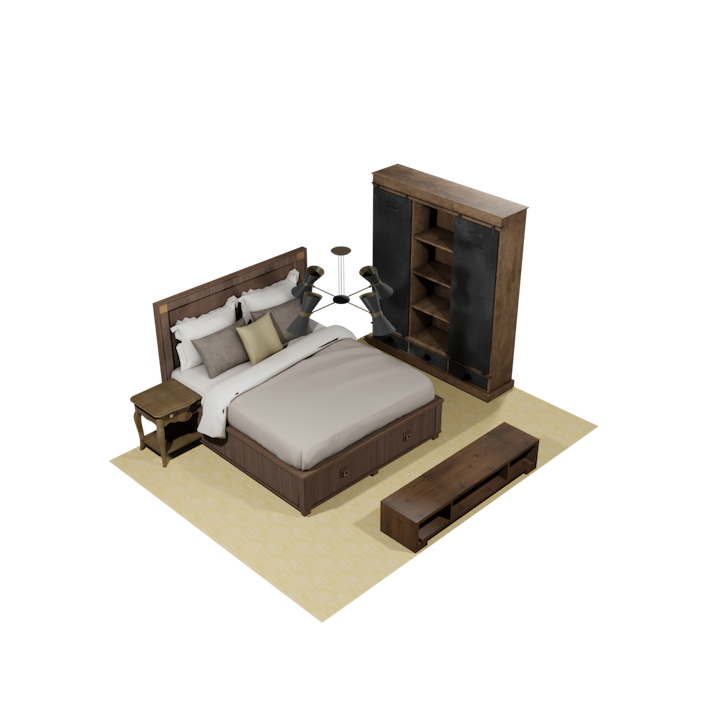}}
    \end{subfigure}
    \vspace{-10pt}

    \begin{subfigure}[b]{0.25\textwidth}
        \centering
        \adj{\includegraphics[width=\textwidth]{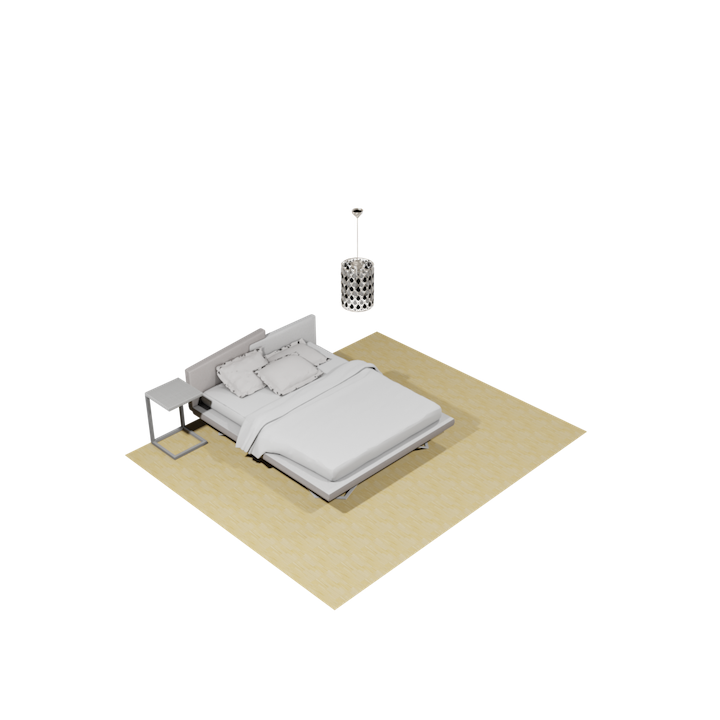}}
        \caption{Partial}
    \end{subfigure}
    \begin{subfigure}[b]{0.25\textwidth}
        \centering
        \adj{\includegraphics[width=\textwidth]{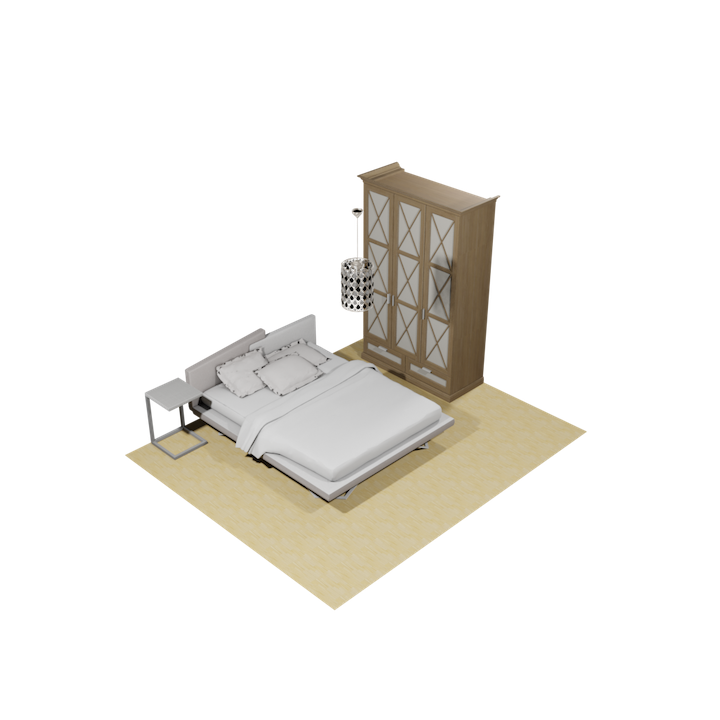}}
        \caption{ATISS}
    \end{subfigure}
    \begin{subfigure}[b]{0.25\textwidth}
        \centering
        \adj{\includegraphics[width=\textwidth]{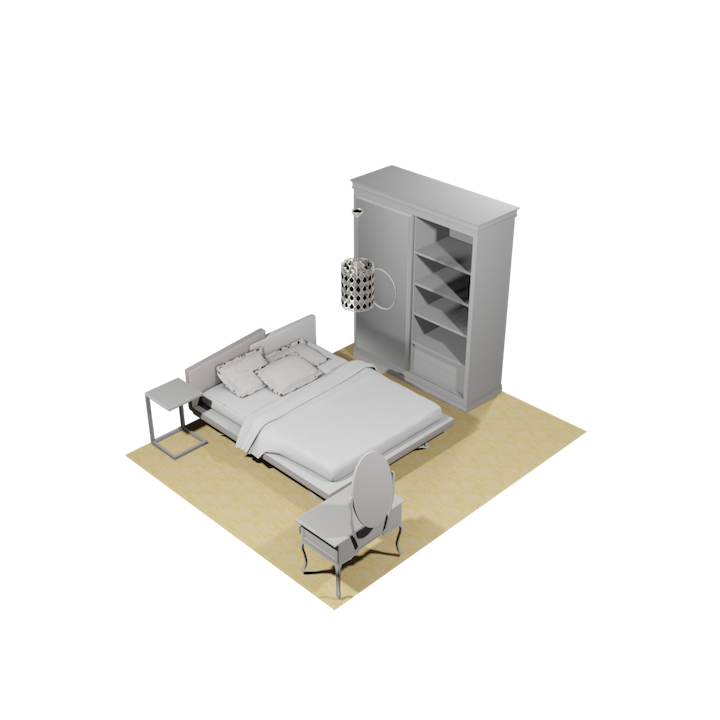}}
        \caption{Ours}
    \end{subfigure}
    
    \caption{\label{fig:scene_comp} \textbf{Partial scene completion:} From the left to right columns are the partial scenes, the completed scenes from ATISS, and the ones generated by our model. \ours{} can distinguish not only color information, but also furniture material and shapes.}
\end{figure}

To provide quantitative results on the performance, we aim to compare how closely our model can complete a partial scene compared to the ground truth test set. For each test scene, we randomly remove $N$ objects and let the model complete it. We focus on two metrics:

\begin{enumerate}
    \item \emph{Retrieval metrics:} Models complete the partial scene by picking and placing $M$ new objects from the asset library. To measure how well the model can retrieve the ground-truth objects, we report the precision and recall of the $M$ retrieved objects with respect to the set of $N$ ground truth furniture.  
    \item \emph{Distance between scene distributions:} For a 3D scene, we used similar setup as in Section \ref{sec:emb} to render eight images from it as the visual representation. We then compare the set of images from the ground truth test scenes to those from the auto-completed scenes using FID and CLIP-FID scores \cite{clean_fid}. Inception distance scores based on a pre-trained visual representation can in principle capture the overall quality of the scenes, including natural placement and cohesiveness. We believe in-perspective renders can better capture the visual details of the scene, whereas top-down images used in prior works would only be informative for layout. We render the images in $512\times512$ resolution for better visual distinctiveness. 
\end{enumerate}

In Table \ref{table:nfurniture_bedroom} we summarize scene completion results for the "bedroom" type. We remove furniture until there is anywhere between 1-5 instances left in the scene. We present metrics separately for different numbers of existing objects in the scene. When this number is low (e.g. 1-2) the task is closer to unconditional generation, as the model has a lot more freedom in assembling a complete scene. As we leave more furniture in the scene (e.g. 4-5), the problem becomes more constrained, as there are fewer ways of consistently completing the scene. We see that \ours{} performs well in most cases, improving precision by an average 53\% and recall by 34\%.  

Looking at FID scores, we see that \ours{} does better compared to the baseline in most cases, and the improvements get larger as the number of pre-populated furniture increases. For the cases where the scene is largely incomplete, in particular the 1-furniture setting, our method lags the baseline in FID scores. We hypothesize that our model might be over-calibrated for precision rather than recall, and is prone to stopping early in these cases. We discuss limitations and potential solutions further in Section \ref{sec:limitations}. 

The full metrics for all room types are presented in the Appendix Table \ref{table:nfurniture_full}. We show a summary table over all room types in Table \ref{table:loo_full}, where we only ask the models to complete a single piece of furniture in an incomplete scene. All quantitative results are averaged over five runs with fixed seeds. \ours{} outperforms the baseline substantially over all room types and metrics. Despite the small dataset sizes for other room types, these results are significant due to the size of the improvements.

\begin{table}
  \caption{Metrics for one-step partial scene completion}
  \label{table:loo_full}
  \centering
  \begin{tabular}{llllll}
    \toprule
    Room Type & Model & FID(↓) & CLIP-FID(↓) & Precision(↑) & Recall(↑) \\
    \midrule
    \multirow{2}{*}{Bedroom} 
    & ATISS & 8.66 & 0.33 & 25.5\% & 23.0\% \\
    & \ours{} & \textbf{7.02} & \textbf{0.25} & \textbf{43.1\%} & \textbf{36.7\%} \\
    \midrule
    \multirow{2}{*}{Dining room} 
    & ATISS & 8.87 & 0.27 & 17.5\% & 14.8\% \\
    & \ours{} & \textbf{6.36} & \textbf{0.19} & \textbf{32.2\%} & \textbf{26.9\%} \\
    \midrule
    \multirow{2}{*}{Library} 
    & ATISS & 33.25 & 1.68 & 10.9\% & 9.6\% \\
    & \ours{} & \textbf{26.23} & \textbf{1.31} & \textbf{24.8\%} & \textbf{22.1\%} \\
    \midrule
    \multirow{2}{*}{Living room} 
    & ATISS & 7.38 & 0.19 & 11.2\% & 9.6\% \\
    & \ours{} & \textbf{5.01} & \textbf{0.13} & \textbf{21.3\%} & \textbf{15.7\%} \\
    \bottomrule
  \end{tabular}
\end{table}

\begin{table}
  \caption{Metrics for partial scene completion of the bedroom type}
  \label{table:nfurniture_bedroom}
  \centering
  \begin{tabular}{llllll}
    \toprule
    Prepopulated \# & Model & FID(↓) & CLIP-FID(↓) & Precision(↑) & Recall(↑) \\
    \midrule
    \multirow{2}{*}{1 furniture} 
    & ATISS & \textbf{16.38} & \textbf{0.80} & 21.6\% & 23.0\% \\
    & \ours{} & 17.83 & 1.03 & \textbf{36.5\%} & \textbf{31.6\%} \\
    \midrule
    \multirow{2}{*}{2 furniture} 
    & ATISS & 14.87 & 0.71 & 26.3\% & 29.5\% \\
    & \ours{} & \textbf{14.08} & \textbf{0.67} & \textbf{40.7\%} & \textbf{42.1\%} \\
    \midrule
    \multirow{2}{*}{3 furniture} 
    & ATISS & 13.51 & 0.64 & 26.9\% & 31.2\% \\
    & \ours{} & \textbf{12.11} & \textbf{0.53} & \textbf{40.4\%} & \textbf{38.2\%} \\
    \midrule
    \multirow{2}{*}{4 furniture} 
    & ATISS & 12.53	& 0.63 & 26.7\% & 31.4\% \\
    & \ours{} & \textbf{10.80} & \textbf{0.46} & \textbf{38.4\%} & \textbf{41.9\%} \\
    \midrule
    \multirow{2}{*}{5 furniture} 
    & ATISS & 11.84 & 0.59 & 26.2\% & 31.5\% \\
    & \ours{} & \textbf{9.88} & \textbf{0.42} & \textbf{39.1\%} & \textbf{42.5\%} \\
    \bottomrule
  \end{tabular}
\end{table}

\subsubsection{Qualitative Results}

\begin{figure}
    \centering
    \raisebox{40pt}{\rotatebox[origin=c]{90}{ATISS}}
    \begin{subfigure}[b]{0.195\textwidth}
        \centering
        \adj{\includegraphics[width=\textwidth]{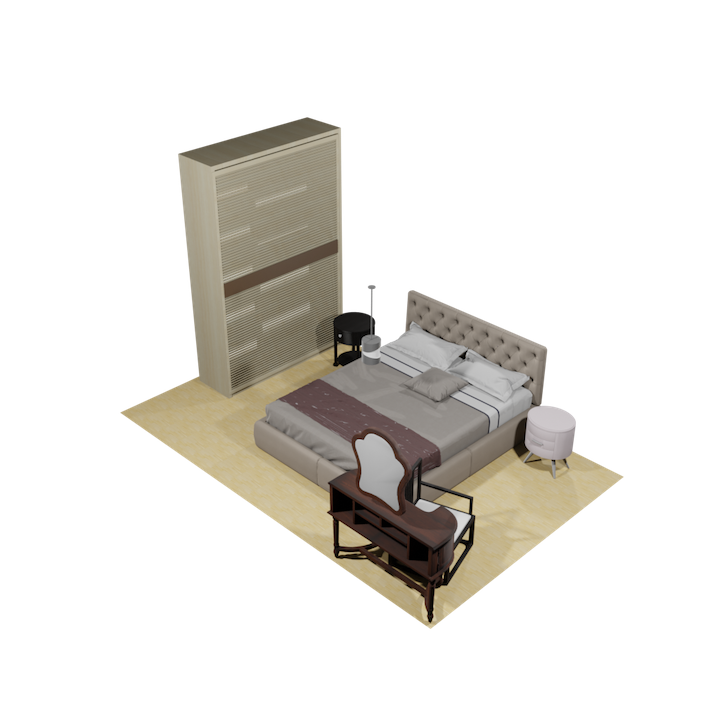}}
    \end{subfigure}\hfill
    \begin{subfigure}[b]{0.195\textwidth}
        \centering
        \adj{\includegraphics[width=\textwidth]{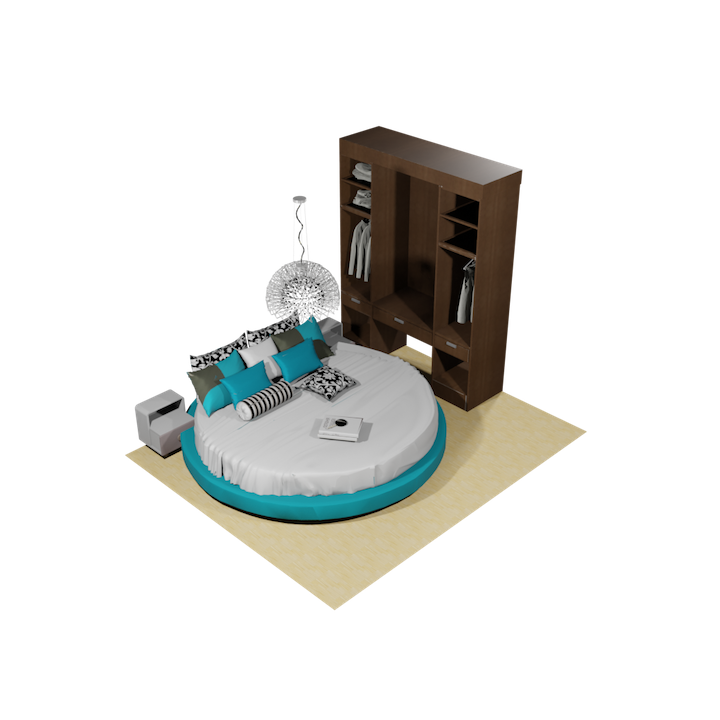}}
    \end{subfigure}\hfill
    \begin{subfigure}[b]{0.195\textwidth}
        \centering
        \adj{\includegraphics[width=\textwidth]{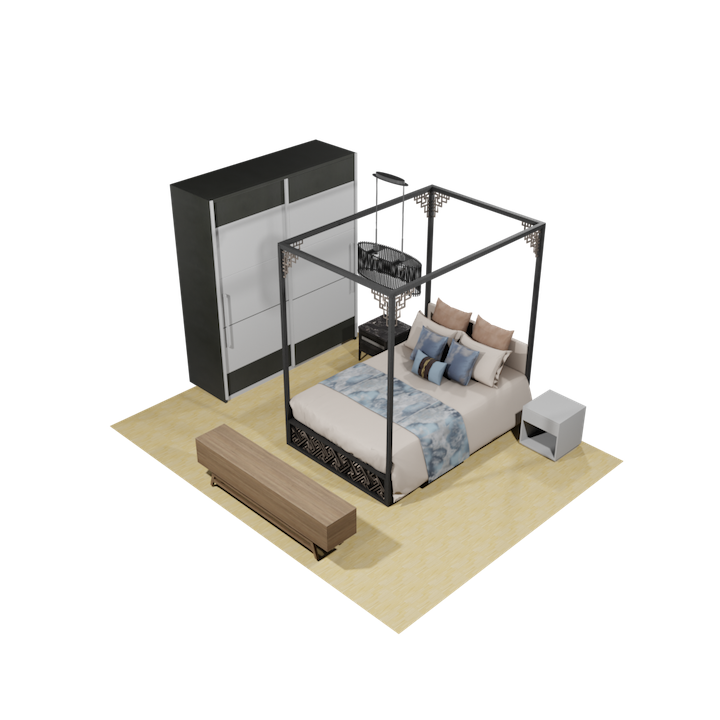}}
    \end{subfigure}\hfill
    \begin{subfigure}[b]{0.195\textwidth}
        \centering
        \adj{\includegraphics[width=\textwidth]{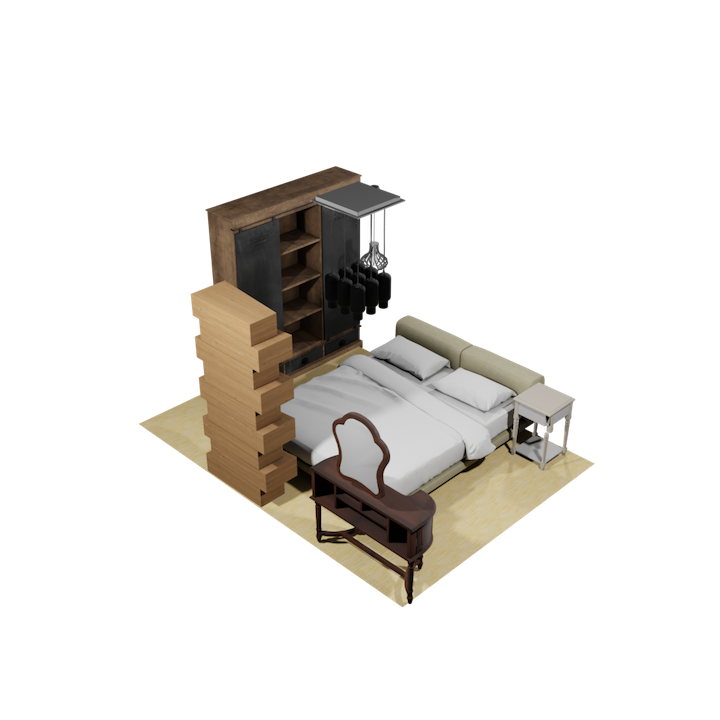}}
    \end{subfigure}\hfill
    \begin{subfigure}[b]{0.195\textwidth}
        \centering
        \adj{\includegraphics[width=\textwidth]{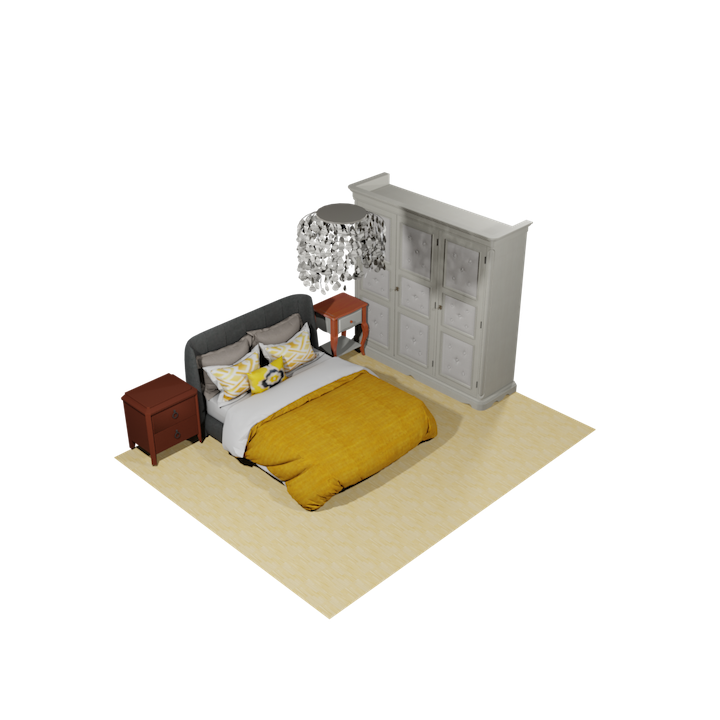}}
    \end{subfigure}\hfill

    \raisebox{40pt}{\rotatebox[origin=c]{90}{Ours}}
    \begin{subfigure}[b]{0.195\textwidth}
        \centering
        \adj{\includegraphics[width=\textwidth]{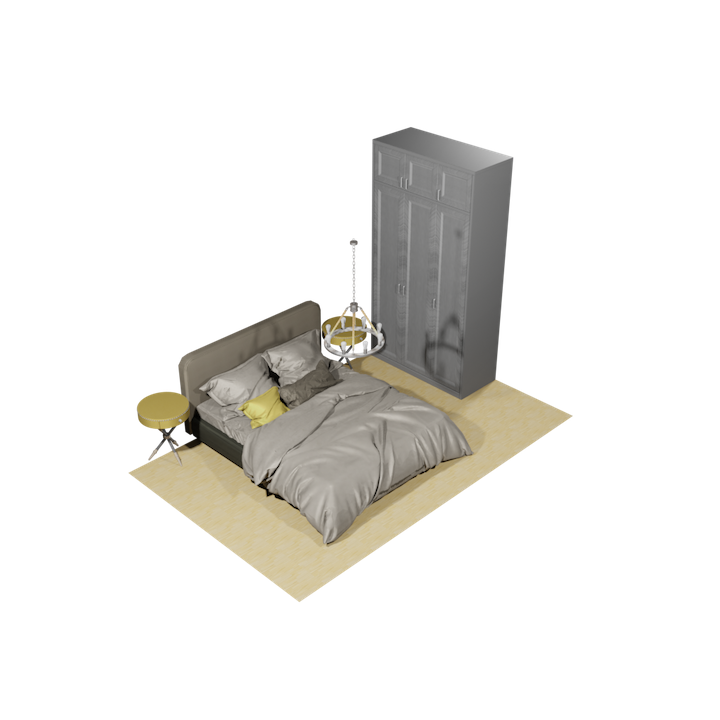}}
    \end{subfigure}\hfill
    \begin{subfigure}[b]{0.195\textwidth}
        \centering
        \adj{\includegraphics[width=\textwidth]{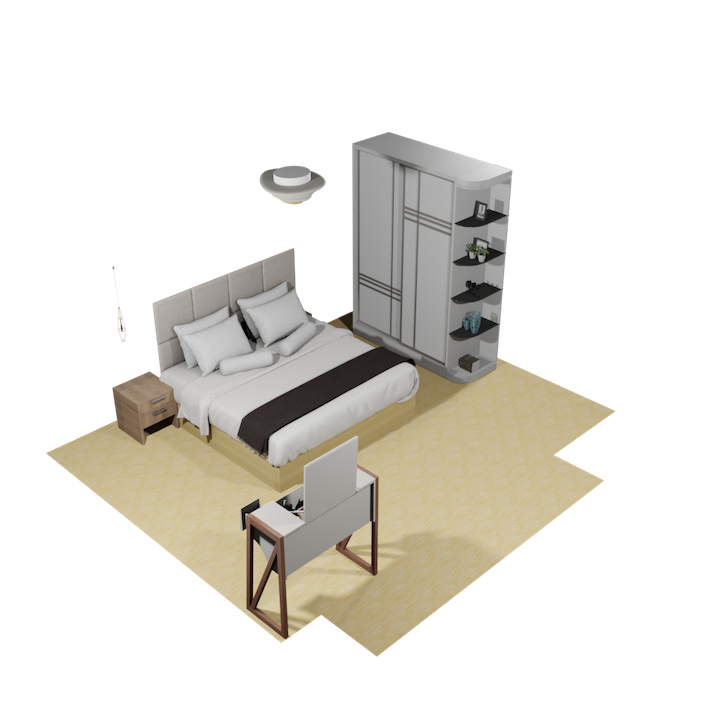}}
    \end{subfigure}\hfill
    \begin{subfigure}[b]{0.195\textwidth}
        \centering
        \adj{\includegraphics[width=\textwidth]{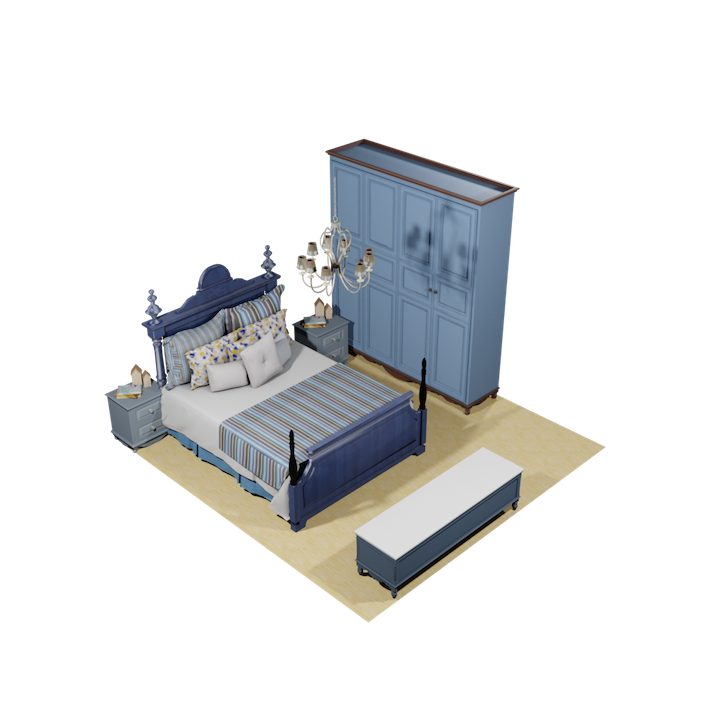}}
    \end{subfigure}\hfill
    \begin{subfigure}[b]{0.195\textwidth}
        \centering
        \adj{\includegraphics[width=\textwidth]{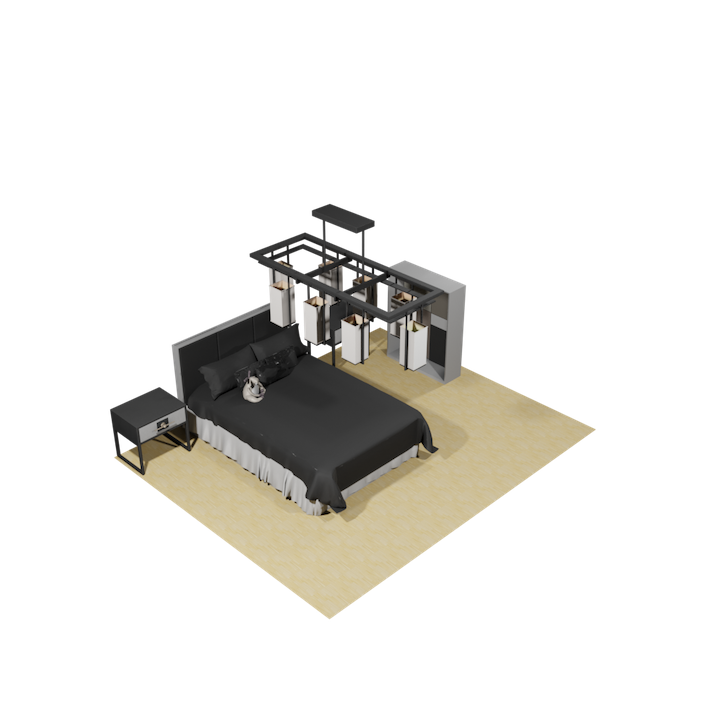}}
    \end{subfigure}\hfill
    \begin{subfigure}[b]{0.195\textwidth}
        \centering
        \adj{\includegraphics[width=\textwidth]{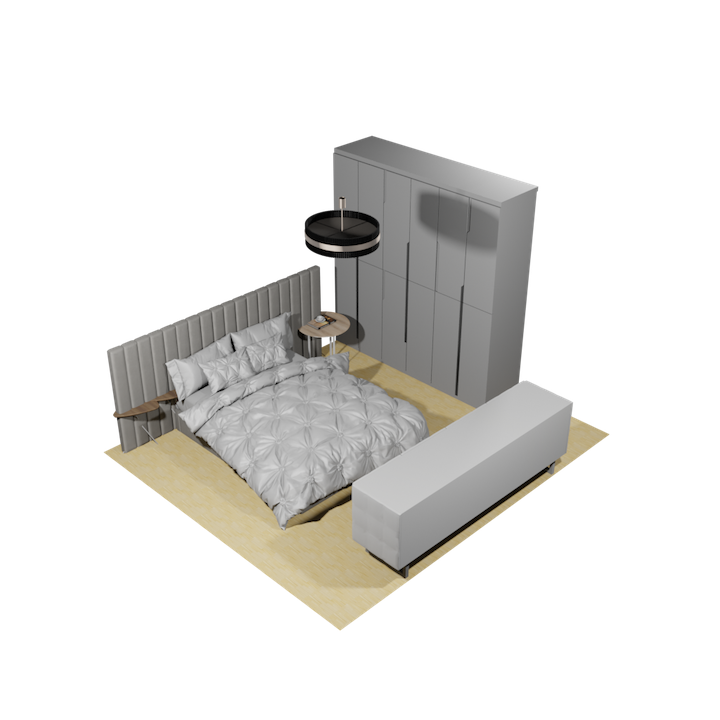}}
    \end{subfigure}\hfill
    \caption{\label{fig:uncond_gen} \textbf{Scene synthesis from room masks}: Visual examples of scenes generated from floor plans where the top row contains samples from ATISS and bottom row by \ours()}
\end{figure}

In this section, we provide some visual examples to showcase the superiority of the scenes generated by our model over the baseline. Figure \ref{fig:uncond_gen} gives a comparison between the two models for (unconditional) scene synthesis from floor plans and Figure \ref{fig:scene_comp} for partial scene completion (with more examples and a comparison against the ground truth in Appendix Figure \ref{app_fig:uncond_gen}, \ref{app_fig:scene_comp_against_baseline} and \ref{app_fig:scene_comp_against_gt}). Note that in both cases, despite being able to produce plausible layout configuration of the room, the baseline model fails to capture style information from the dataset. \ours{} can produce scenes which follow specific patterns like style and color, and therefore make the samples more artistically realistic and well-designed, such as consistently modern furniture, or a particular dominant color (more examples from \ours{} in Figure \ref{fig:title}).

\subsection{Text-guided Applications}\label{sec:text_guided}
Thanks to the furniture semantic CLIP embedding, \ours{} can naturally execute many text-conditioned tasks such as scene synthesis and text-guided furniture replacement with decent zero-shot performance. 

\subsubsection{Text-guided Scene Synthesis}

We first showcase that our model can take in user-specified style prompts and generate scenes accordingly. We achieve this by embedding the text prompt into an edit vector $t$ which is scaled to the same length by the norm of the current predicted embedding $\hat{e}$, and then merged using linear interpolation with a weight $w$ before being used for object retrieval. In our trials, we empirically find weights between 0.2 to 0.5 to work well. To avoid interfering with the style-consistency learned by our model, we apply a exponential decay to the weight factor at each timestamp. This technique has several benefits: 1) our model will take advantage of the learned prior on style-consistency and be less reliant on the prompts at later stages and 2) when the desired furniture is too far away from the data distribution, the model will not be affected too much by the edit vector and can still choose the best from available options. In Figure \ref{fig:title}, we qualitatively show the diversity and faithfulness of generated scenes with example prompts (more examples in Appendix Figure \ref{app_fig:text_gen}). \ours{} is able to both adapt to simple material or texture cues such as ``wooden" or ``striped", and entertain more abstract design prompts such as ``European". To the best of our knowledge, this is the first successful demonstration of text-guided scene synthesis with instance-level manipulation ability under the zero-shot setting.

\subsubsection{Text-guided Furniture Replacement}

\begin{figure}
    \centering
    \begin{subfigure}[t]{0.48\textwidth}
        \centering
        \adj{\includegraphics[width=0.5\textwidth]{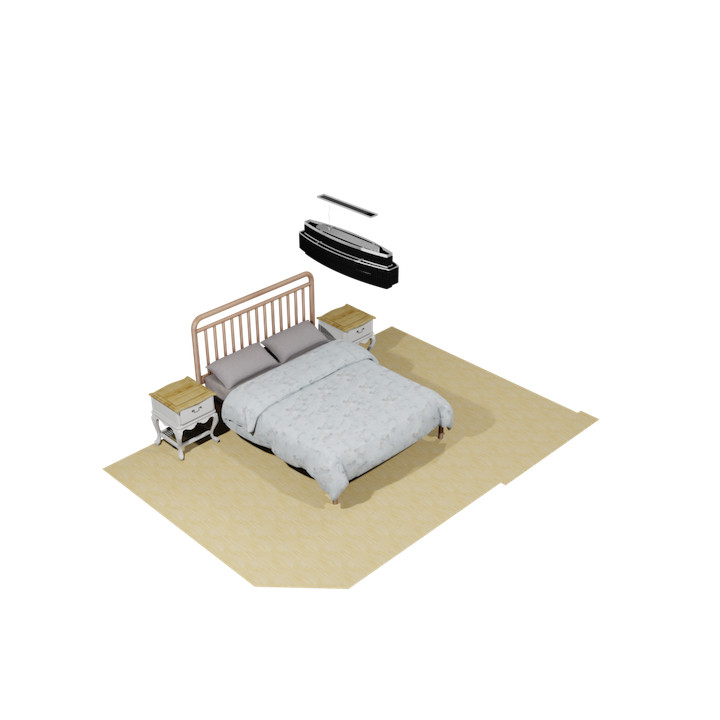}}
        \adj{\includegraphics[width=0.5\textwidth]{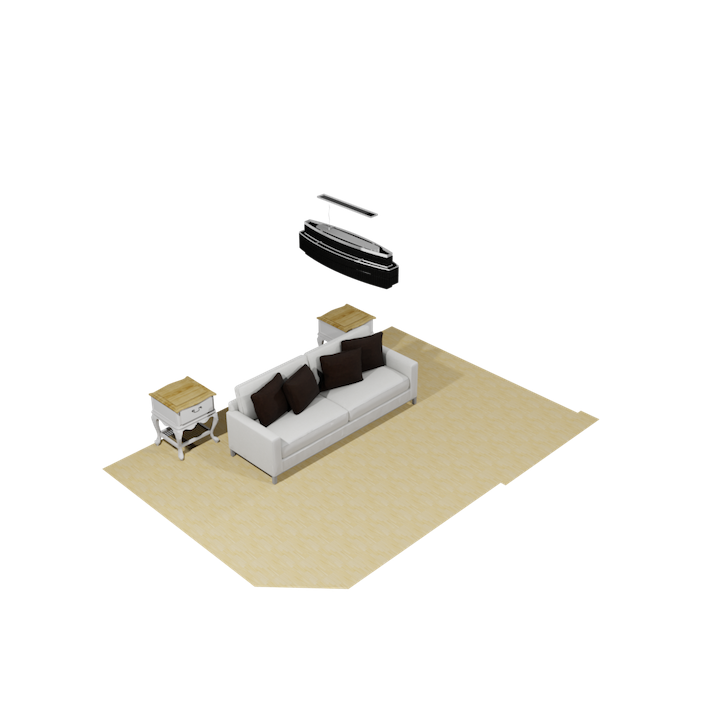}}
        \caption{From the bed to ``a black and white lounge sofa''}
    \end{subfigure}
    \begin{subfigure}[t]{0.48\textwidth}
        \centering
        \adj{\includegraphics[width=0.5\textwidth]{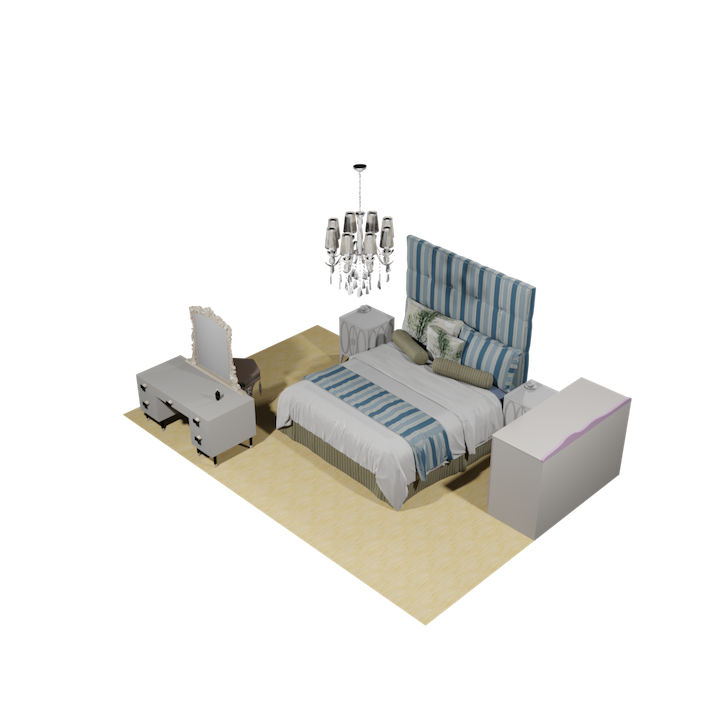}}
        \adj{\includegraphics[width=0.5\textwidth]{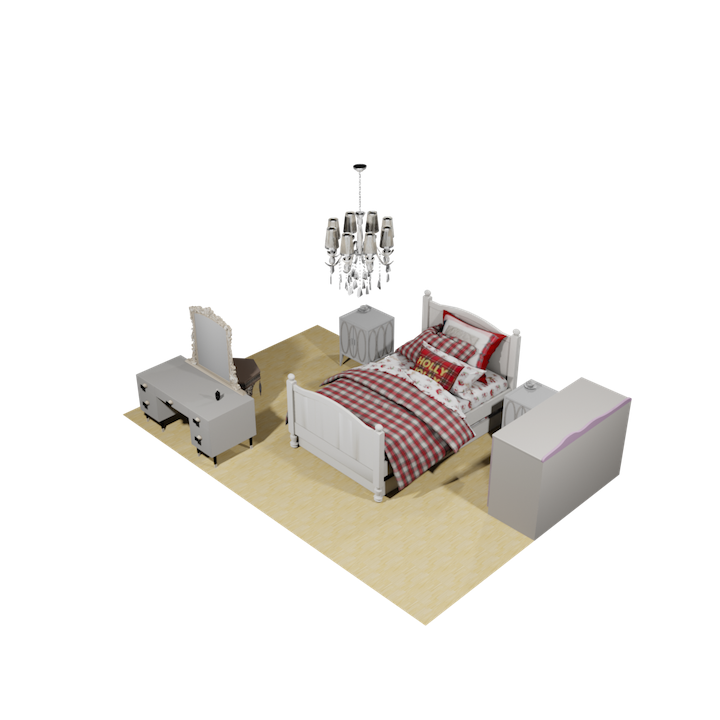}}
        \caption{From the blue bed to ``a red-striped bed''}
    \end{subfigure}
    
    \caption{\label{fig:text_repl} \textbf{Furniture replacement using text prompts:} Each pair consists of the original scene on the left and the one with the replaced furniture on the right, with the text prompt below it.}
\end{figure}

In addition to synthesizing the whole scene, \ours{} is also capable of modifying specific instances using text. To this end, we retrieve target objects using text descriptions which are embedded using the same CLIP embedder. In Figure \ref{fig:text_repl}, we demonstrated several examples in which our model can reliably retrieve relevant furniture instances and place them accordingly. Note that \ours{} adjusts the placement of the new object based on the instance-level information, and can re-adjust as needed instead of blindly replacing the previous object at the same position (e.g. when trading a bed for a lounge sofa). More examples are in Appendix Figure \ref{app_fig:text_repl}.

\section{Limitations} \label{sec:limitations}

\begin{figure}
    \centering
    \begin{subfigure}[b]{0.25\textwidth}
        \centering
        \adj{\includegraphics[width=\textwidth]{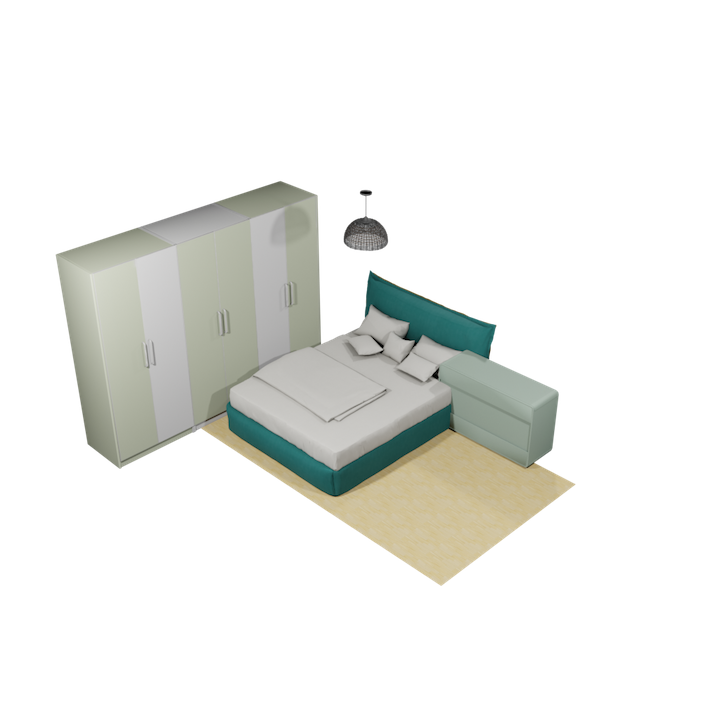}}
        \caption{Out of bounds \label{fig:fail_out_of_bounds}}
    \end{subfigure}
    \begin{subfigure}[b]{0.25\textwidth}
        \centering
        \adj{\includegraphics[width=\textwidth]{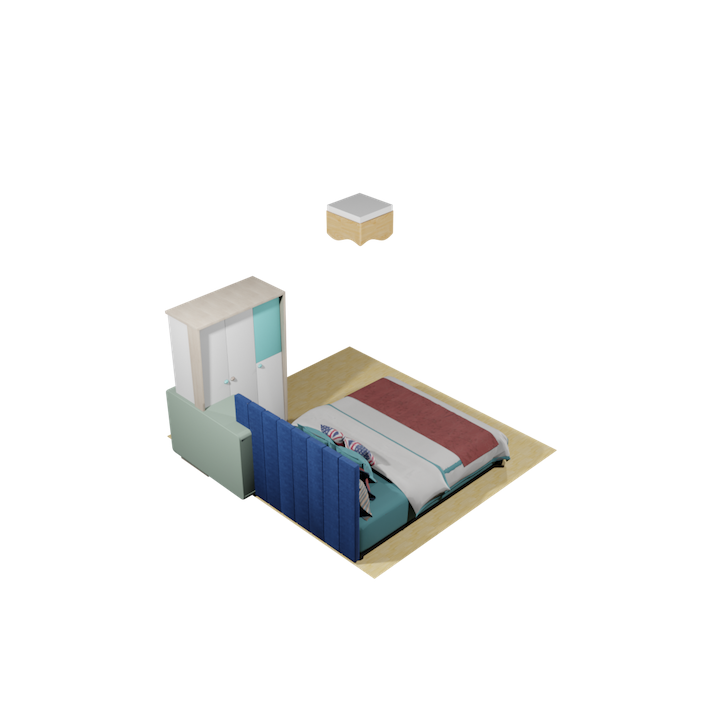}}
        \caption{Dual placement \label{fig:fail_dual_placement}}
    \end{subfigure}
    \begin{subfigure}[b]{0.25\textwidth}
        \centering
        \adj{\includegraphics[width=\textwidth]{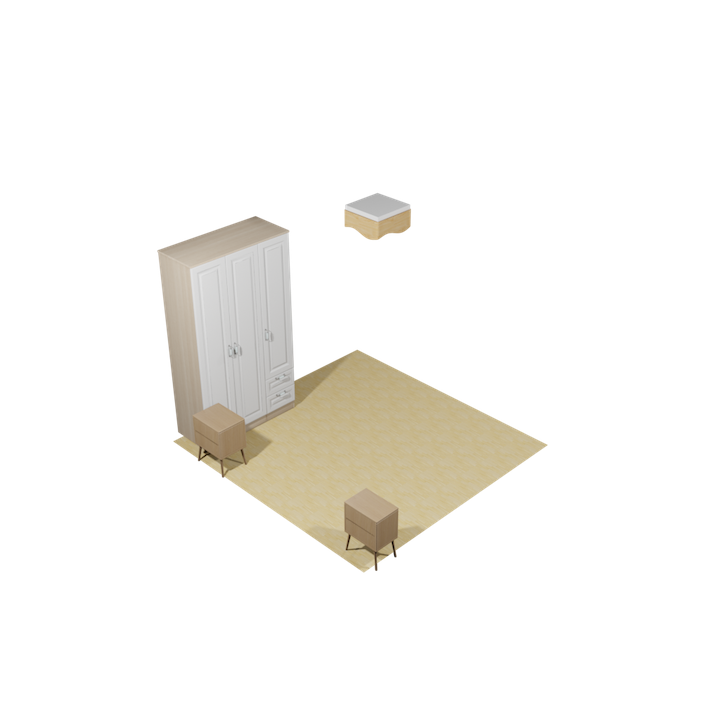}}
        \caption{Early stopping \label{fig:fail_early_stopping}}
    \end{subfigure}
    \caption{\label{fig:failing} \textbf{Commonly seen failing examples.}}
\end{figure}

With a relatively small training dataset (Table \ref{tab:dataset}), our model sometimes fails to produce reasonable results. We list three common failure modes in Figure \ref{fig:failing}: Figure \ref{fig:fail_out_of_bounds} shows a case where a furniture is placed outside the floor plan. Despite visually unrealistic, we found that similar malformed scenes are present in the training data. Since our model is intended to be purely data-driven, we left for future works to inject human design priors into the dataset if needed with learned methods \cite{enhancer} or manual constraints. Sometimes, the model places duplicated furniture as in Figure \ref{fig:fail_dual_placement}. We hypothesize it to be similar to unwanted repetitive phrases in text generation models. The last commonly seen pitfall is early stopping as in Figure \ref{fig:fail_early_stopping}, which is an artifact of training a separate head for the stop token on heavily unbalanced label distribution. We believe these shortcomings could be potentially addressed with better architectures, more training data, or explicit calibration of the synthesis length. Unfortunately, building large scale, high quality indoor scene datasets with dense annotations is expensive and time-consuming. This could be good motivation to invest in data collection efforts, or pursue unsupervised and semi-supervised methods.

\section{Conclusions}
In this paper, we introduce \ours{}, a style-consistent indoor scene model with semantic furniture embedding, which can synthesize and auto-complete scenes that follow the style and aesthetics of the dataset while also preserving the furniture configurations. Thanks to the multi-modality and semantic furniture embedding, \ours{} can accomplish many user-controlled tasks, such as text-guided furniture replacement and scene synthesis with decent zero-shot performance. We have shown both qualitatively and quantitatively that our model outperforms the baseline in different settings. To the best of our knowledge, our model is the first to take visual characteristics of furniture instances into account. We hope \ours{} can be extended to larger, more diverse datasets and new tasks including full building interiors, thereby enabling many downstream tasks that require automatic scene synthesis, such as navigational agent training, virtual world construction, spatial computing platforms, and architectural design. 

{
    \small
    \bibliographystyle{unsrt}
    \bibliography{main}
}

\clearpage
\section{Appendix for \ours{}: Style-Consistent Indoor Scene Synthesis
with Semantic Furniture Embedding}
\appendix
\newcommand{\hrulesep}{\unskip\ \hrule\ }

\section{Full Scene Completion Metrics} \label{app:full_metrics}
In Section \ref{sec:scene_comp}, we list the metrics for the bedroom type because the bedroom scenes have the most data with correct layouts and furniture count. Here we provide the full evaluation metrics table of all room types in Table \ref{table:loo_full}.  Overall the conclusions form the bedroom scenes hold for other room types.

\begin{table}
  \caption{Metrics of partial scene completion of all room types}
  \label{table:nfurniture_full}
  \centering
  \begin{tabular}{lllllll}
    \toprule
    Room type & Prepopulated \# & Model & FID(↓) & CLIP-FID(↓) & Precision(↑) & Recall(↑) \\
    \midrule
    \multirow{10}{*}{Bedroom}
        & \multirow{2}{*}{1 furniture} 
          & ATISS & \textbf{16.38} & \textbf{0.80} & 21.6\% & 23.0\% \\
        & & \ours{} & 17.83 & 1.03 & \textbf{36.5\%} & \textbf{31.6\%} \\
        \cmidrule{2-7}
        & \multirow{2}{*}{2 furniture} 
          & ATISS & 14.87 & 0.71 & 26.3\% & 29.5\% \\
        & & \ours{} & \textbf{14.08} & \textbf{0.67} & \textbf{40.7\%} & \textbf{42.1\%} \\
        \cmidrule{2-7}
        & \multirow{2}{*}{3 furniture} 
          & ATISS & 13.51 & 0.64 & 26.9\% & 31.2\% \\
        & & \ours{} & \textbf{12.11} & \textbf{0.53} & \textbf{40.4\%} & \textbf{38.2\%} \\
        \cmidrule{2-7}
        & \multirow{2}{*}{4 furniture} 
          & ATISS & 12.53	& 0.63 & 26.7\% & 31.4\% \\
        & & \ours{} & \textbf{10.80} & \textbf{0.46} & \textbf{38.4\%} & \textbf{41.9\%} \\
        \cmidrule{2-7}
        & \multirow{2}{*}{5 furniture} 
          & ATISS & 11.84 & 0.59 & 26.2\% & 31.5\% \\
        & & \ours{} & \textbf{9.88} & \textbf{0.42} & \textbf{39.1\%} & \textbf{42.5\%} \\
    \midrule
    \multirow{10}{*}{Dining room}
        & \multirow{2}{*}{1 furniture} 
          & ATISS & 21.42	& 1.21 & 12.5\% & 12.9\% \\
        & & \ours{} & 21.66 & 1.18 & \textbf{21.8\%} & \textbf{25.1\%} \\
        \cmidrule{2-7}
        & \multirow{2}{*}{2 furniture} 
          & ATISS & 20.60 & 1.13 & 18.7\% & 19.5\% \\
        & & \ours{} & 20.24 & \textbf{1.01} & \textbf{27.9\%} & \textbf{31.6\%} \\
        \cmidrule{2-7}
        & \multirow{2}{*}{3 furniture} 
          & ATISS & 19.51 & 1.02 & 20.6\% & 21.7\% \\
        & & \ours{} & 19.19 & \textbf{0.95} & \textbf{29.5\%} & \textbf{33.5\%} \\
        \cmidrule{2-7}
        & \multirow{2}{*}{4 furniture} 
          & ATISS & 18.61 & 0.95 & 21.5\% & 23.0\% \\
        & & \ours{} & 18.13 & \textbf{0.90} & \textbf{30.1\%} & \textbf{34.2\%} \\
        \cmidrule{2-7}
        & \multirow{2}{*}{5 furniture} 
          & ATISS & 17.16 & 0.80 & 21.2\% & 23.1\% \\
        & & \ours{} & 17.10 & 0.78 & \textbf{29.4\%} & \textbf{34.5\%} \\
    \midrule
    \multirow{10}{*}{Library}
        & \multirow{2}{*}{1 furniture} 
          & ATISS & 49.84 & 3.19 & 10.7\% & 11.1\% \\
        & & \ours{} & \textbf{47.56} & \textbf{3.08} & \textbf{16.7\%} & \textbf{21.3\%} \\
        \cmidrule{2-7}
        & \multirow{2}{*}{2 furniture} 
          & ATISS & 44.91 & 2.73 & 11.2\% & 11.7\% \\
        & & \ours{} & \textbf{41.60} & \textbf{2.60} & \textbf{20.4\%} & \textbf{26.9\%} \\
        \cmidrule{2-7}
        & \multirow{2}{*}{3 furniture} 
          & ATISS & 40.74 & 2.37 & 13.9\% & 15.0\% \\
        & & \ours{} & \textbf{37.78} & \textbf{2.27} & \textbf{22.6\%} & \textbf{27.5\%} \\
        \cmidrule{2-7}
        & \multirow{2}{*}{4 furniture} 
          & ATISS & 38.84 & 2.16 & 12.7\% & 14.0\% \\
        & & \ours{} & \textbf{34.93} & \textbf{2.03} & \textbf{20.7\%} & \textbf{26.0\%} \\
        \cmidrule{2-7}
        & \multirow{2}{*}{5 furniture} 
          & ATISS & 37.32 & 2.03 & 14.4\% & 14.8\% \\
        & & \ours{} & \textbf{33.18} & \textbf{1.93} & \textbf{19.6\%} & \textbf{24.9\%} \\
    \midrule
    \multirow{10}{*}{Living room}
        & \multirow{2}{*}{1 furniture} 
          & ATISS & \textbf{17.74} & \textbf{0.98} & 6.4\%           & 6.9\%           \\
        & & \ours{} & 19.98          & 1.10          & \textbf{12.3\%} & \textbf{13.2\%} \\
        \cmidrule{2-7}
        & \multirow{2}{*}{2 furniture} 
          & ATISS & \textbf{17.16} & 0.96          & 8.9\%           & 9.6\%           \\
        & & \ours{} & 18.34          & \textbf{0.91} & \textbf{16.2\%} & \textbf{17.5\%} \\
        \cmidrule{2-7}
        & \multirow{2}{*}{3 furniture} 
          & ATISS & \textbf{16.69} & 0.90          & 10.0\%          & 10.8\%          \\
        & & \ours{} & 17.48          & \textbf{0.85} & \textbf{19.7\%} & \textbf{21.4\%} \\
        \cmidrule{2-7}
        & \multirow{2}{*}{4 furniture} 
          & ATISS & 16.16          & 0.86          & 10.6\%          & 12.2\%          \\
        & & \ours{} & 16.54          & \textbf{0.76} & \textbf{20.7\%} & \textbf{22.7\%} \\
        \cmidrule{2-7}
        & \multirow{2}{*}{5 furniture} 
          & ATISS & 15.56          & 0.83          & 11.6\%          & 13.1\%          \\
        & & \ours{} & 15.92          & \textbf{0.75} & \textbf{21.6\%} & \textbf{23.9\%} \\
    \bottomrule
  \end{tabular}
\end{table}

\section{Rendering Settings} \label{app:rendering}

To facilitate reproducibility, we detail the rendering settings that are specific to our experiment in this section. We first discuss the mesh object material standardization process and then the camera settings for the eight-view rendering of objects and scenes.

Since the model dataset \cite{3dfuture} has many instances with abnormal material properties like 0 diffuse or 1 specular, which would make rendered images look either too dark or bright, in order to render images more consistently and generalizable to all meshes, we manually adjust the material properties when loading the mesh from the dataset. During the experiments, we found that setting the ambient, diffuse, and specular to a ratio of (0.4, 0.4, 0.1) gives the most consistent lighting results.

We specify the settings with which we render images of an object or a scene from eight canonical directions to create object embeddings and evaluate models. We use \cite{simple_3dviz} as the rendering package for its efficiency and simplicity, and EGL as the backend. We summarize the camera settings in Table \ref{table:cam_setting}, and all parameters unspecified are the default values.

Note that we rely on \cite{simple_3dviz} as our rendering package during evaluation for its efficiency but all figures shown in the paper are post-processed using ray-tracing from Blender\cite{blender} with CYCLE engine under the three-point-light setup. We also fix the texture of the floor plan to make sure all metrics are not influenced by the choice during the evaluation.

\begin{table}
  \caption{Camera settings for simple-3dviz.}
  \label{table:cam_setting}
  \centering
  \begin{tabular}{llllll}
    \toprule
    Task & Distance & Elevation & Look at & Azimuth \\
    \midrule
    Object embedding & 3 & 1 & (0,0,0) & Eight views \\
    Scene visualization & 8 & 2.5 & (0,1,0) & Eight views \\
    \bottomrule
  \end{tabular}
\end{table}

\section{More Visual Examples} \label{app:visual_examples}

\begin{figure}
    \centering
    \begin{subfigure}[b]{0.48\textwidth}
        \centering
        \adj{\includegraphics[width=0.5\textwidth]{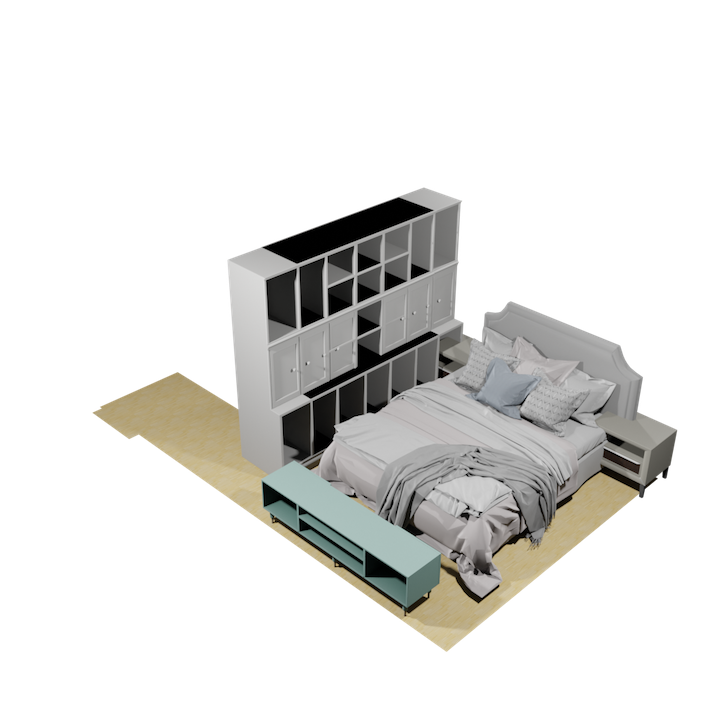}}
        \adj{\includegraphics[width=0.5\textwidth]{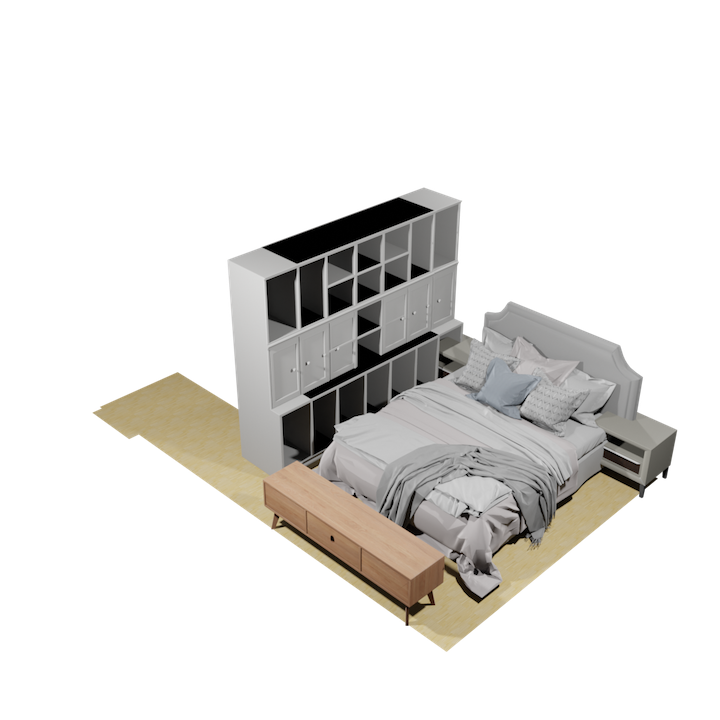}}
        \caption{From a cyan to ``a solid wooden TV stand''}
    \end{subfigure}
    \begin{subfigure}[b]{0.48\textwidth}
        \centering
        \adj{\includegraphics[width=0.5\textwidth]{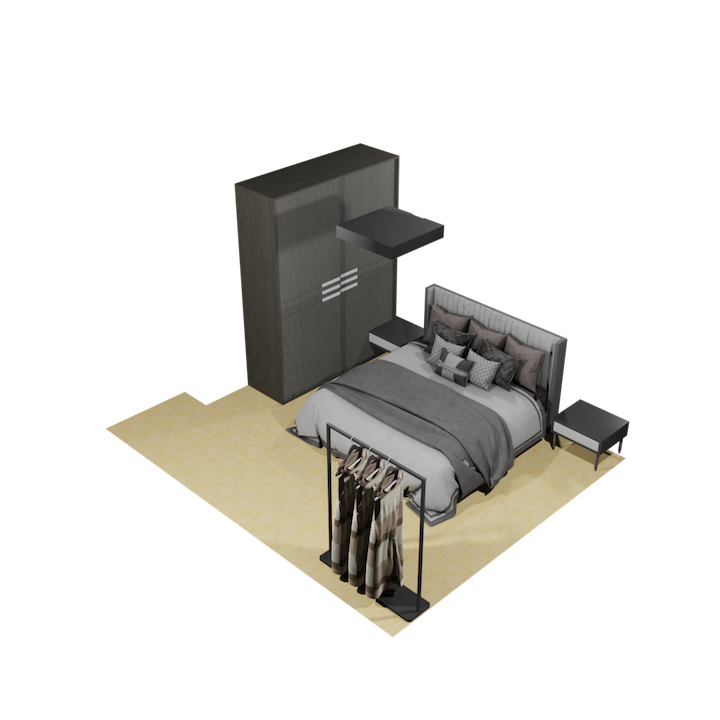}}
        \adj{\includegraphics[width=0.5\textwidth]{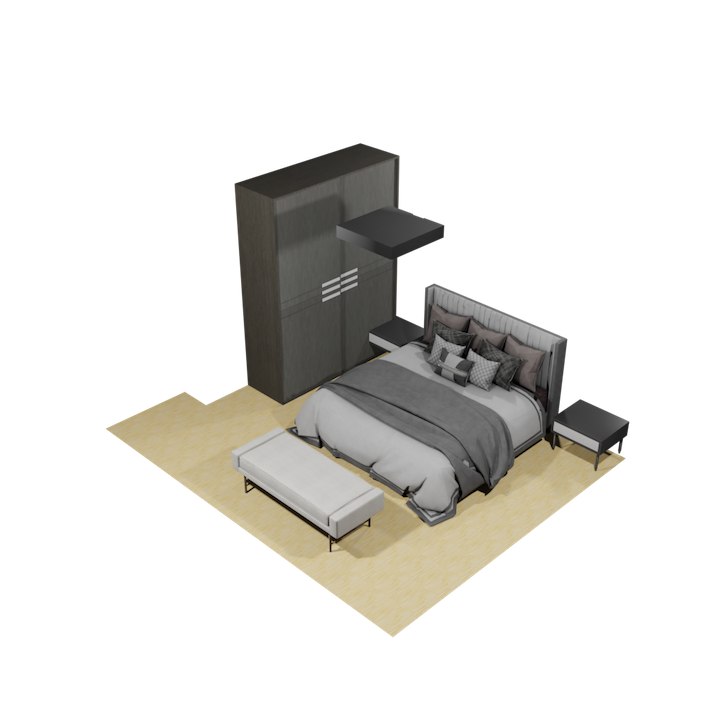}}
        \caption{From the clothes hanger to ``a white bed ottoman''}
    \end{subfigure}
    
    \begin{subfigure}[b]{0.48\textwidth}
        \centering
        \adj{\includegraphics[width=0.5\textwidth]{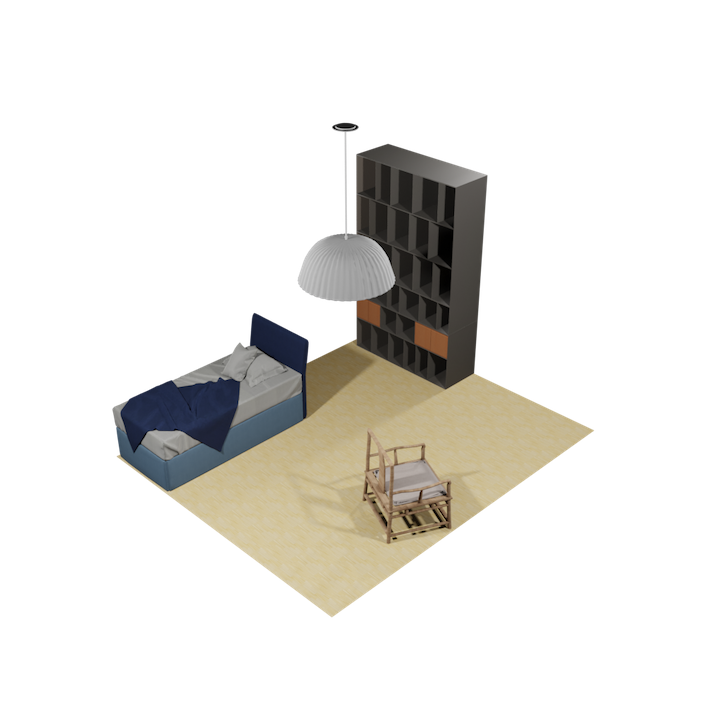}}
        \adj{\includegraphics[width=0.5\textwidth]{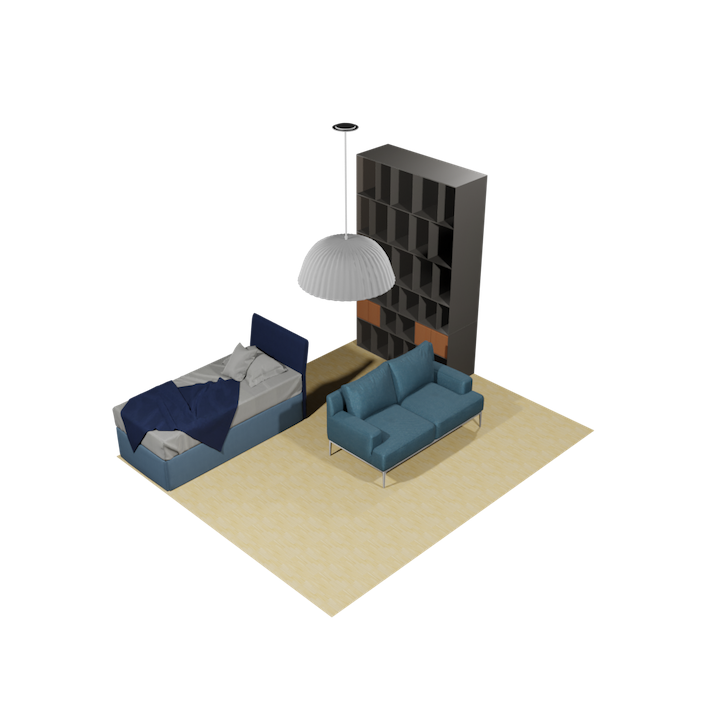}}
        \caption{From the chair to ``a blue modern sofa''}
    \end{subfigure}
    \begin{subfigure}[b]{0.48\textwidth}
        \centering
        \adj{\includegraphics[width=0.5\textwidth]{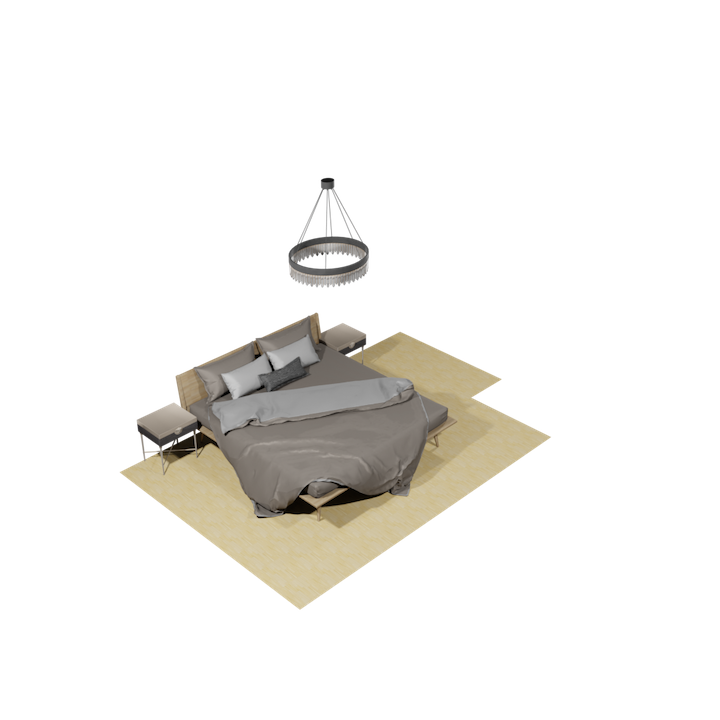}}
        \adj{\includegraphics[width=0.5\textwidth]{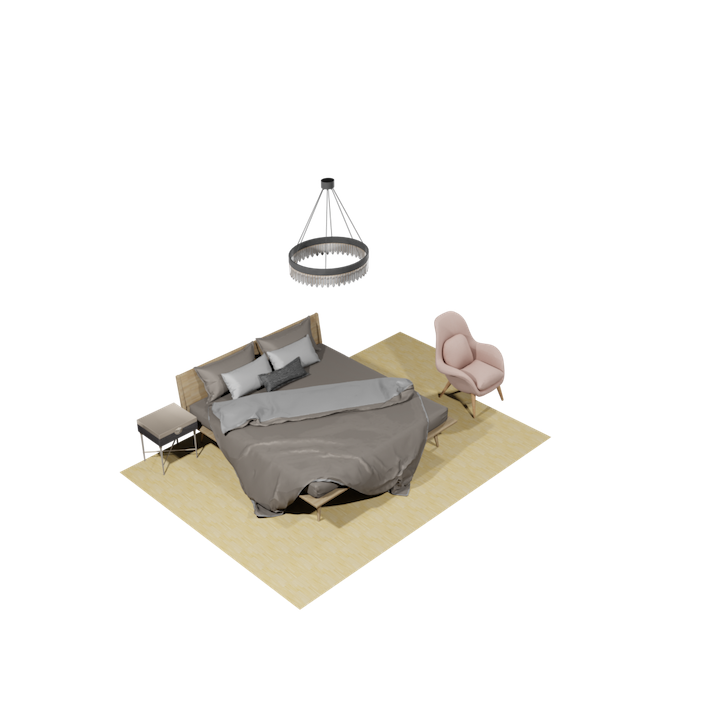}}
        \caption{From the nightstand to ``a small pink chair''}
    \end{subfigure}
    
    \caption{\label{app_fig:text_repl} \textbf{Furniture replacement using text prompts:} \ours{} can replace a furniture in the scene to a new instance described by a text prompt, with instance-dependent transformation. Each pair consists of the original scene on the left and the one with the replaced furniture on the right, with the text prompt below it}
\end{figure}

\begin{figure*}
    \centering
    \begin{subfigure}[b]{0.2\textwidth}
        \centering
        \adj{\includegraphics[width=\textwidth]{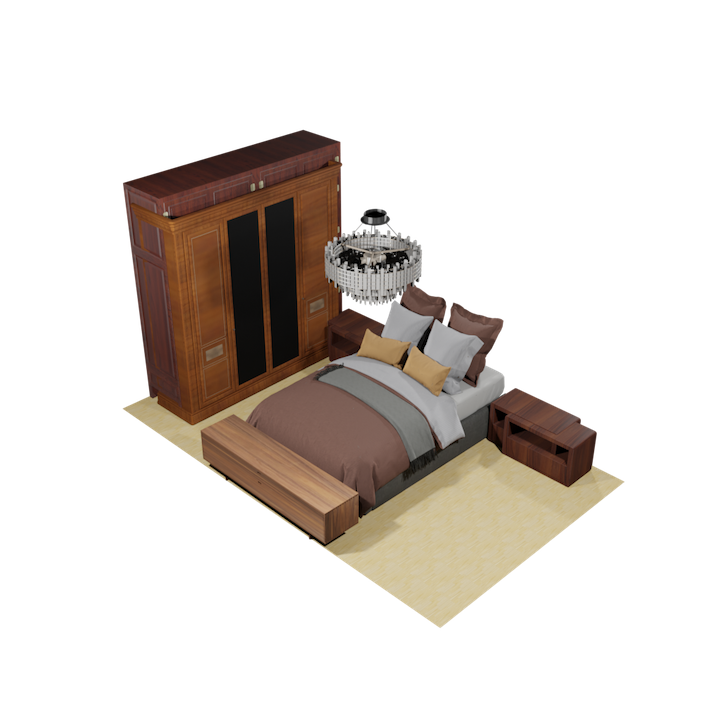}}
        \caption{Dark Wood}
    \end{subfigure}\hfill
    \begin{subfigure}[b]{0.2\textwidth}
        \centering
        \adj{\includegraphics[width=\textwidth]{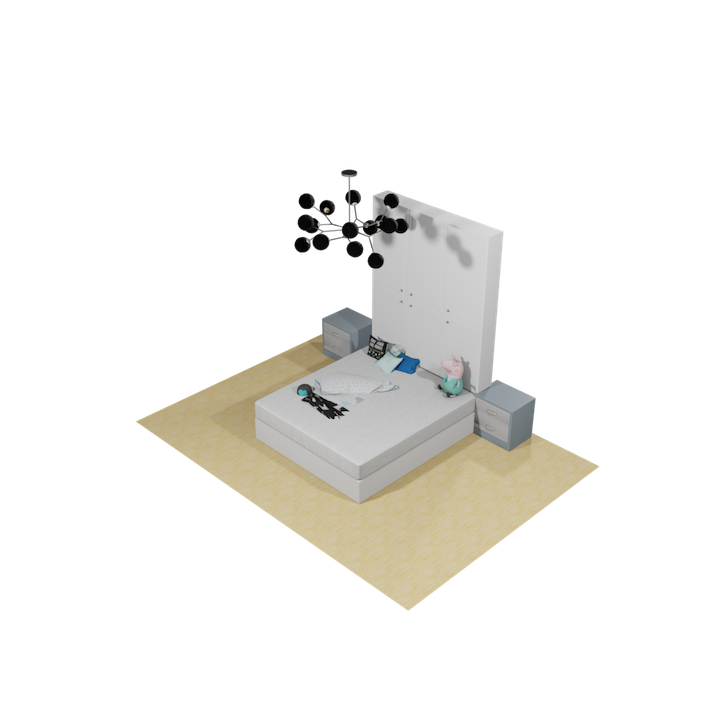}}
        \caption{Cute White}
    \end{subfigure}\hfill
    \begin{subfigure}[b]{0.2\textwidth}
        \centering
        \adj{\includegraphics[width=\textwidth]{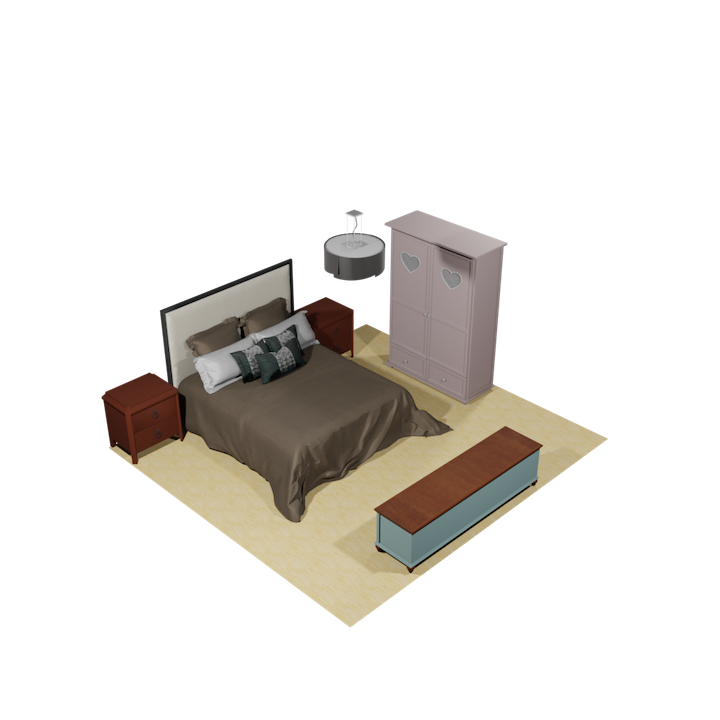}}
        \caption{Colorful}
    \end{subfigure}\hfill
    \begin{subfigure}[b]{0.2\textwidth}
        \centering
        \adj{\includegraphics[width=\textwidth]{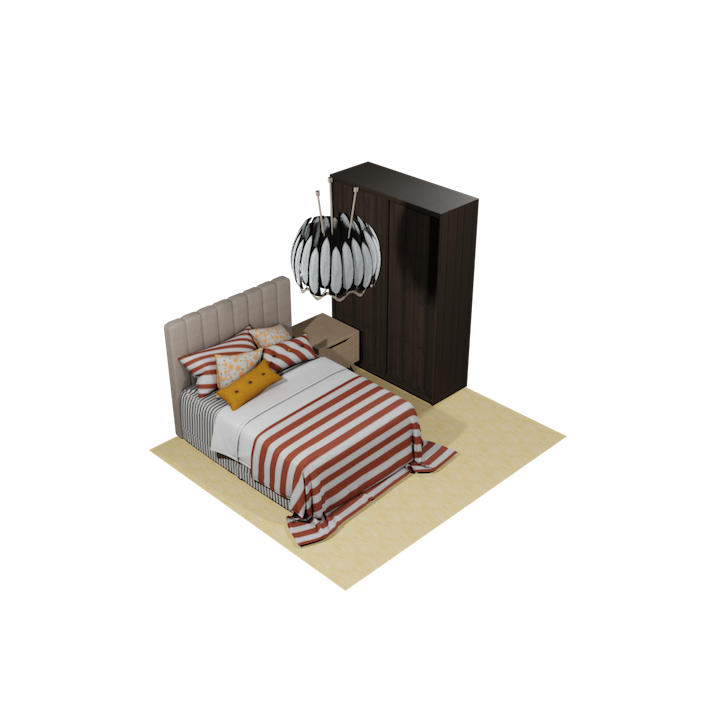}}
        \caption{Red Striped}
    \end{subfigure}\hfill
    \begin{subfigure}[b]{0.2\textwidth}
        \centering
        \adj{\includegraphics[width=\textwidth]{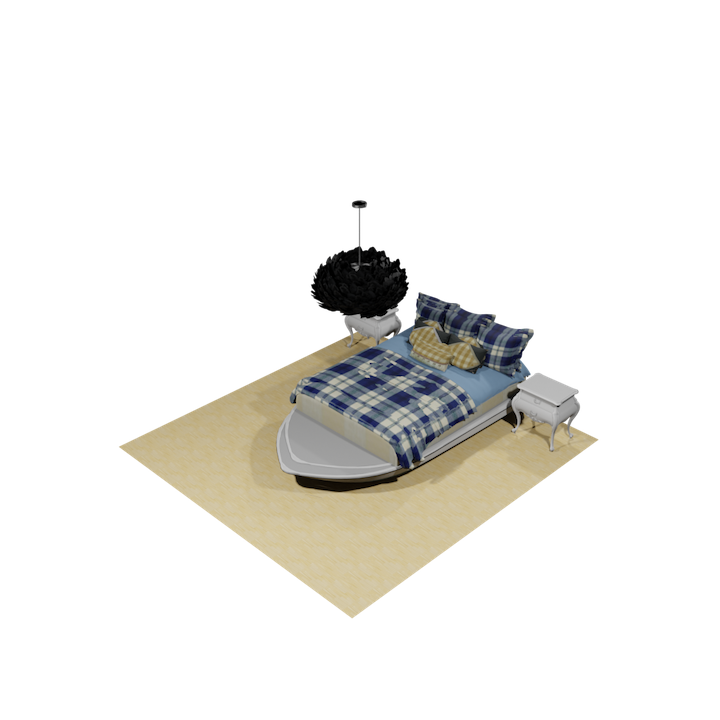}}
        \caption{Triangular}
    \end{subfigure}

    \begin{subfigure}[b]{0.2\textwidth}
        \centering
        \adj{\includegraphics[width=\textwidth]{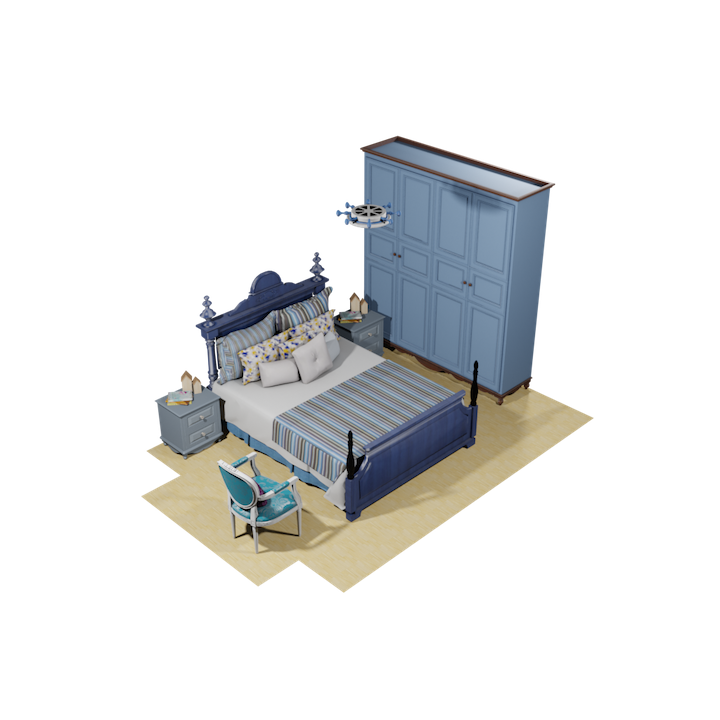}}
        \caption{Royal Blue}
    \end{subfigure}\hfill
    \begin{subfigure}[b]{0.2\textwidth}
        \centering
        \adj{\includegraphics[width=\textwidth]{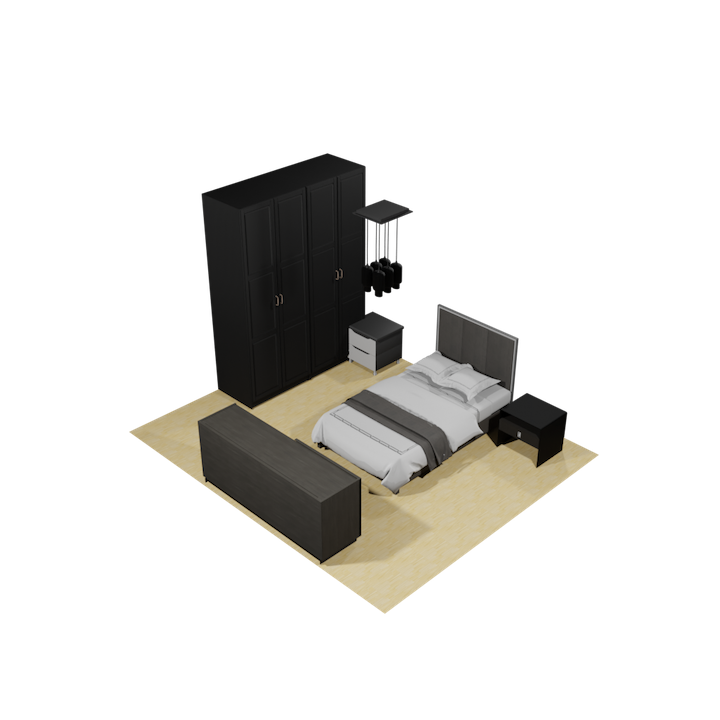}}
        \caption{Industrial}
    \end{subfigure}\hfill
    \begin{subfigure}[b]{0.2\textwidth}
        \centering
        \adj{\includegraphics[width=\textwidth]{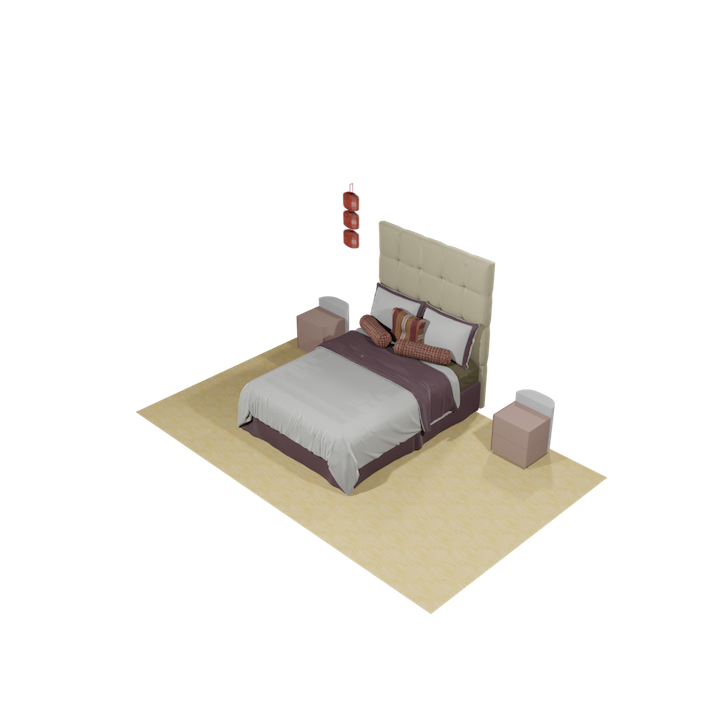}}
        \caption{Childish Pink}
    \end{subfigure}\hfill
    \begin{subfigure}[b]{0.2\textwidth}
        \centering
        \adj{\includegraphics[width=\textwidth]{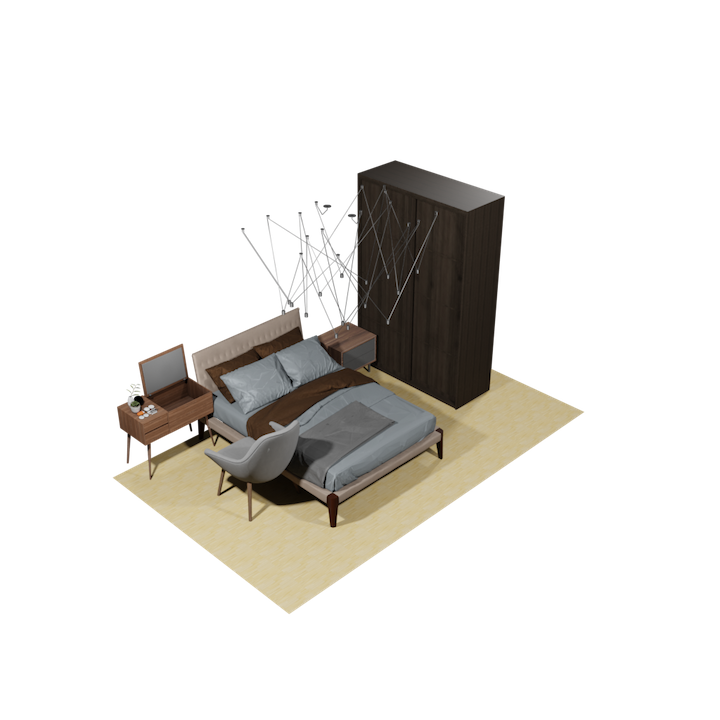}}
        \caption{Designer}
    \end{subfigure}\hfill
    \begin{subfigure}[b]{0.2\textwidth}
        \centering
        \adj{\includegraphics[width=\textwidth]{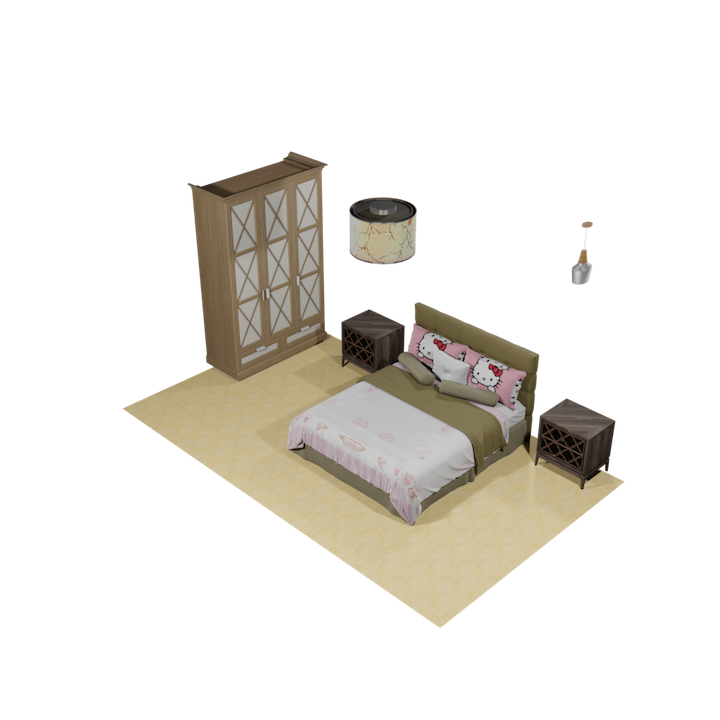}}
        \caption{Hello Kitty}
    \end{subfigure}
    
    \caption{\label{app_fig:text_gen} \textbf{Scene synthesis from room masks:} Visual examples of scenes generated from floor plans where the first row contains samples from ATISS and the bottom two rows by \ours{}.}
\end{figure*}

\begin{figure*}
    \centering

    \raisebox{40pt}{\rotatebox[origin=c]{90}{ATISS}}
    \begin{subfigure}[b]{0.95\textwidth}
        \centering
        \adj{\includegraphics[width=0.2\textwidth]{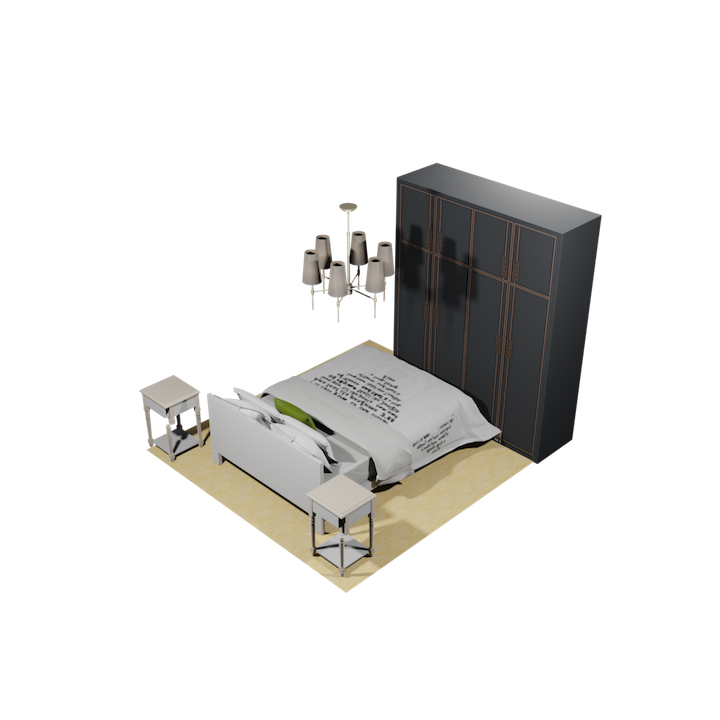}}
        \adj{\includegraphics[width=0.2\textwidth]{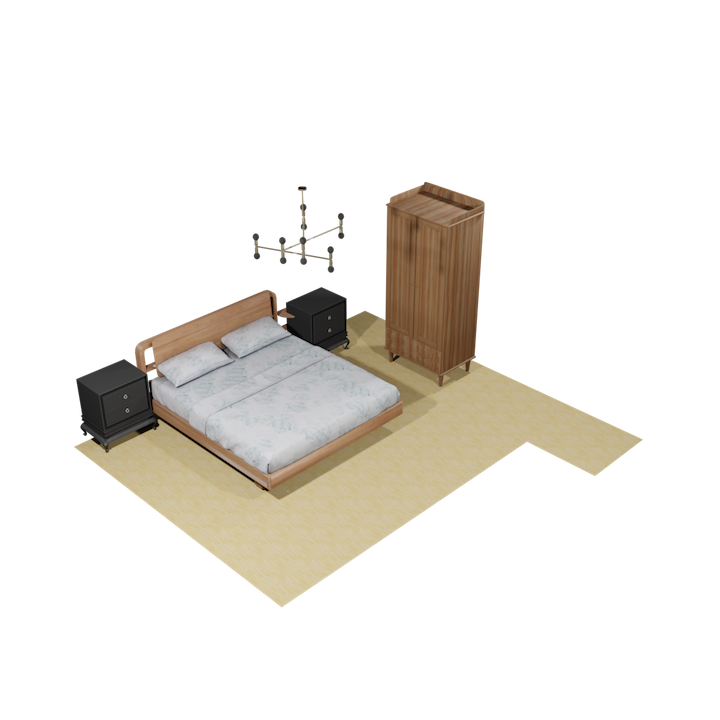}}
        \adj{\includegraphics[width=0.2\textwidth]{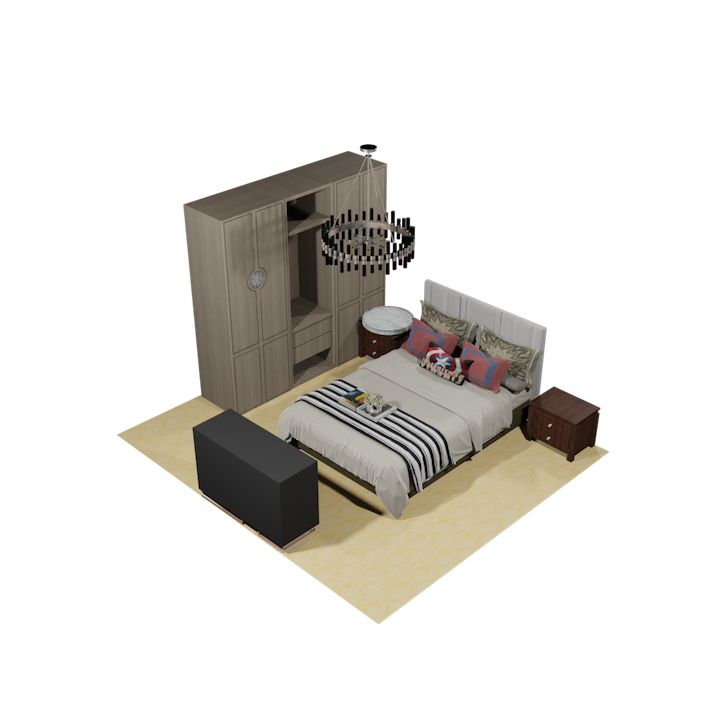}}
        \adj{\includegraphics[width=0.2\textwidth]{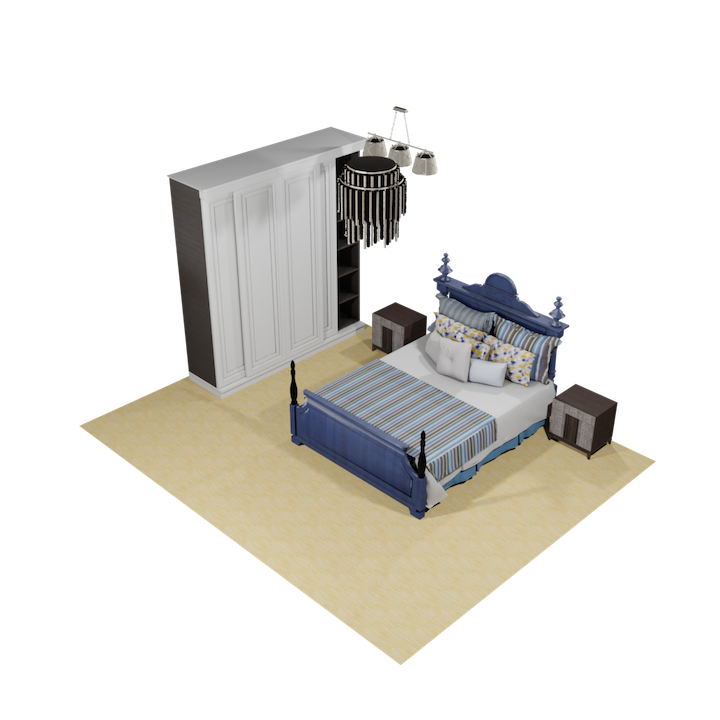}}
        \adj{\includegraphics[width=0.2\textwidth]{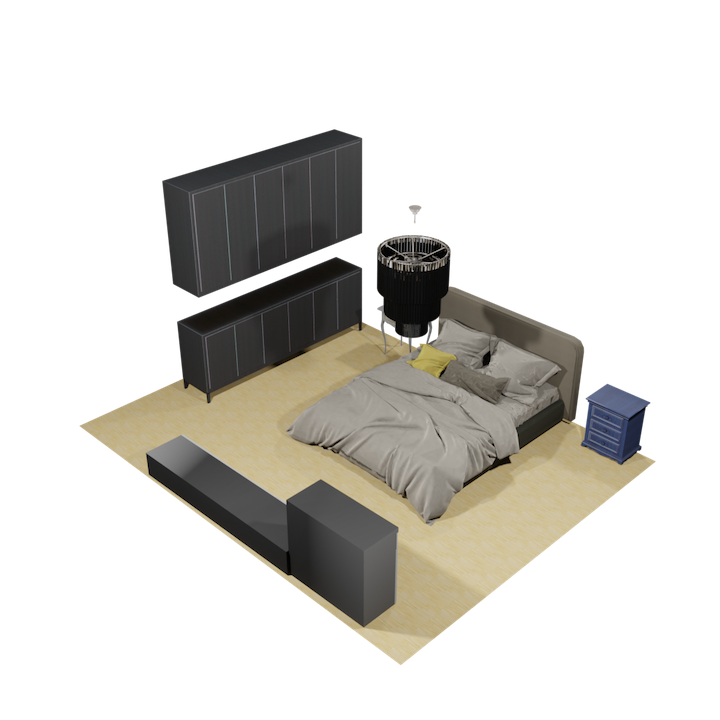}}
    \end{subfigure}
    \vspace{10pt}\hrulesep

    \raisebox{70pt}{\rotatebox[origin=c]{90}{Ours}}
    \begin{subfigure}[b]{0.95\textwidth}
        \centering
        \adj{\includegraphics[width=0.2\textwidth]{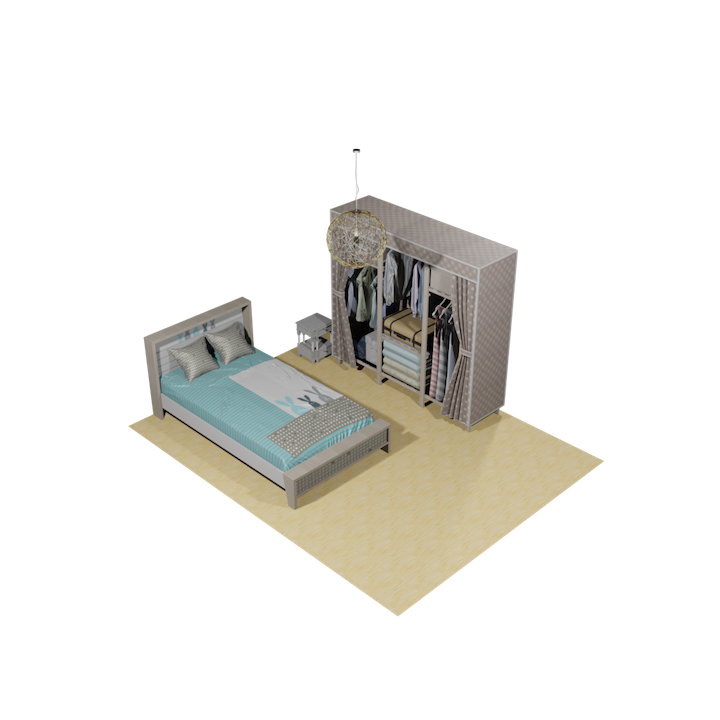}}
        \adj{\includegraphics[width=0.2\textwidth]{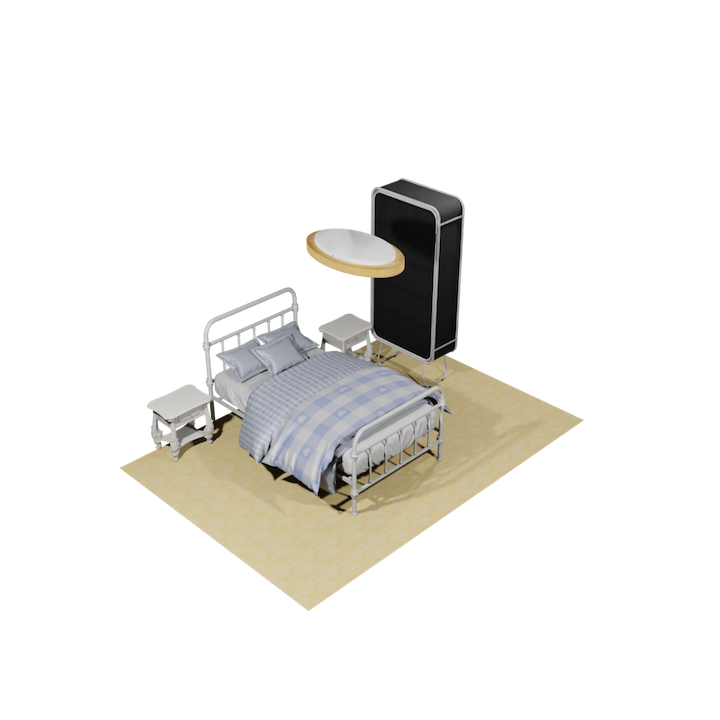}}
        \adj{\includegraphics[width=0.2\textwidth]{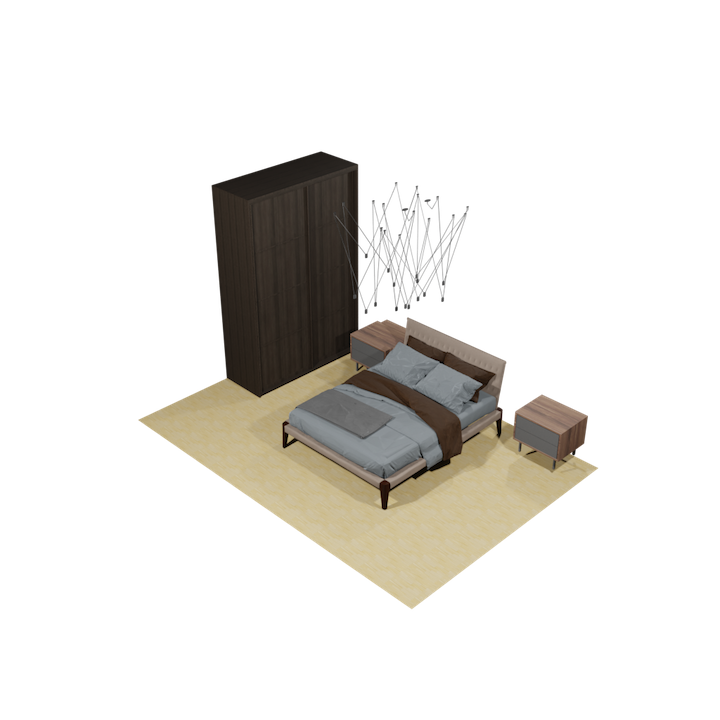}}
        \adj{\includegraphics[width=0.2\textwidth]{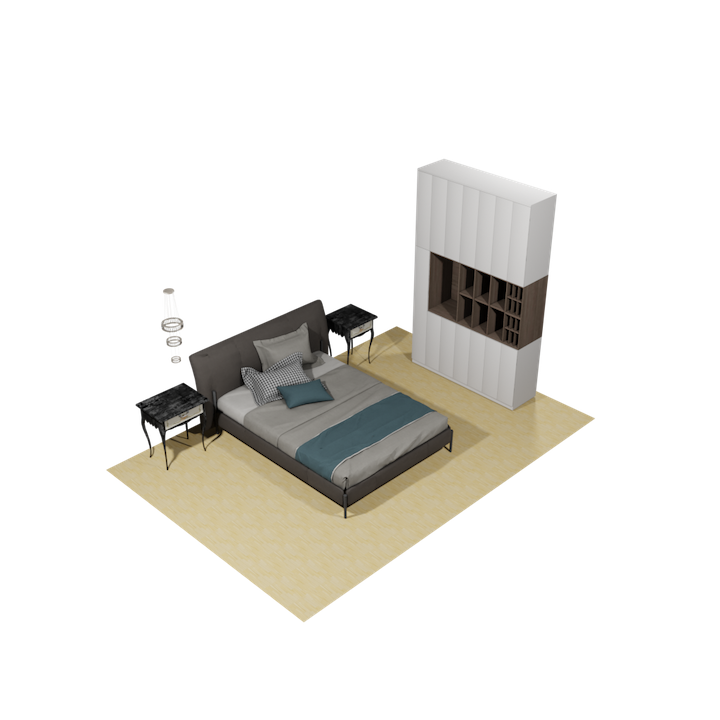}}
        \adj{\includegraphics[width=0.2\textwidth]{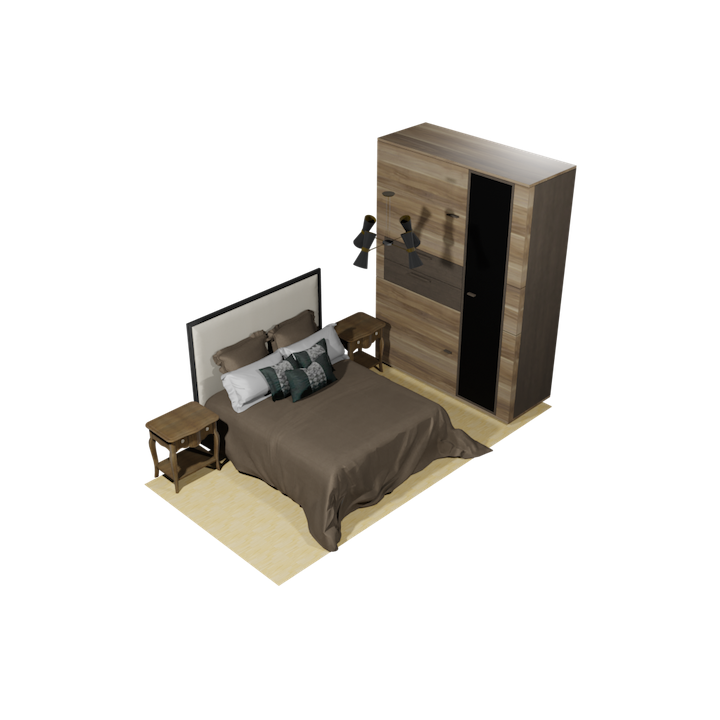}}
        
        \adj{\includegraphics[width=0.2\textwidth]{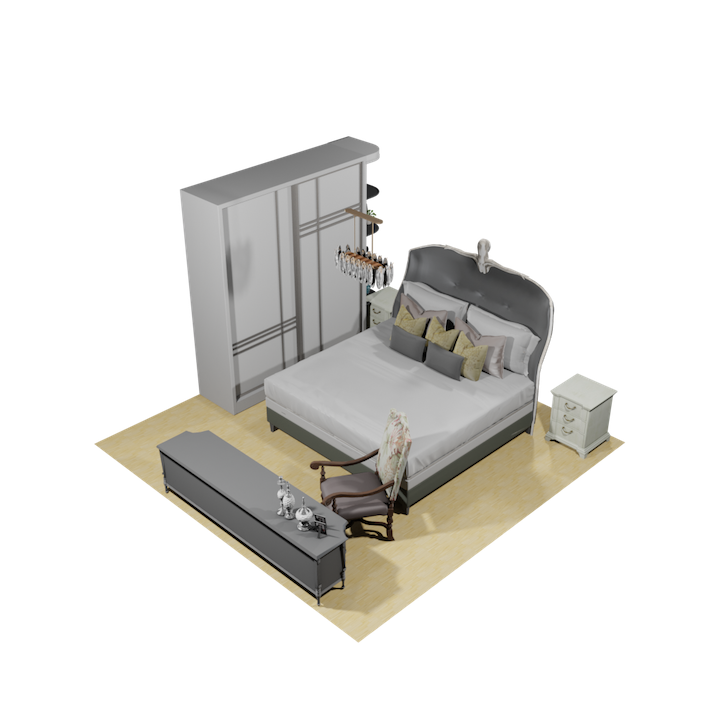}}
        \adj{\includegraphics[width=0.2\textwidth]{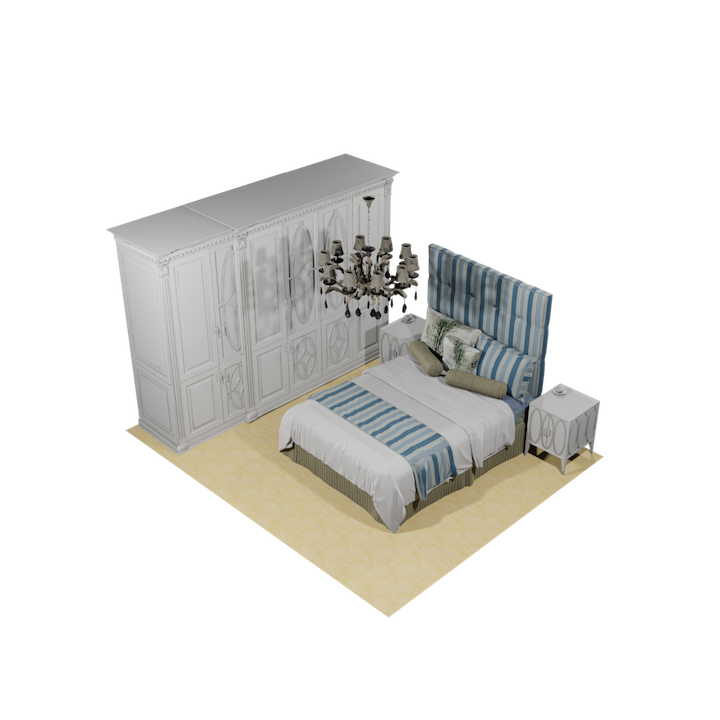}}
        \adj{\includegraphics[width=0.2\textwidth]{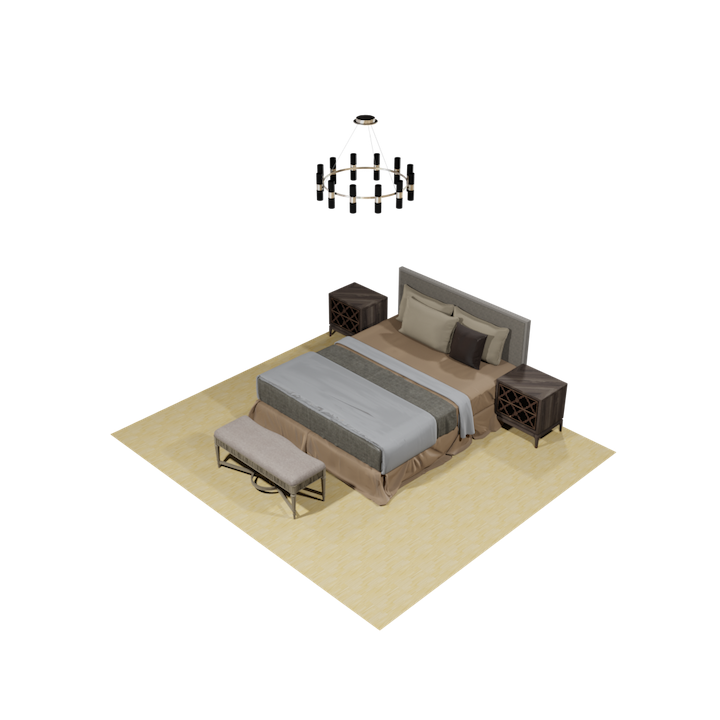}}
        \adj{\includegraphics[width=0.2\textwidth]{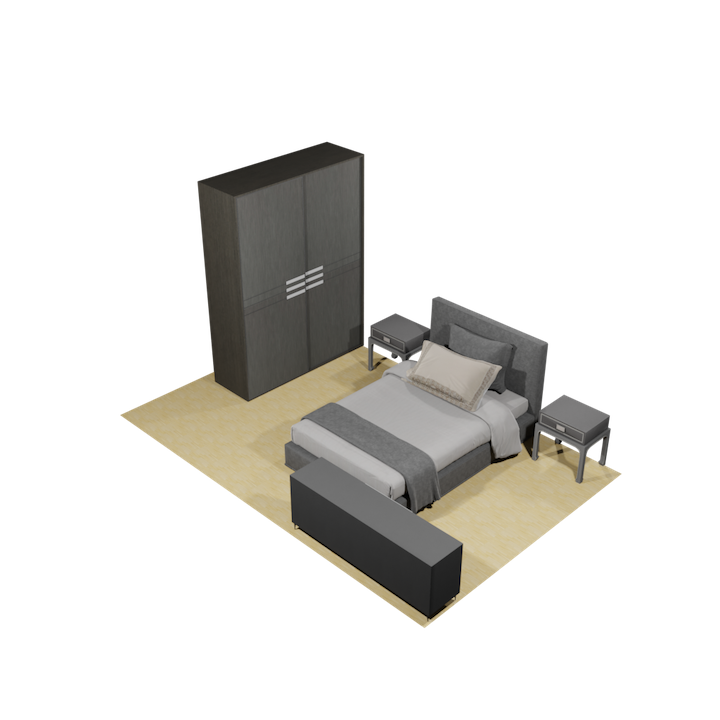}}
        \adj{\includegraphics[width=0.2\textwidth]{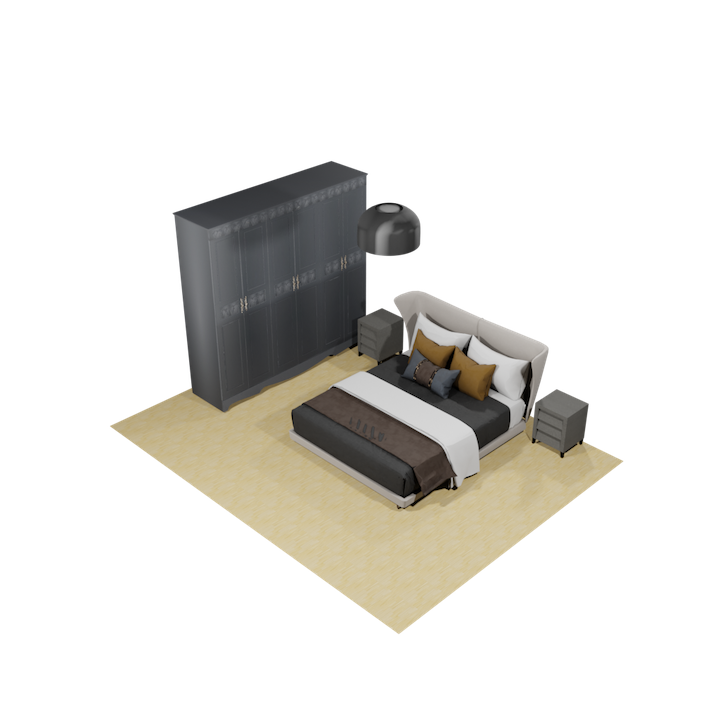}}
    \end{subfigure}
    
    \caption{\label{app_fig:uncond_gen} \textbf{Visual comparison of scene synthesis from floor plans:} The top row contains samples from the baseline model and the ones below the divider are samples from our model.}
\end{figure*}

\begin{figure}
    \captionsetup[subfigure]{labelformat=empty}
    \centering
    
    \begin{subfigure}[b]{0.25\textwidth}
        \centering
        \adj{\includegraphics[width=\textwidth]{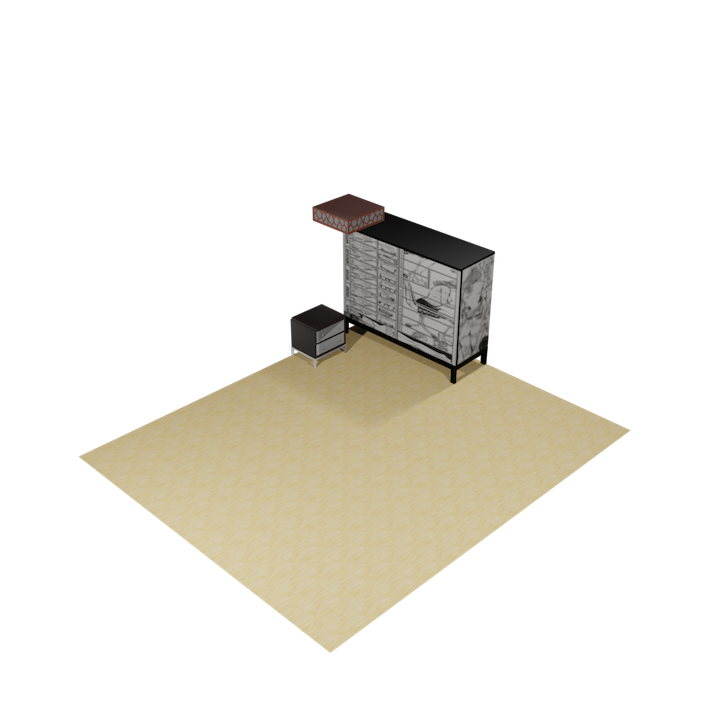}}
    \end{subfigure}
    \begin{subfigure}[b]{0.25\textwidth}
        \centering
        \adj{\includegraphics[width=\textwidth]{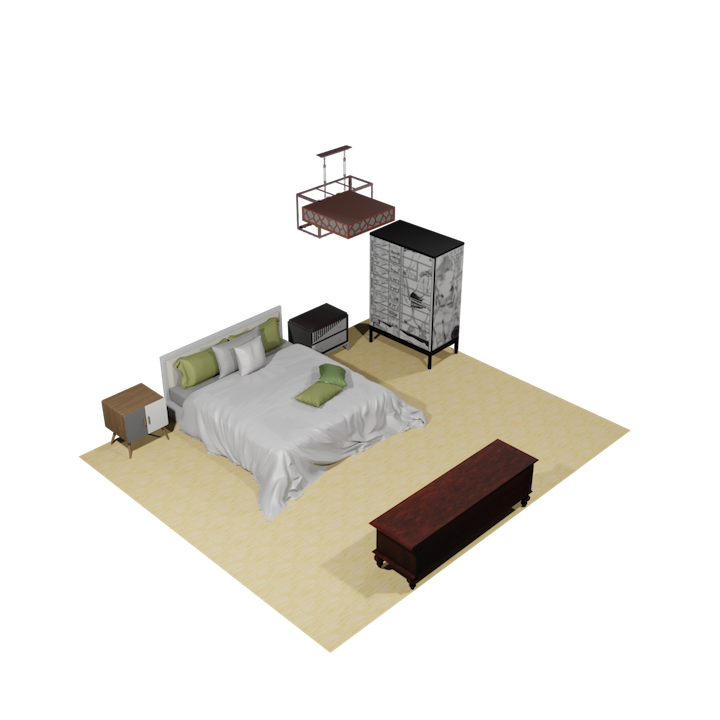}}
    \end{subfigure}
    \begin{subfigure}[b]{0.25\textwidth}
        \centering
        \adj{\includegraphics[width=\textwidth]{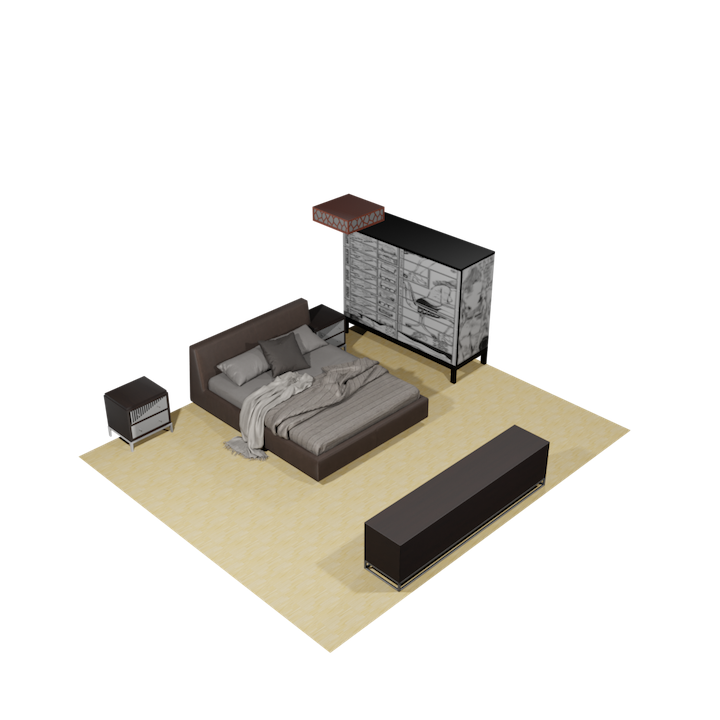}}
    \end{subfigure}

    \begin{subfigure}[b]{0.25\textwidth}
        \centering
        \adj{\includegraphics[width=\textwidth]{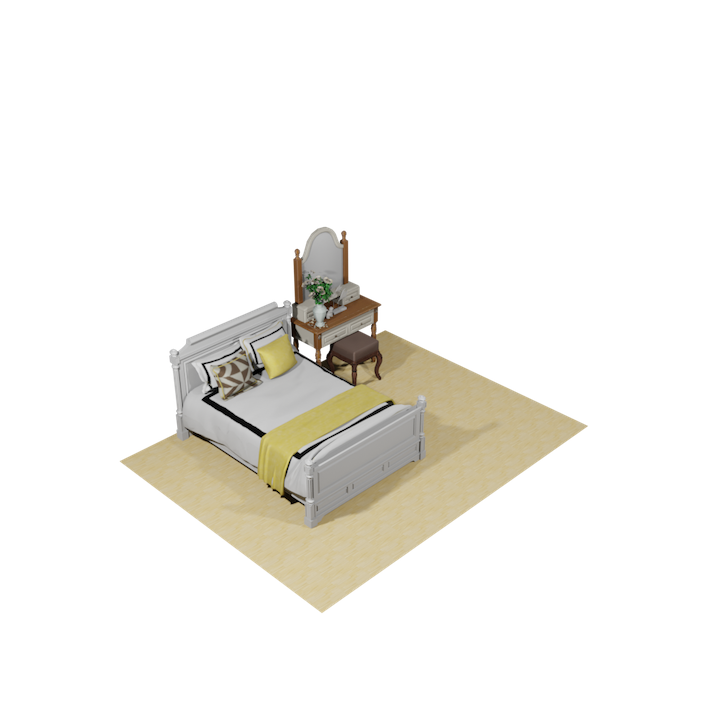}}
    \end{subfigure}
    \begin{subfigure}[b]{0.25\textwidth}
        \centering
        \adj{\includegraphics[width=\textwidth]{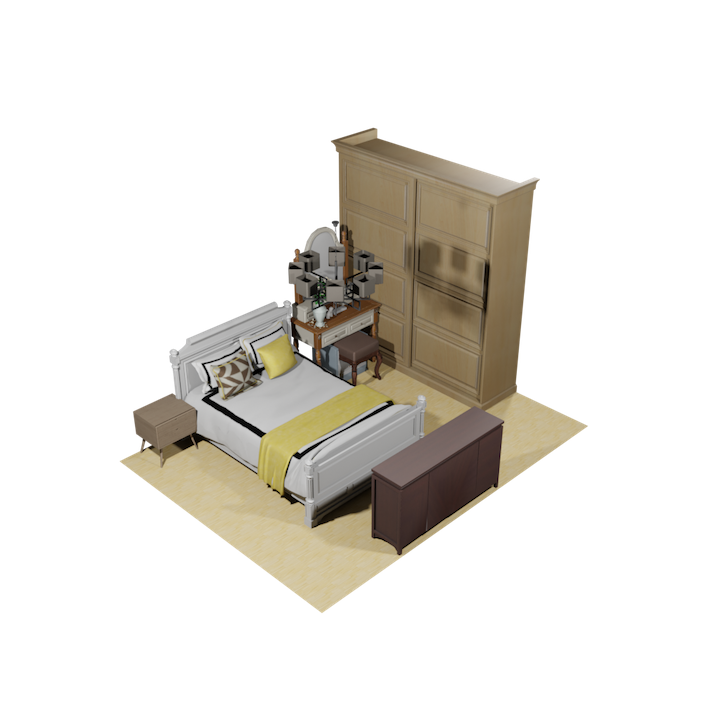}}
    \end{subfigure}
    \begin{subfigure}[b]{0.25\textwidth}
        \centering
        \adj{\includegraphics[width=\textwidth]{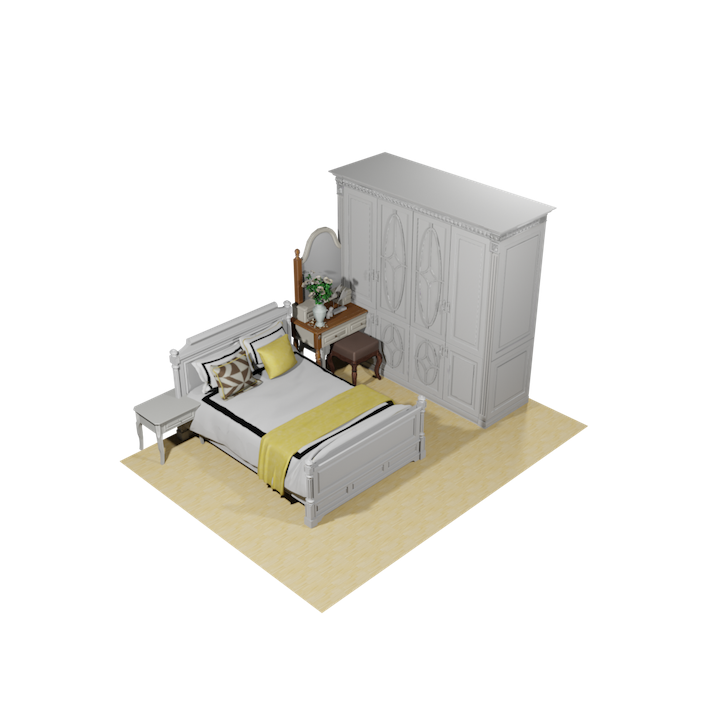}}
    \end{subfigure}

    \begin{subfigure}[b]{0.25\textwidth}
        \centering
        \adj{\includegraphics[width=\textwidth]{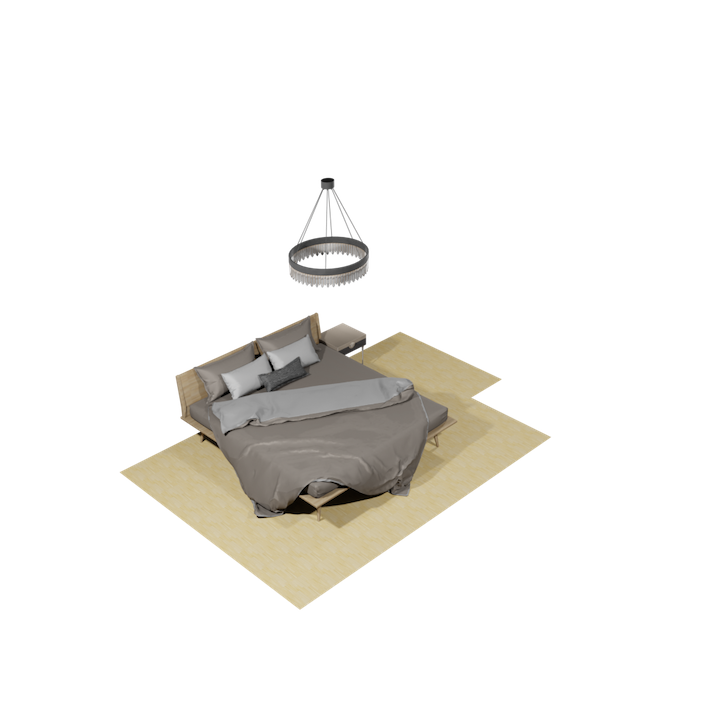}}
    \end{subfigure}
    \begin{subfigure}[b]{0.25\textwidth}
        \centering
        \adj{\includegraphics[width=\textwidth]{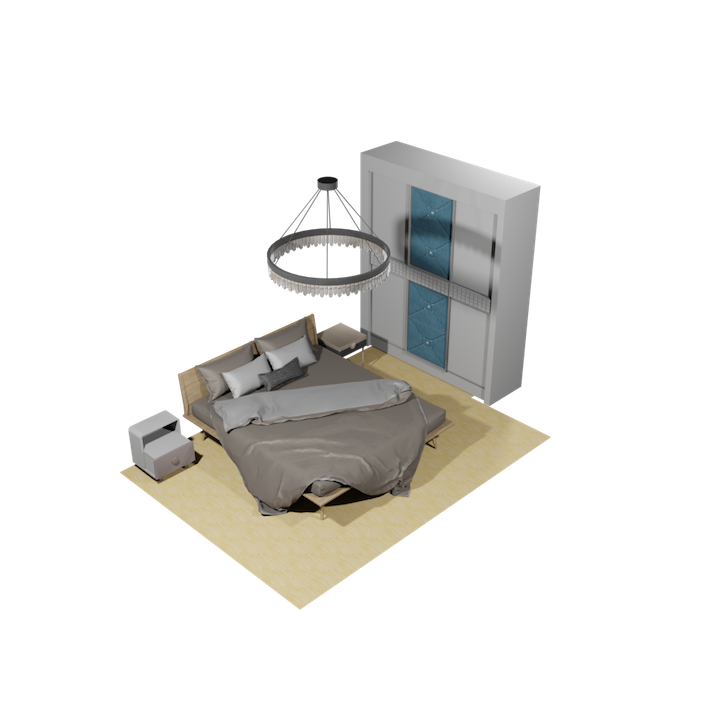}}
    \end{subfigure}
    \begin{subfigure}[b]{0.25\textwidth}
        \centering
        \adj{\includegraphics[width=\textwidth]{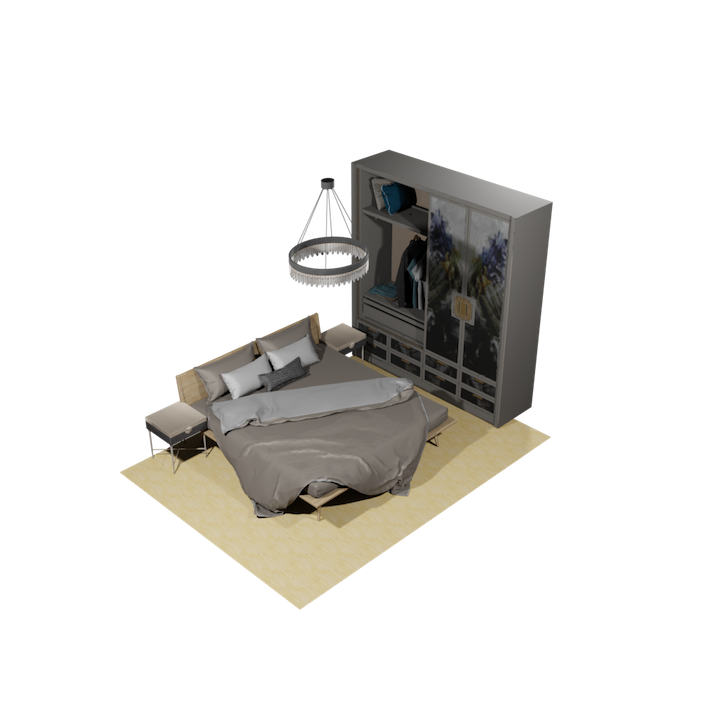}}
    \end{subfigure}

    \begin{subfigure}[b]{0.25\textwidth}
        \centering
        \adj{\includegraphics[width=\textwidth]{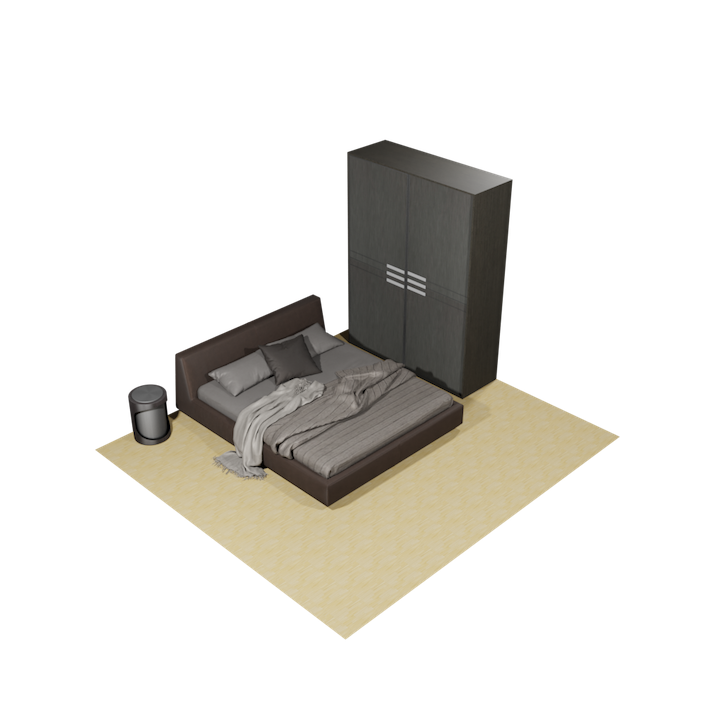}}
    \end{subfigure}
    \begin{subfigure}[b]{0.25\textwidth}
        \centering
        \adj{\includegraphics[width=\textwidth]{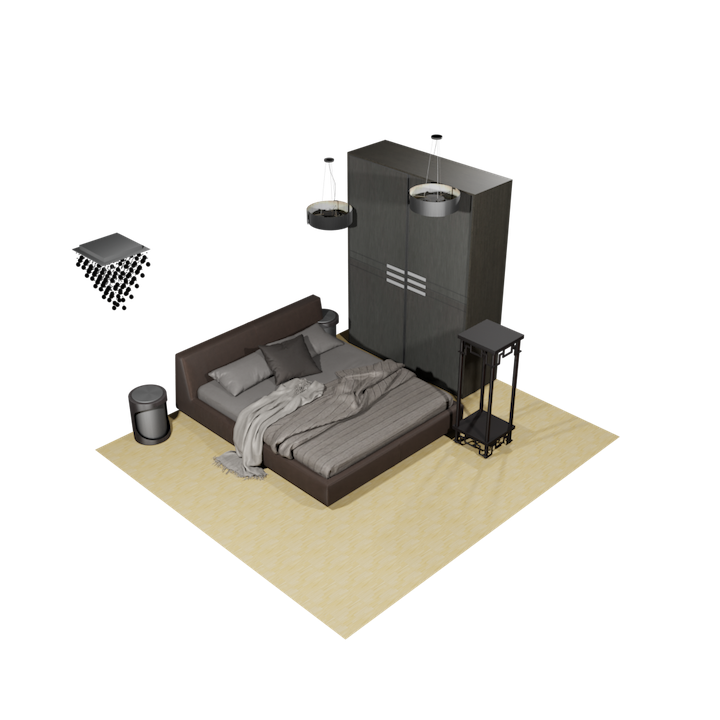}}
    \end{subfigure}
    \begin{subfigure}[b]{0.25\textwidth}
        \centering
        \adj{\includegraphics[width=\textwidth]{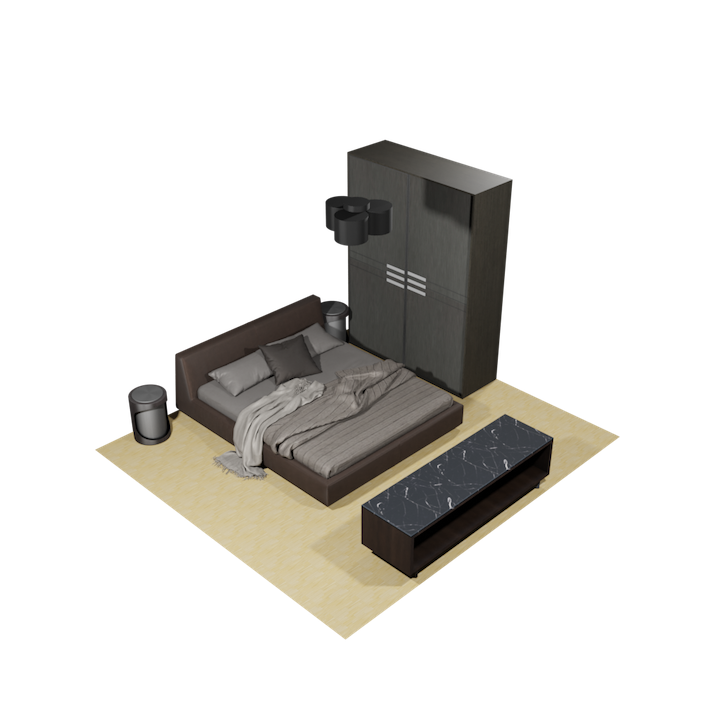}}
    \end{subfigure}

    \begin{subfigure}[b]{0.25\textwidth}
        \centering
        \adj{\includegraphics[width=\textwidth]{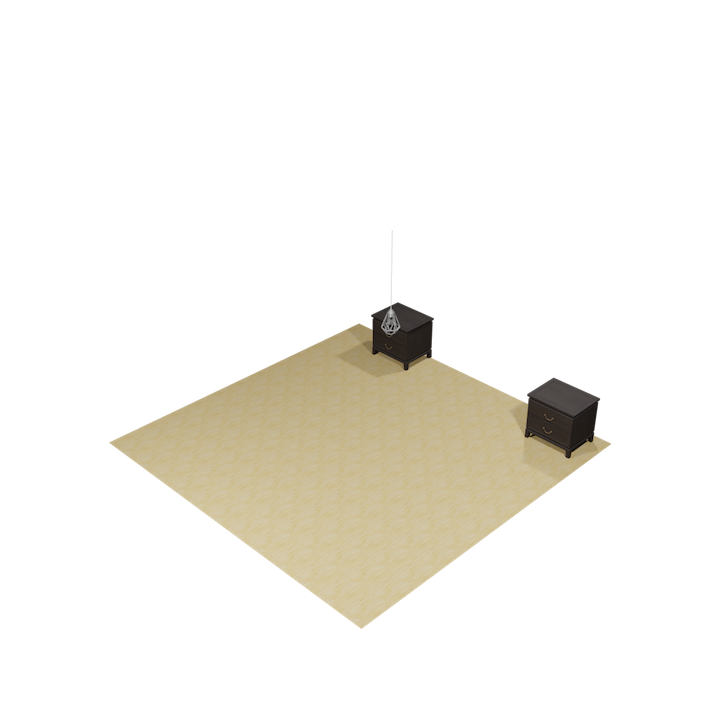}}
    \end{subfigure}
    \begin{subfigure}[b]{0.25\textwidth}
        \centering
        \adj{\includegraphics[width=\textwidth]{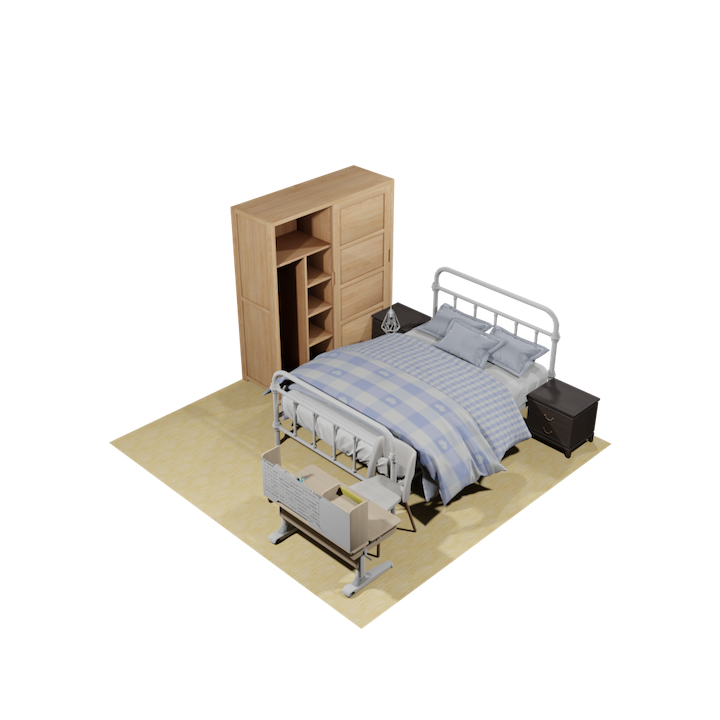}}
    \end{subfigure}
    \begin{subfigure}[b]{0.25\textwidth}
        \centering
        \adj{\includegraphics[width=\textwidth]{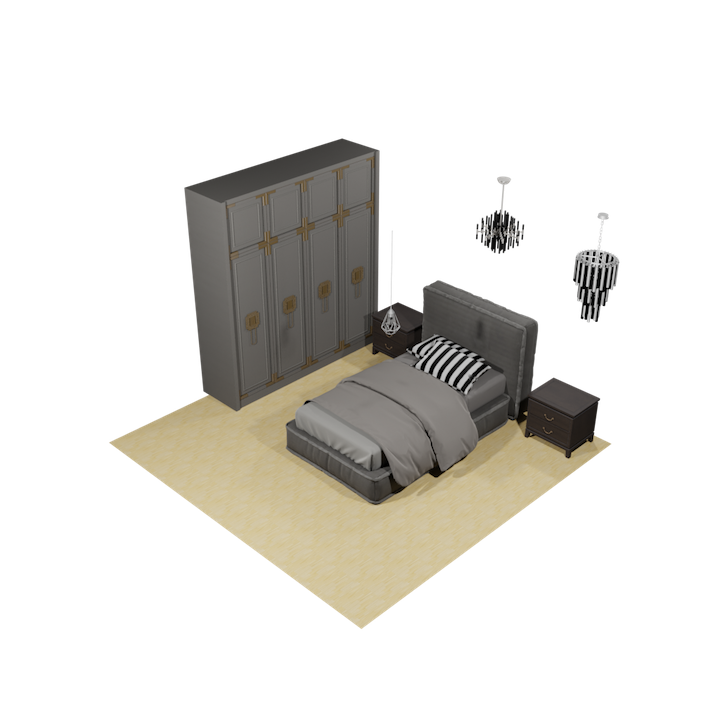}}
    \end{subfigure}

    \begin{subfigure}[b]{0.25\textwidth}
        \centering
        \adj{\includegraphics[width=\textwidth]{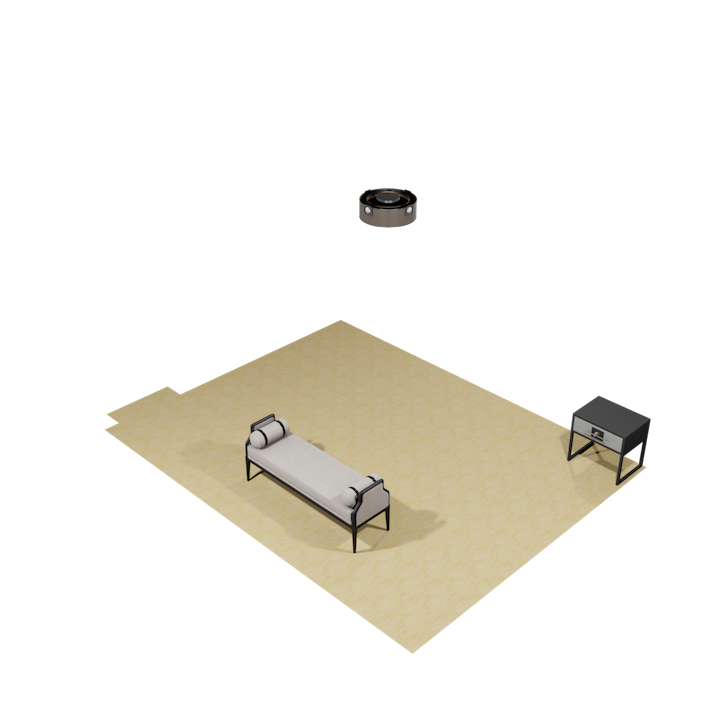}}
        \caption{Partial}
    \end{subfigure}
    \begin{subfigure}[b]{0.25\textwidth}
        \centering
        \adj{\includegraphics[width=\textwidth]{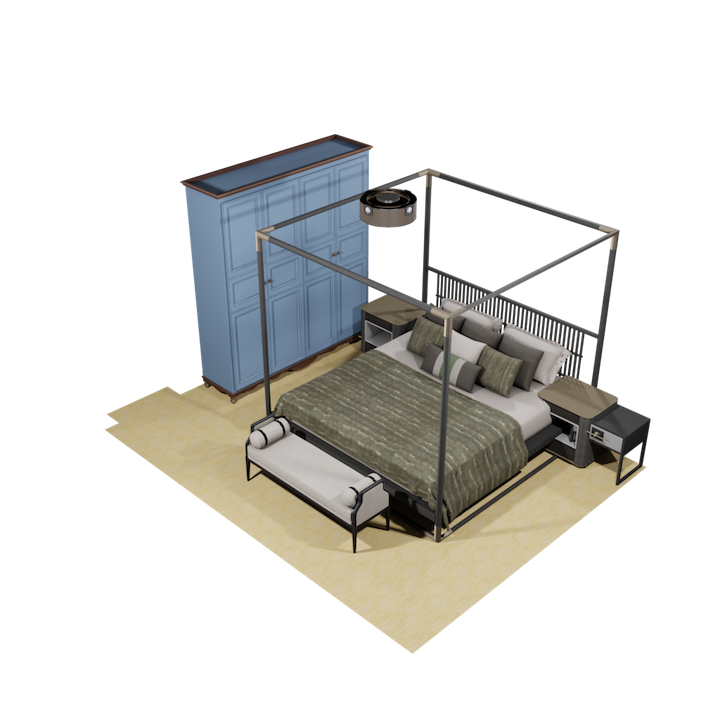}}
        \caption{Baseline}
    \end{subfigure}
    \begin{subfigure}[b]{0.25\textwidth}
        \centering
        \adj{\includegraphics[width=\textwidth]{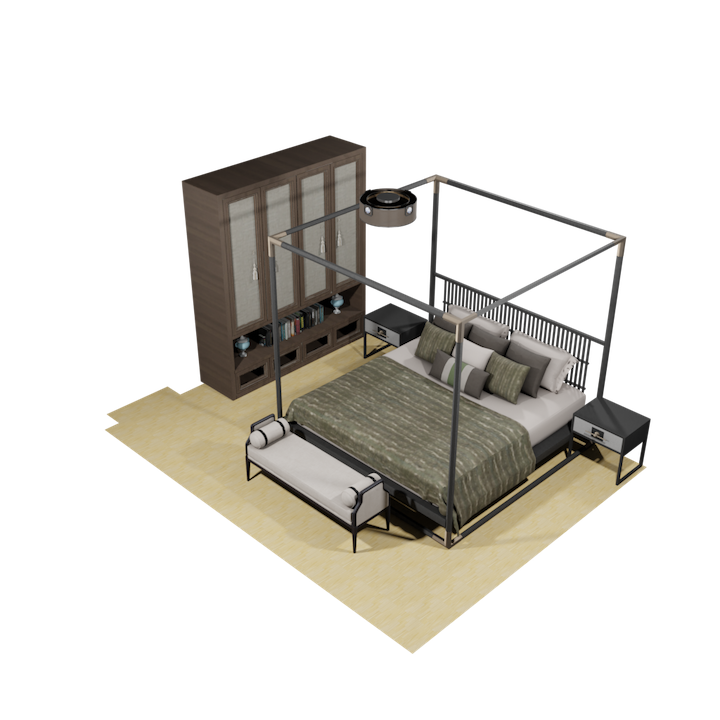}}
        \caption{Ours}
    \end{subfigure}

    \caption{\label{app_fig:scene_comp_against_baseline} \textbf{Partial scene completion of our model against the baseline:} From the leftmost to the rightmost columns are the partial scenes, the completed scenes by the baseline, and those by our model.}
\end{figure}

\begin{figure}
    \centering    
    \captionsetup[subfigure]{labelformat=empty}
    \centering
    
    \begin{subfigure}[b]{0.25\textwidth}
        \centering
        \adj{\includegraphics[width=\textwidth]{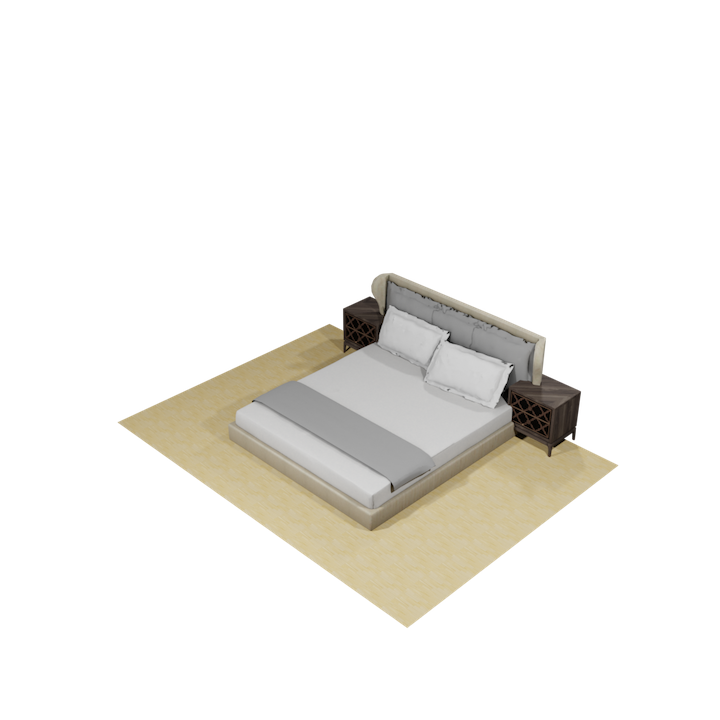}}
    \end{subfigure}
    \begin{subfigure}[b]{0.25\textwidth}
        \centering
        \adj{\includegraphics[width=\textwidth]{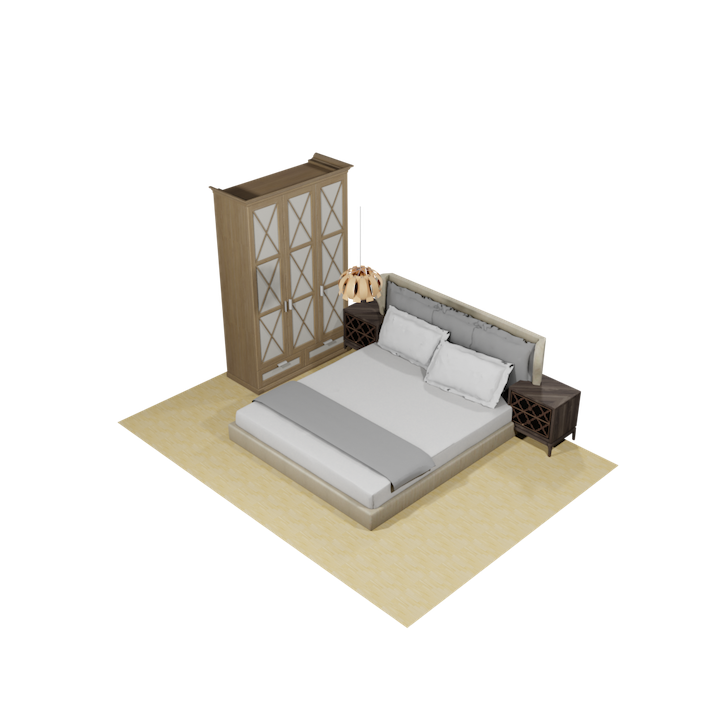}}
    \end{subfigure}
    \begin{subfigure}[b]{0.25\textwidth}
        \centering
        \adj{\includegraphics[width=\textwidth]{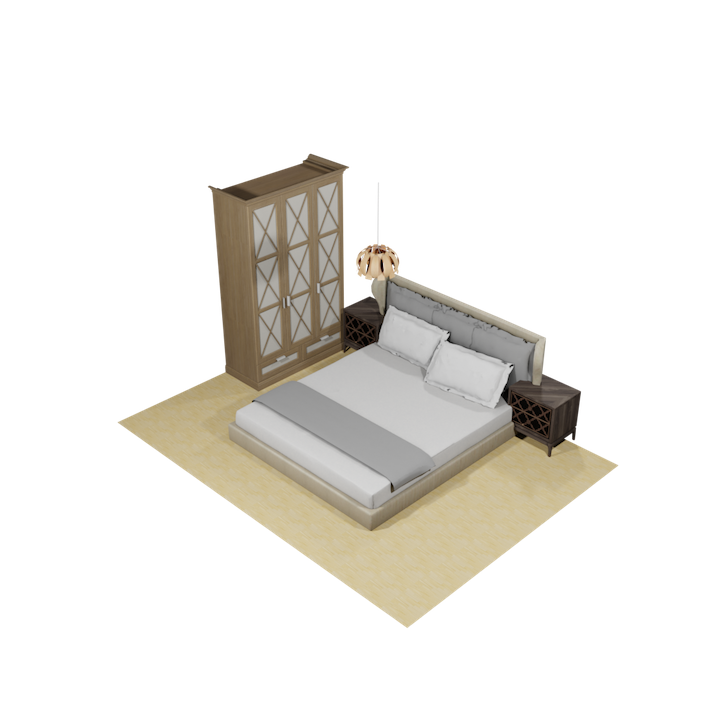}}
    \end{subfigure}

    \begin{subfigure}[b]{0.25\textwidth}
        \centering
        \adj{\includegraphics[width=\textwidth]{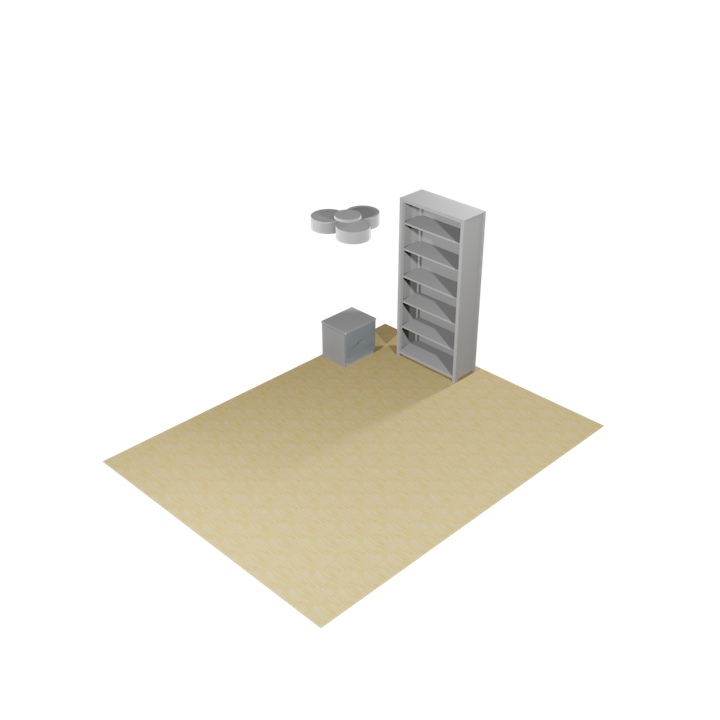}}
    \end{subfigure}
    \begin{subfigure}[b]{0.25\textwidth}
        \centering
        \adj{\includegraphics[width=\textwidth]{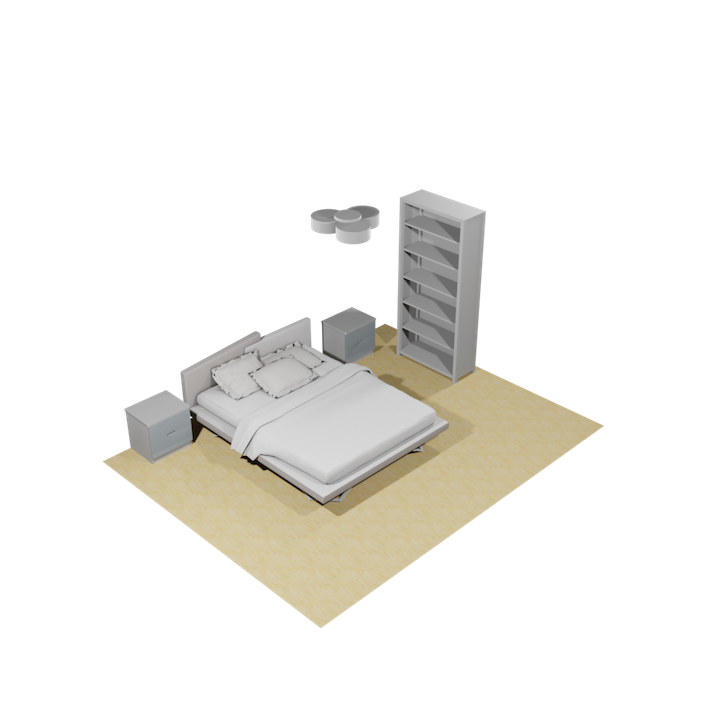}}
    \end{subfigure}
    \begin{subfigure}[b]{0.25\textwidth}
        \centering
        \adj{\includegraphics[width=\textwidth]{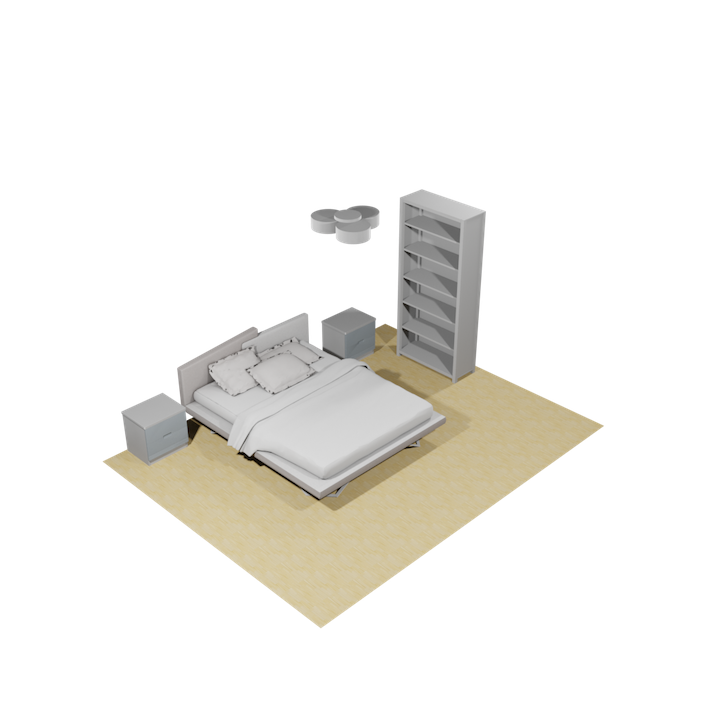}}
    \end{subfigure}

    \begin{subfigure}[b]{0.25\textwidth}
        \centering
        \adj{\includegraphics[width=\textwidth]{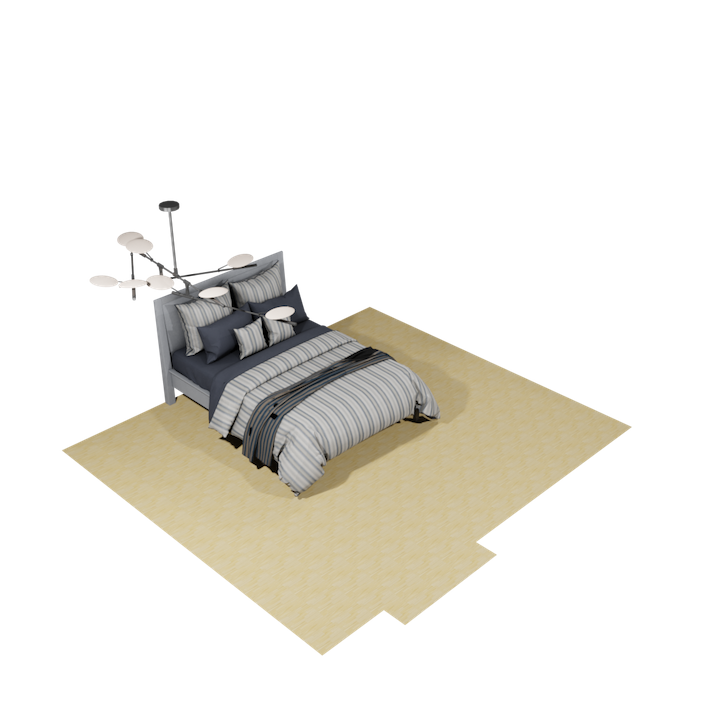}}
    \end{subfigure}
    \begin{subfigure}[b]{0.25\textwidth}
        \centering
        \adj{\includegraphics[width=\textwidth]{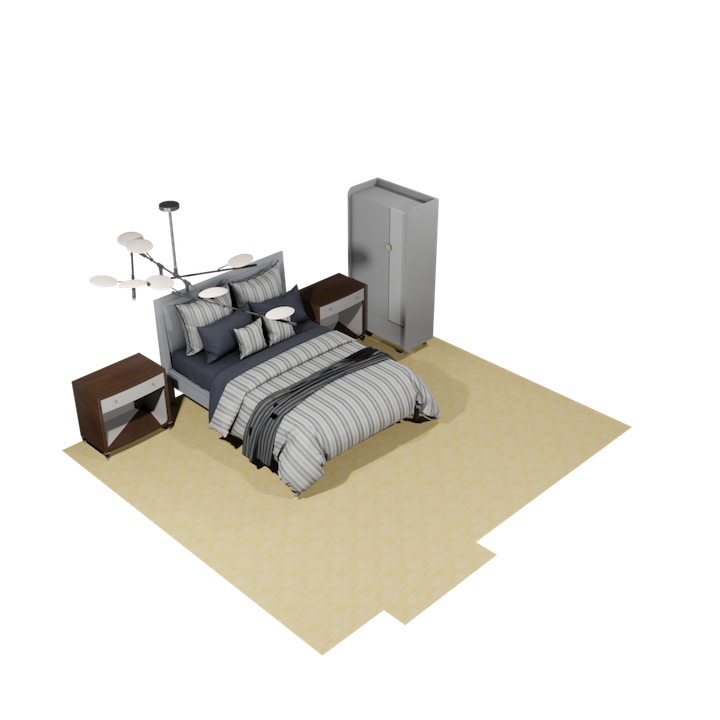}}
    \end{subfigure}
    \begin{subfigure}[b]{0.25\textwidth}
        \centering
        \adj{\includegraphics[width=\textwidth]{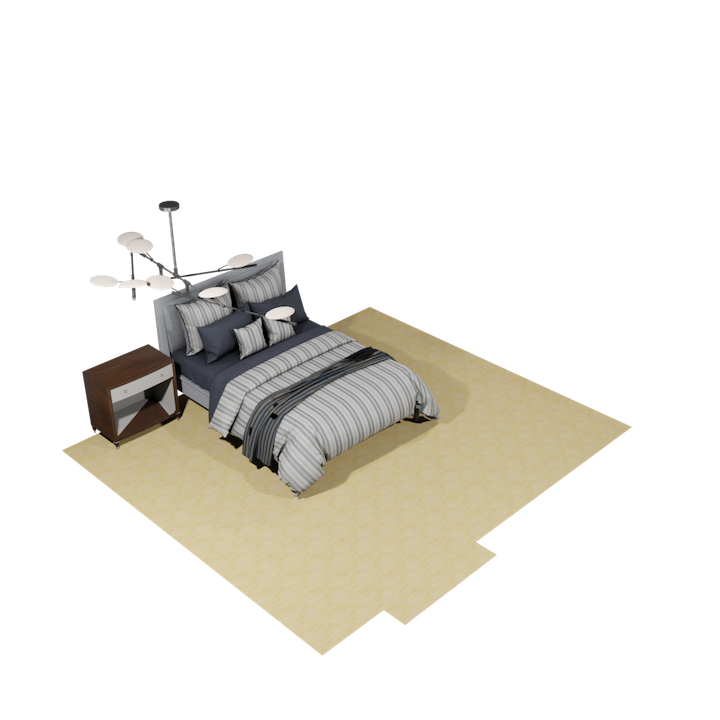}}
    \end{subfigure}

    \begin{subfigure}[b]{0.25\textwidth}
        \centering
        \adj{\includegraphics[width=\textwidth]{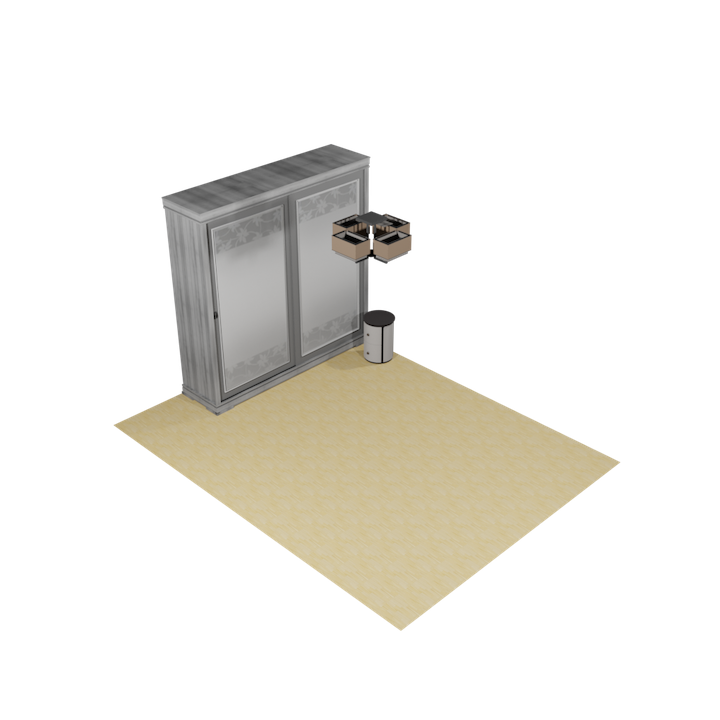}}
    \end{subfigure}
    \begin{subfigure}[b]{0.25\textwidth}
        \centering
        \adj{\includegraphics[width=\textwidth]{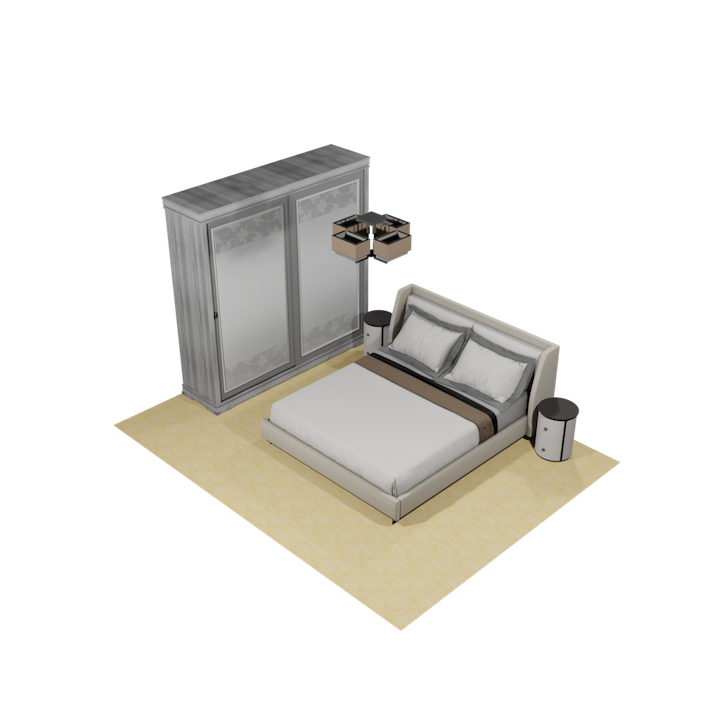}}
    \end{subfigure}
    \begin{subfigure}[b]{0.25\textwidth}
        \centering
        \adj{\includegraphics[width=\textwidth]{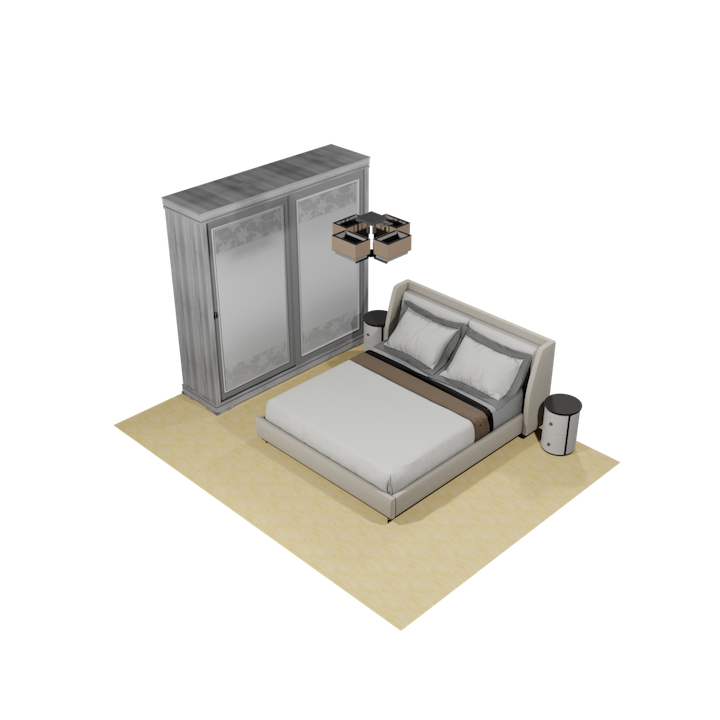}}
    \end{subfigure}

    \begin{subfigure}[b]{0.25\textwidth}
        \centering
        \adj{\includegraphics[width=\textwidth]{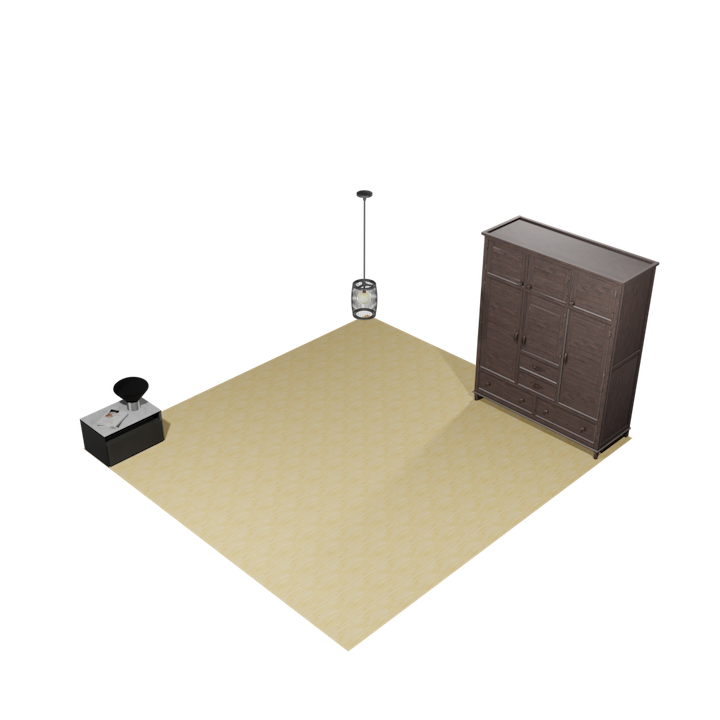}}
    \end{subfigure}
    \begin{subfigure}[b]{0.25\textwidth}
        \centering
        \adj{\includegraphics[width=\textwidth]{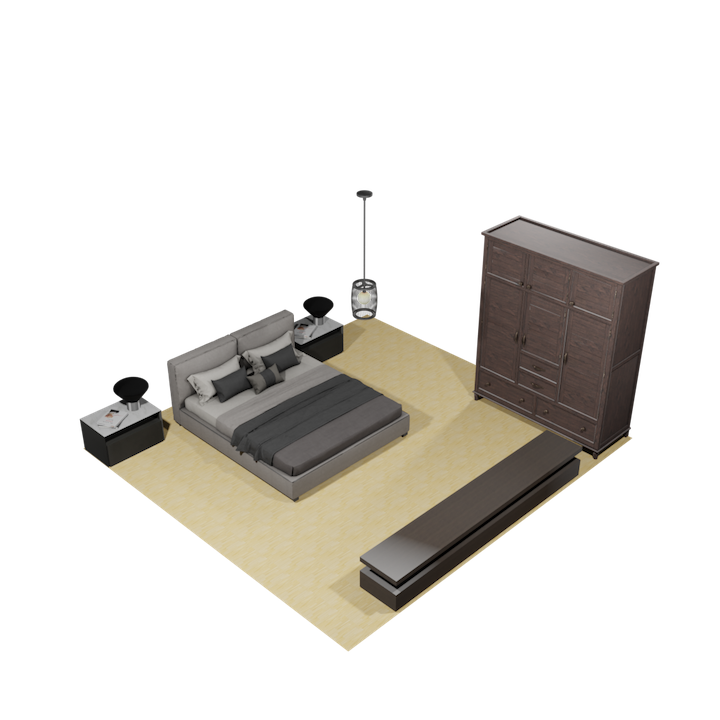}}
    \end{subfigure}
    \begin{subfigure}[b]{0.25\textwidth}
        \centering
        \adj{\includegraphics[width=\textwidth]{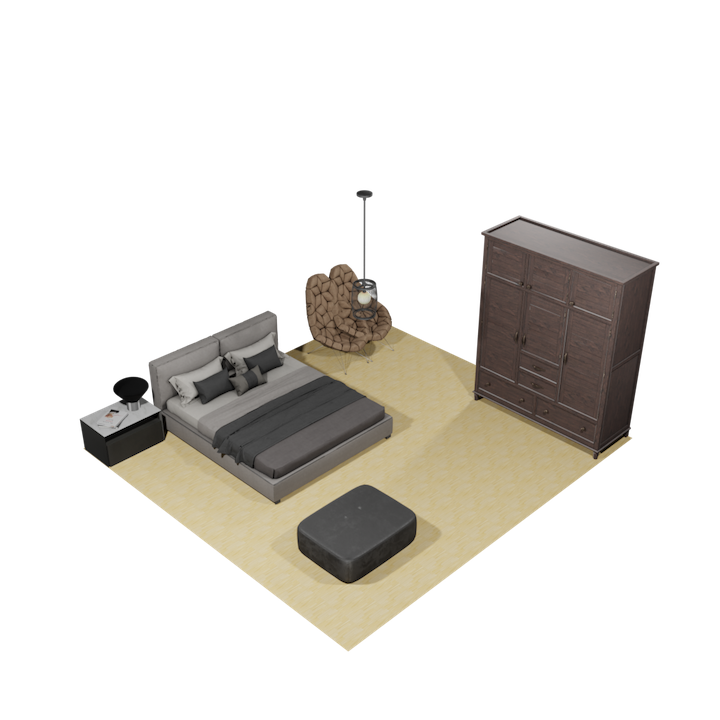}}
    \end{subfigure}

    \begin{subfigure}[b]{0.25\textwidth}
        \centering
        \adj{\includegraphics[width=\textwidth]{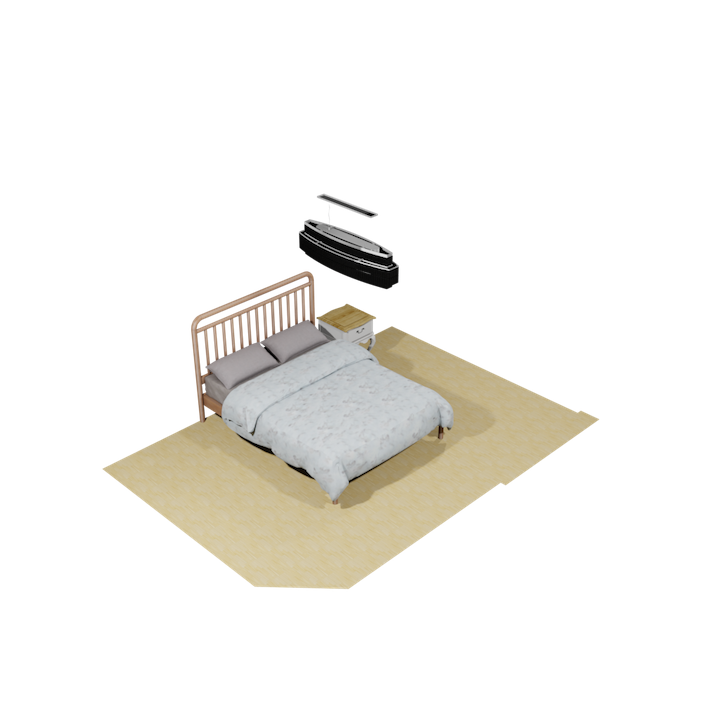}}
        \caption{Partial}
    \end{subfigure}
    \begin{subfigure}[b]{0.25\textwidth}
        \centering
        \adj{\includegraphics[width=\textwidth]{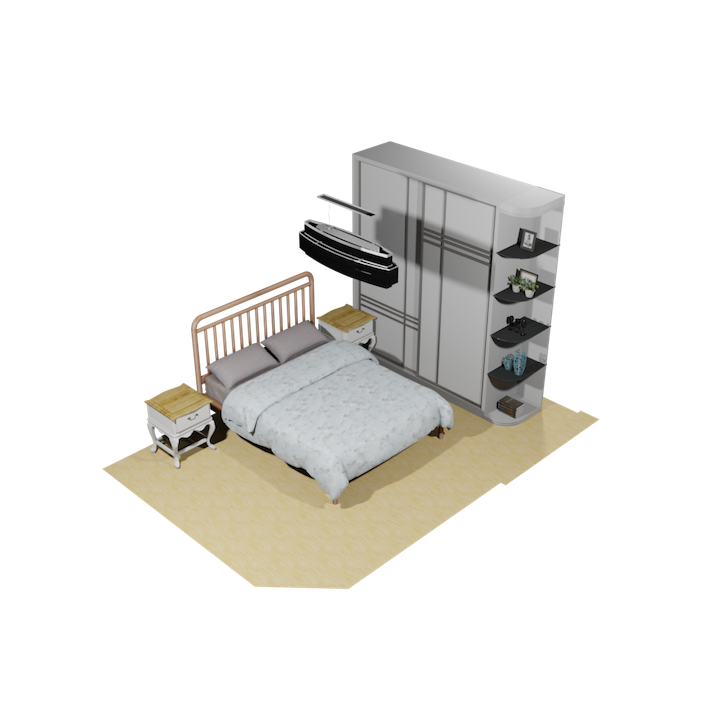}}
        \caption{Ours}
    \end{subfigure}
    \begin{subfigure}[b]{0.25\textwidth}
        \centering
        \adj{\includegraphics[width=\textwidth]{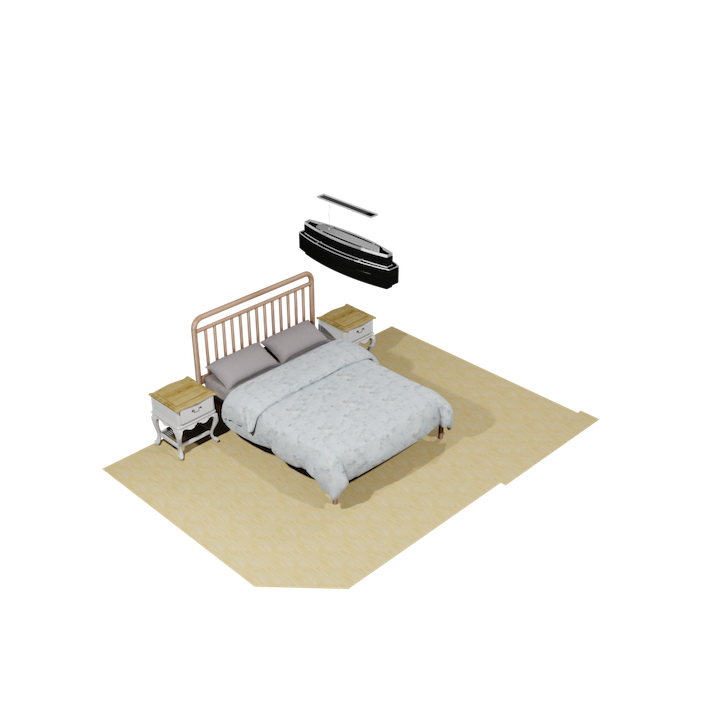}}
        \caption{Ground Truth}
    \end{subfigure}
    
    \caption{\label{app_fig:scene_comp_against_gt} \textbf{Partial scene completion of our model against the ground truth scenes:} From the leftmost to the rightmost columns are the partial scenes, the completed scenes by our model, and the ground truth scenes.}
\end{figure}

In this section, we provide more visual samples of scenes generated by our model in different settings to complement our previous quantitative and qualitative results and analysis:

\begin{enumerate}
    \item \textit{Text-guided Furniture Replacement}: Figure \ref{app_fig:text_repl} shows examples of replacing a certain furniture in an existing scene with text prompts describing the new one.
    
    \item \textit{Text-guided Synthesis}: In Figure \ref{app_fig:text_gen}, we list more examples of how our model can generate scenes which follow the visual clues from the text prompts.

    \item \textit{Scene Synthesis from Floor Layouts}: We showcase more examples of the scenes generated based on test floor plans by our model in Figure \ref{app_fig:uncond_gen}, on the top of which we also give some samples from ATISS for contrast.

    \item \textit{Partial Scene Completion}: Figure \ref{app_fig:scene_comp_against_baseline} compares our model against the baseline in partial scene completion task (same as \ref{fig:scene_comp}), which qualitatively demonstrates the aesthetic superiority over ATISS. Figure \ref{app_fig:scene_comp_against_gt} shows how well our model performs against the ground truth scenes. Note that our model can not only reconstruct the ground truth faithfully but also generalize to other furniture.
\end{enumerate}

\end{document}